%% file: ms.tex
\documentclass[sn-mathphys,Numbered,table,xcdraw]{sn-jnl}

\usepackage{graphicx}%
\usepackage{multirow}%
\usepackage{amsmath,amssymb,amsfonts}%
\usepackage{amsthm}%
\usepackage{mathrsfs}%
\usepackage[title]{appendix}%
\usepackage[table,xcdraw]{xcolor}%
\usepackage{textcomp}%
\usepackage{manyfoot}%
\usepackage{booktabs}%
\usepackage{algorithm}%
\usepackage{algorithmicx}%
\usepackage{algpseudocode}%
\usepackage{listings}%
\usepackage{float}
\usepackage{placeins}

\usepackage{xspace}
\usepackage{mathtools}
\usepackage{rotating}
\usepackage[nohyperlinks, printonlyused, withpage, smaller]{acronym}
\DeclarePairedDelimiter\abs{\lvert}{\rvert}

\newcommand{\eg}{e.\,g.,\xspace}
\newcommand{\ie}{i.\,e.,\xspace}
\newcommand{\cf}{cf.\xspace}
\newcommand{\Eg}{E.\,g.,\xspace}

\usepackage{microtype}

\newcommand{\eat}[1]{}

\theoremstyle{thmstyleone}%
\theoremstyle{thmstyletwo}%

\theoremstyle{thmstylethree}%
\newtheorem{definition}{Definition}%

\begin{document}

\title[Local and Global Representations of Attention on Time Series]{Extracting Interpretable Local and Global Representations from Attention on Time Series}

\author*[1]{\fnm{Leonid} \sur{Schwenke}}\email{leonid.schwenke@uni-osnabrueck.de}

\author[1,2]{\mbox{\fnm{Martin} \sur{Atzmueller}}}\email{martin.atzmueller@uni-osnabrueck.de}

\affil[1]{\orgname{Osnabr\"uck University}, \orgaddress{Semantic Information Systems Group, \street{Wachsbleiche 27}, \postcode{49090}, \city{Osnabr\"uck}, \country{Germany}}}
\affil[2]{\orgname{German Research Center for Artificial Intelligence (DFKI)},
\orgaddress{\street{Hamburger Str. 24}, \postcode{49084}, \city{Osnabrück},  \country{Germany}}}

\abstract{
This paper targets two transformer attention based interpretability methods working with local abstraction and global representation, in the context of time series data. We distinguish local and global contexts, and provide a comprehensive framework for both general interpretation options. We discuss their specific instantiation via different methods in detail, also outlining their respective computational implementation and abstraction variants. Furthermore, we provide extensive experimentation demonstrating the efficacy of the presented approaches. In particular, we perform our experiments using a selection of univariate datasets from the UCR UEA time series repository where we both assess the performance of the proposed approaches, as well as their impact on explainability and interpretability/complexity. Here, with an extensive analysis of hyperparameters, the presented approaches demonstrate an significant improvement in interpretability/complexity, while capturing many core decisions of and maintaining a similar performance to the baseline model. Finally, we draw general conclusions outlining and guiding the application of the presented methods.

}
\keywords{Transformer, Attention, Symbolic Approximation, Interpretability, Data Abstraction, Time Series Classification}

\maketitle
\DeclareRobustCommand{\apen}{\ac{ApEn}\xspace}
\DeclareRobustCommand{\ce}{\ac{CE}\xspace}
\DeclareRobustCommand{\cv}{\ac{CV}\xspace}

\DeclareRobustCommand{\crcam}{\ac{CCAM}\xspace}
\DeclareRobustCommand{\crcams}{\ac{CCAM}s\xspace}

\DeclareRobustCommand{\fcam}{\ac{FCAM}\xspace}
\DeclareRobustCommand{\fcams}{\ac{FCAM}s\xspace}

\DeclareRobustCommand{\gcr}{\ac{GCR}\xspace}
\DeclareRobustCommand{\gcrs}{\ac{GCR}s\xspace}
\DeclareRobustCommand{\pgcr}{\ac{GCR-P}\xspace}
\DeclareRobustCommand{\pgcrs}{\ac{GCR-P}s\xspace}
\DeclareRobustCommand{\tgcrs}{\ac{GCR-T}s\xspace}
\DeclareRobustCommand{\tgcr}{\ac{GCR-T}\xspace}

\DeclareRobustCommand{\gswa}{\ac{GSA}\xspace}
\DeclareRobustCommand{\gswpa}{\ac{GSA-P}\xspace}
\DeclareRobustCommand{\tgswa}{\ac{GSA-T}\xspace}

\DeclareRobustCommand{\gtm}{\ac{GTM}\xspace}
\DeclareRobustCommand{\gtms}{\ac{GTM}s\xspace}

\DeclareRobustCommand{\gva}{\ac{GVA}\xspace}

\DeclareRobustCommand{\laam}{\ac{LAMA}\xspace}
\DeclareRobustCommand{\laams}{\ac{LAMA}s\xspace}

\DeclareRobustCommand{\lasa}{\ac{LASA}\xspace}
\DeclareRobustCommand{\lasas}{\ac{LASA-S}\xspace}

\DeclareRobustCommand{\laav}{\ac{LAVA}\xspace}
\DeclareRobustCommand{\mha}{\ac{MHA}\xspace}
\DeclareRobustCommand{\nlp}{\ac{NLP}\xspace}
\DeclareRobustCommand{\sampen}{\ac{SampEn}\xspace}
\DeclareRobustCommand{\sax}{\ac{SAX}\xspace}
\DeclareRobustCommand{\svden}{\ac{SvdEn}\xspace}

\DeclareRobustCommand{\xai}{\ac{XAI}\xspace}

\input{s1_intro}
\input{s3_background_knowledge}
\input{s4_methods}
\input{s5_results}
\input{s2_related_work}

\input{s6_discussion}
\input{s7_conclusion_outlook}
\bibliography{main}
\input{s8_appendix}

\newpage

\end{document}

%% file: s1_intro.tex
\section{Introduction}
\label{sec:introduction}
In the last years, the research interest in explainable and interpretable Artificial Intelligence (AI) strongly increased~\cite{BarredoArrieta:20}. This research area, which is often also referred to as \xai, provides methods and tools to increase the transparency and understanding of black box AI models and their results. This is especially important for critical application areas where trust and confidence into the model is mandatory to justify its use~\cite{BarredoArrieta:20,rojat2021explainable}. Nevertheless, the understanding of deep neural networks is still limited and thus researchers are still facing several open challenges, \eg~\cite{carvalho2019machine,VAT:21}. In particular, this holds for time series data, due to the less intuitive nature of the data, \eg compared to \nlp or \cv, \cf~\cite{rojat2021explainable}. An important part of XAI is interpretability~\cite{palacio2021xai}, \eg ``the degree to which an observer can understand the cause of a decision''~\citep{miller2019explanation}.

Advances towards better interpretability emerged though the Transformer architecture and its sub-architectures with its internal Attention mechanism. Nowadays, the Transformer makes up the state-of-the-art for multiple challenges. While most research on Transformers focuses on \nlp and \cv, recently also multiple new successful work on time series data emerged~\citep{wen2022transformers}. 
However, while the Attention mechanism inside the \mha seem to highlight the importance of different input combinations and therefore promote interpretability; \citep{baan2019understanding} found, though, that Attention is only partially interpretable, where multiple heads can be pruned without reducing the accuracy \cite{pruthi2019learning}. Therefore, overall in general we still do not fully understand how the MHA contributes to the performance of Transformer and how we can handle the Attention values, of which multiple exist throughout the whole architecture \cite{pruthi2019learning,baan2019understanding,clark2019does, baan2019transformer, ramsauer2020hopfield}. It is argued that Attention highlights the importance of elements, regardless of the class \cite{baan2019understanding,wang2020language,SA:21:local}. Multiple works thus focus on analysing the interpretability of the Attention mechanism in the Transformer~\citep{baan2019understanding, pruthi2019learning, ramsauer2020hopfield, clark2019does, wang2020language, baan2019transformer}; those are, however, mostly related to the context of NLP and CV. In the context of time series data interpretation might work differently, \eg the experiments from \cite{wen2022transformers} indicate that time series tasks are more successful on lower layer counts (3-6), compared to 12-128 in CV and NLP. This might suggest that the information flow in Transformers and its Attention works differently on time series tasks. \cite{ismail2020benchmarking} highlights this further, by showing that current saliency methods from different contexts are not well applicable on time series data. This even further underlines the need for an exploration of Attention on time series data, which we want to advance as part of this work.

\begin{figure}[htb!]
	\centering
	\includegraphics[width=1.0\columnwidth]{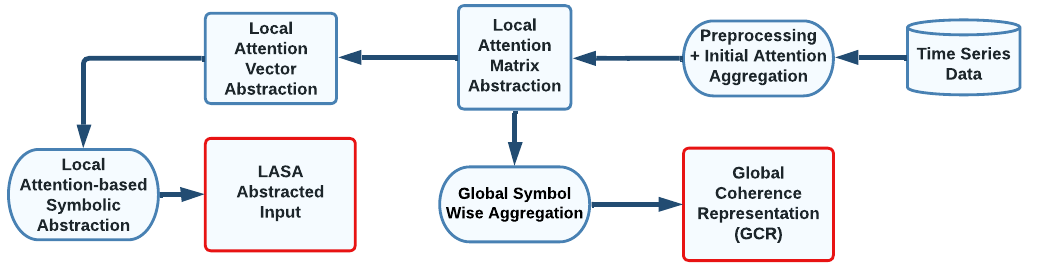}
	\caption{Simple Pipeline of the aggregation process of the global and local approach.}
	\label{fig:simplePipeline}
\end{figure}

To tackle this challenge, we already introduced two methods to enhance the interpretability specifically on time series data. One is called \lasa \citep{SA:21:local} and is focusing on the abstraction of local input data, to find the most interesting local characteristics for each class. The second one is the \gcr \citep{SA:21:global}, a global class dependent saliency based representation to show globally how all symbols influences each other at each possible time point. Figure \ref{fig:simplePipeline} shows a simple pipeline for both methods to highlight their relation.

This paper is an adapted and significantly extended revision of~\cite{SA:21:local, schwenke2021abstracting, SA:21:global}.
In summary, we have extended on the following issues:
\begin{enumerate}
    \item We provide a comprehensive view on the topic, for an integrated discussion of~\cite{SA:21:local, schwenke2021abstracting, SA:21:global}, providing a uniform terminology.
    \item Furthermore, we introduce the following extensions of our methods:
    \begin{enumerate}
        \item \lasas is a variation of \lasa which uses the widely known Shapelets algorithm \cite{ye2009time} on the abstracted data, to extract more simple shapes in the less complex data. The idea would be, that the Shapelets are therefore also more simple and thus better accessible.
        \item \tgcr uses a threshold to cut off low Attention values while creating the \gcr and thus should show that less attended data can be somewhat neglected.
        \item Because the \gcr does not consider how other classes look like, we introduce the \pgcr , which adds a penalty to each class representation, to better highlight the most significant differences between classes.
    \end{enumerate}
    \item Finally, we provide extended experimentation, exploring more XAI metrics, analysing more datasets as well as more variables and hyperparameters.
\end{enumerate}
For reproducibility, our code regarding implementation and experiments is provided in a publicly available GitHub repository\footnote{\url{https://github.com/cslab-hub/localAndGlobalJournal}}.

The remainder of the paper is structured as follows: In Section \ref{sec:backgroundknowledge} we first give an introduction to various methods and metrics, which are relevant for the remainder of the paper. Section \ref{sec:methods} focuses on our proposed local and global method and variations, including their mathematical and algorithmic formulas, as well as their application. We follow up in Section \ref{sec:results} with our experiment setup and our results, which we also partially discuss and interpret. In Section \ref{sec:relatedWork} we set our work into relation to other literature, to better define currently ongoing research and to better point out the novel elements of our approach. Afterwards in Section \ref{sec:discussion} we continue the discussion on the bigger picture of our completed results. Last but not least, we finish with a conclusion and give an outlook in Section \ref{sec:conclusion}. As further addition, we provide some complementary information and results in the Appendix. 

%% file: s3_background_knowledge.tex
\section{Background}
\label{sec:backgroundknowledge}
In this section, we provide introduce basic concepts, the Transformer architecture, as well as metrics used for performance and XAI evaluation.


\subsection{Trend Mining Methods}
This section describes the two prominent methods to help find relevant trend-structures in the data. 
\subsubsection{Symbolic Aggregate ApproXimation}
\label{sec:sax}

\begin{figure}[h!]
	\centering
	\includegraphics[width=0.92\columnwidth]{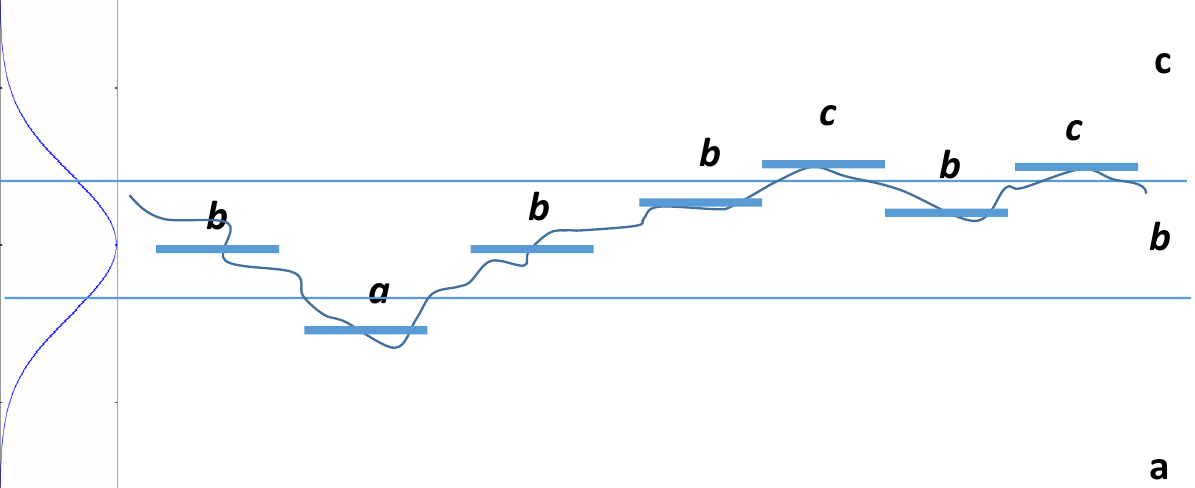}
	\caption{Example visualisation for a SAX discretization, \cf~~\cite{atzmueller2017explanation}: Each data point from the original time series is mapped to a discrete symbol (a, b, c) based on the quantiles from the standard normal distribution.}
	\label{fig:sax}
\end{figure}

The \sax is a prominent aggregation and abstraction technique for time series analysis, \eg~\cite{Lin:2003}, also enhancing interpretability and computational sensemaking via its specific representation, \cf~\cite{atzmueller2017explanation,RWA:19}. To reduce the complexity of a real-valued time series domain, the SAX algorithm transforms this into a discrete symbolic string representation in order to facilitate interpretation. The resulting high-level representation of time series data enhances the general trend of the data via its symbolic representation (\cf~Figure \ref{fig:sax}). Previously, this has been used, \eg to improve \emph{motif detection} in data due to the simpler data shape. One drawback of this method is that detailed data changes are getting lost, hence data where small changes are important need a large set of symbols. Also, \citep{rojat2021explainable} summarized some SAX based XAI methods for deep learning. 

\subsubsection{Shapelets}
\label{sec:shapelets}
Shapelets \cite{ye2009time} are time series subsequences -- \ie corresponding to or being interpretable as specific graphical shapes -- which aim to represent the most distinguished trend features for a class.
While the initial algorithm only finds one Shapelet per class, \citep{mueen2011logical} proposed a method for identifying multiple Shapelets per class. The current state-of-the-art approach \cite{rakthanmanon2013fast} uses the SAX algorithm to speed up the Shapelet mining. Due to the visual accessibility and the ability to find the most distinctions  between classes, Shapelets are seen as highly interpretable. For a more detailed discussion on existing deep learning XAI methods based on Shapelets, we refer to \citep{rojat2021explainable}. 

\subsection{Transformer Model}
\emph{Transformers}~\cite{vaswani2017Attention,tay2020efficient} have emerged as a prominent Deep Learning architecture for handling sequential data~\cite{vaswani2017Attention}. Originally designed for translation tasks in NLP, they are still typically applied in the NLP domain~\cite{vaswani2017Attention,devlin2018bert,clark2019does,tay2020efficient,bracsoveanu2020visualizing} as well as in CV~\cite{dosovitskiy2020image,carion2020end,caron2021emerging,yang2021focal},
Transformers have also recently started to be successfully applied to time series problems~\cite{wen2022transformers,song2018attend,lim2019temporal}, \eg addressing efficient architectures~\cite{tay2020efficient} and enhanced prediction approaches~\cite{li2019enhancing,wu2020adversarial} on time series. Especially due to the Transformer's ability to capture long-range dependencies, this architecture is gaining even more interest in the time series community \cite{wen2022transformers}.
It is important to note, that the given limitations of Transformers are currently a rather strong research topic in general; recently many slightly modified Transformer architectures arose which take on different limitations of the original Transformer~\cite{tay2020efficient}.
In comparison to those approaches, we consider the MHA structure in its original form~\cite{vaswani2017Attention} to create a standard baseline for our approach. Therefore, we also do not focus on scalability and runtime, but rather on comparability between the given tasks over multiple hyperparameters under consideration of our abstraction, interpretation and visualization approaches.

The \emph{Transformer} architecture~\cite{vaswani2017Attention,tay2020efficient} was originally applied on sequence-to-sequence tasks; it consists of an encoder and decoder block. For other use cases, like \eg in our classification tasks context, just the encoder is used (see Figure \ref{fig:transformerEncoder}). The following sections discuss the encoder part of the Transformer architecture as well as its included MHA in more detail.

\subsubsection{Encoder Model}
\label{sec:encoderModel}

\begin{figure}[ht!]
	\centering
	\includegraphics[width=0.95\columnwidth]{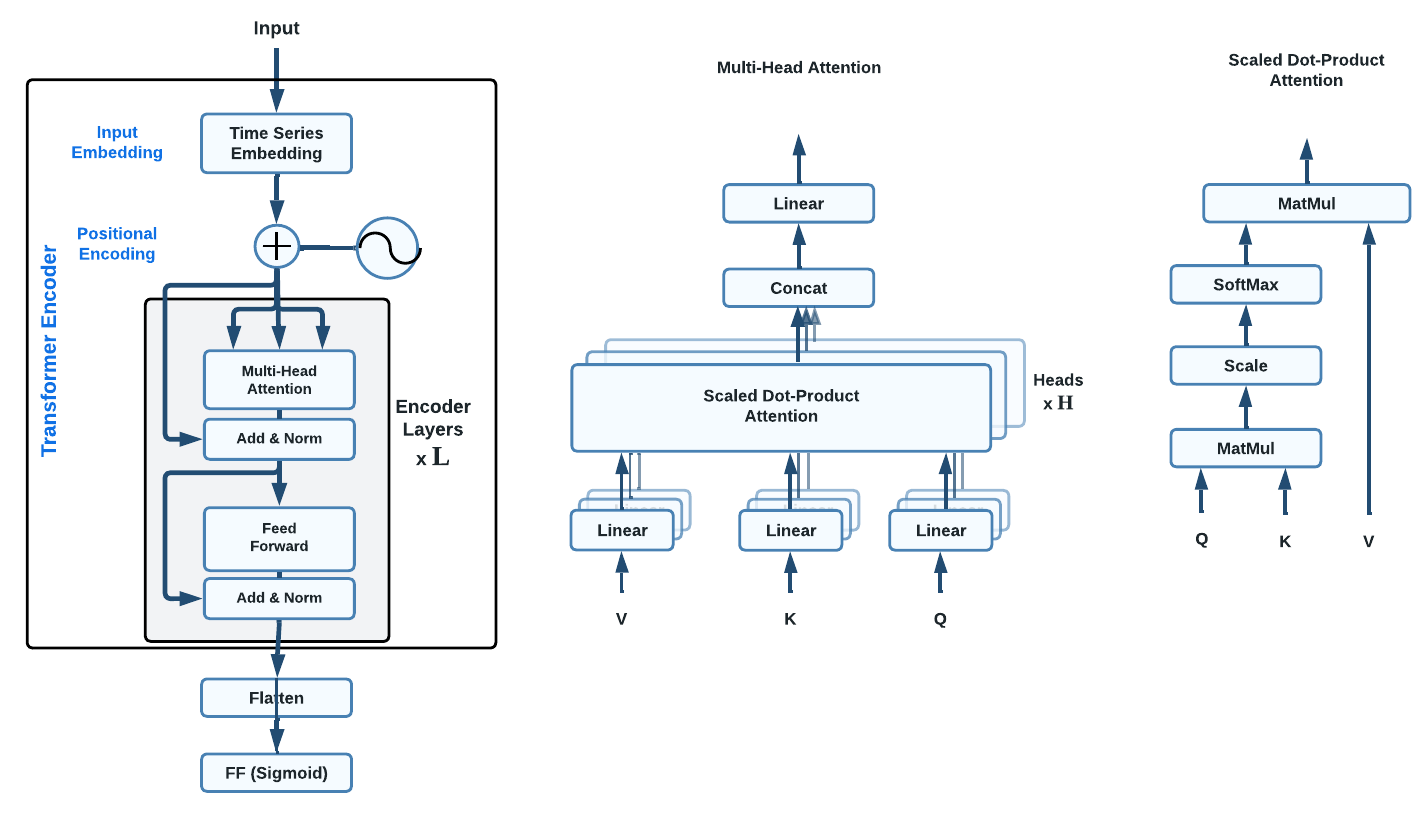}
	\caption{Transformer encoder \citep{vaswani2017Attention} model we adapted for our experiments. On the right, the Multi-Head Attention and Scale Dot-Product Attention components are shown. Image adapted from \citep{vaswani2017Attention}.}
	\label{fig:transformerEncoder}
\end{figure}

In general, the encoder from the Transformer architecture receives an input sequence $X_{in} = \{x_1, x_2,..., x_n\}$ of length $n$, where each element $x_i \in X_{in}$ of the input sequence has the dimensionality $d$. In Figure~\ref{fig:transformerEncoder}, a slightly modified encoder model can be seen, adapted for our classification task, which we also apply for our evaluation. In our experiments, $X_{in}$ is always a univariate time series, hence $d = 1$ in this case. Our proposed method is currently only adapted to univariate data, because multivariate data open up some further challenges, \eg differentiating the Attention between sequences and/or considering the cross-sequential influence between input symbols. However, in this work we test rolling out multivariate data into one sequence, to somewhat reduce this limitation. As for the embedding, we apply a simple mapping -- described in more detail in Section \ref{sec:prepro}. A positional encoding is afterwards added onto the embedded input. We use the original positional encoding, proposed by \citep{vaswani2017Attention}:
\begin{equation}
\mathit{PE}_{(\mathit{pos}, 2i)} = \sin(\mathit{pos}/10000^\frac{2i}{d_{m}})
\end{equation}\begin{equation}
\mathit{PE}_{(\mathit{pos}, 2i+1)} = \cos(\mathit{pos}/10000^\frac{2i}{d_{m}})
\end{equation}
\\
where $pos$ is the sequence position, $i$ is the dimensional position of the vector-element and $d_{m}$ is a parameter giving the output dimension of each sublayer in the encoder, including the positional encoding. Next $L$ times the encoder block is applied, including the Multi-Head Attention (MHA) -- which we want to better understand with our experiments -- as a core mechanic. As shown in Figure \ref{fig:transformerEncoder}, first the MHA is applied (described in more detail in Section \ref{sec:mha}), followed by a normalization layer:
\begin{equation}\text{LayerNorm}(F(X), X)\end{equation}
which uses the output from the previous sub-layer $F(X)$ (in our case the MHA or a point-wise feed-forward layer (FFN)) and the input $X$ of the previous sub-layer. This is also called skip-connection. The output from the normalization layer is feed into an FFN, which is again followed by a normalization layer. 

\begin{equation}\mathit{FFN(X)} = \mathit{ReLu}(XW_1 + b_1) W_2 + b_2\end{equation}
\\
with the learnable weights $W_1 \in \mathbb{R}^{d_{m} \times d_{\mathit{ff}}}$, $W_2 \in  \mathbb{R}^{d_{m} \times d_{\mathit{ff}}}$ and biases $b_1 \in \mathbb{R}^{d_{\mathit{ff}}}$, $b_2 \in \mathbb{R}^{d_{m}}$. The variable $d_{\mathit{ff}}$ is the so called inner-layer dimensionality. For the activation function, we use $\mathit{ReLU}$. As for our specific and slightly modified classification model; the output of the $N$-th encoder block is flattened afterwards and classified with a fully connected sigmoid based layer. 

\FloatBarrier
\subsubsection{Multi-Head Attention}
\label{sec:mha}

\begin{figure}[ht!]
	\centering
	\includegraphics[width=0.92\columnwidth]{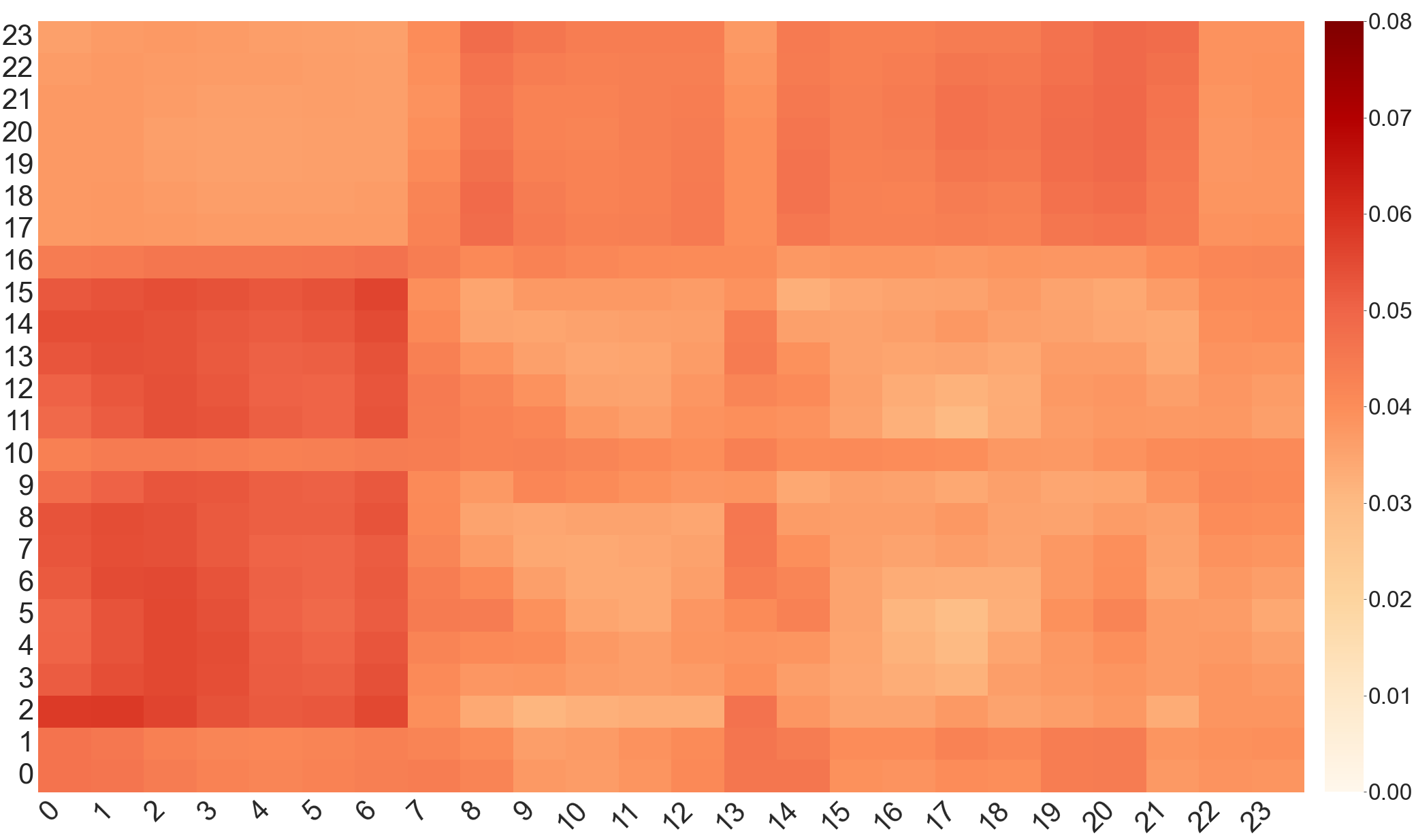}
	\caption{Example for an Attention matrix from one head of the MHA. Both axes show the same input sequence (self-Attention) and both sequences start in the bottom left corner. It can be seen how the elements at each position highlight each other in the given Attention head.}
	\label{fig:AttentionExample}
\end{figure}

The Multi-Head Attention (MHA) and its Attention matrices are core elements of the Transformer architecture and are noted as one of the mechanics that contribute the most to the success of the Transformer model. As we already elaborated, the influence of the MHA on the results is still not fully understood, nor are the Attention matrices fully interpretable. Some work aimed to analyse different aspects of the MHA matrices, which are normally referred to as Attention matrices, for which one example of an Attention matrix is depicted in Figure \ref{fig:AttentionExample}. Here, \citep{ramsauer2020hopfield} analysed the pattern filtering ability of Transformer Attention, showing that the first layers do a more generic averaging, while the later ones are more class specific and are the ones still learned for fine-tuning at the end of training. \citep{clark2019does} and \citep{wang2020language} show further examples of the later layer providing more detailed information for their interpretation methods. 

Another example of successfully extracting information over Attention was given by \citep{wang2020language}. They used the strongest Attention connections between words to extract knowledge, showing that high Attention values can be used to find important linguistic connections between words. We already used this property of Attention in \citep{SA:21:local,schwenke2021abstracting} to show that abstracting data is also possible on time series data over different aggregated Attention matrices.

Figure \ref{fig:transformerEncoder} shows the internal components of the MHA. It consists of $H$ parallel \textit{Scaled Dot-Product Attention} heads, which receive $V$, $K$ and $Q$ as input. In most cases, so called self-Attention is applied where all inputs are the same, as shown in Figure \ref{fig:transformerEncoder}.

In~\citep{vaswani2017Attention} the MHA is defined as follows:
\begin{equation}
\text{MultiHead}(Q,K,V ) = \text{Concat}(\text{head}_1,...,\text{head}_h)W^O
\end{equation}
\begin{equation}
\text{with head}_i = \text{Attention}(QW^Q_i ,KW^K_i ,V W^V_i )
\end{equation}
\begin{equation}
\text{Attention}(Q,K,V ) = \text{softmax}(\frac{QK^T}{\sqrt{d_k}})V
\end{equation}
$\text{Attention}(Q,K,V)$ is the scale-dot-product, $\sqrt{d_k}$ a scaling factor based on the dimensionality of $K$ and $W^Q_i$, $W^K_i$, $W^V_i$ are learnable weights in the dimensionality of $Q$, $K$, $V$.
$A = \text{softmax}(\frac{QK^T}{\sqrt{d_k}})$ is commonly referred to as the Attention matrix $A$, of size $l \times l$ with each element in [0,1], hence showing how strong each input in V is attended. In the following, $A_h^n$ refers to the Attention matrix in the $h$-th head of the $n$-th encoder layer.

Figure \ref{fig:AttentionExample} shows one example of an Attention matrix, \ie for self-Attention. One can read the Attention matrix as follows: Each axis represents the same sequence -- for self-Attention -- where in our visualizations the sequences start in the bottom left corner. Attention value $a_{ij}\in A$ highlights how strong one element $x_i\in X$ at a specific position $i$ in the encoder input sequence $X$ attends another element  $x_j \in X$ in the sequence at position $j$, thus $a_{11}$ shows how the first element in the sequence X attends itself. While it might seem that the Attention values correspond to importance, this is not that simple \cite{jain2019Attention,baan2019understanding,clark2019does}, especially because the MHA is only one part of the whole model and multiple Attention matrices exist throughout the whole architecture. Therefore, it is not clear how to read, combine or evaluate the Attention matrices. Additionally to note is that $X_h^l$ is the corresponding input related to $A_h^l$, \ie an aggregation of all previous inputs and operations, but for the analysis of (aggregated) Attention often $X_{in}$ is taken. This further shows that interpreting Attention is not quite as simple as it seems. In this work, we analyse different aggregations of Attention matrices on time series data, show some examples on what can be achieved with quite simple interpretation methods and how the different aggregations affect the information extraction of those methods.
\subsection{Metrics}
\label{sec:metricDef}
Besides typical performance measurements also multiple explanation and interpretation evaluation measurements getting more important. To clarify which measurements we use and how we calculate them, we present an overview in the following section.

\subsubsection{Performance Metrics}
\label{sec:perfMetrics}
To compare the results and measure the performance of our analysed methods, we use four classic different performance metrics. To calculate each we use the Sklearn \cite{sklearn_api} toolkit. We namely look into the accuracy, recall, precision and F1-score.

\subsubsection{Complexity Metrics}

To assess the abstraction ability of our local approach, we introduce four complexity measurements. The general idea is that the simpler the time series, the easier it is accessible to humans, in the sense, that simpler patterns in the data promote the interpretability of the underlying problem similarly as \emph{motifs} and \emph{shapelets} do. With those measures, we want to further quantify the simplicity vs. accuracy ratio of multiple aggregations, in order to better understand the effects different aggregation properties have on the MHA, as we already did with three metrics for our local approach in \cite{schwenke2021abstracting}. To calculate the metrics Singular Value Decomposition Entropy, Approximate Entropy and Sample Entropy, we use the Antropy package\footnote{\url{https://github.com/raphaelvallat/antropy} (accessed 12.09.23; Version: 0.1.4)}.

\paragraph{CID Complexity Estimation}
\citep{batista2014cid} introduced a \ce for their Complexity-Invariant Distance (CID) metric. The CE measures the complexity of a time series by stretching it and comparing the length of the line, where a simple line would be the least complex case. It has no parameters and only two requirements, that the time series are of the same length and are Z-normalized. Those properties make this complexity metric easy to use and easy to understand. \citep{batista2014cid} suppose that CE is an estimation of the intrinsic dimension and shows experiments which suggest that it is a property of the shape rather than the dimensionality.

\begin{equation}\mathit{CE}(X) = \sqrt{\sum\limits_{i=1}^{N-1} (q_i - q_{i+1})^2} \end{equation}
\\
where $N$ denotes the length of the time series $X$, and $q_i\in X$ its $i$-th element.

\paragraph{Singular Value Decomposition Entropy}
Our second complexity measure is the \svden which measures the Shannon Entropy for the vector components which can construct the dataset \cite{li2008analysis} and is defined as following: 

\begin{equation}
    \text{SvdEn} = - \sum\limits^{k}_{i=1} \overline{\lambda}_i \ln( \overline{\lambda}_i)
\end{equation}
\begin{equation}
    \overline{\lambda}_i = \frac{\lambda_i}{ \sum\limits^{k}_{j=1} \lambda_j}
\end{equation}

where $\sum \lambda_k = 1$ and $\lambda_k$ is a singular value of the matrix $M$, which indicates the signal complexity. $M$ is a $n \times m$ matrix generated by sampling vectors in the form of sliding window subsequences of length $m$. Afterwards it is decomposed into $M = USV^T$, where $U$ is a $m\times k$ matrix, $V$ a $n\times k$ matrix and $S$ a diagonal $k\times k$ matrix, which consists of the singular values in form of a diagonal vector $S = \text{diag}(\lambda_1 \lambda_2, ..., \lambda_k)$ \cite{li2008analysis,broomhead1986extracting}. In other words, SvdEn shows the deviation of the singular values to the uniform distribution and thus a higher value indicates a higher complexity. Compared to more spectrum-derived entropy estimators, SvdEn is less effected by exogenous oscillations \cite{sabatini2000analysis}.

\paragraph{Approximate Entropy}
\apen \cite{pincus1991approximate} measures the randomness and similarity of patterns inside a time series, by dividing it into blocks of length $m$ and measures the logarithmic frequency of the closeness and the repeatability of found patterns \cite{delgado2019approximate}. \cite{delgado2019approximate} summarizes ApEn as follows: ``ApEn is a parameter that measures correlation, persistence or regularity in the sense that low ApEn values reflect that the system is very persistent, repetitive and predictive, with apparent patterns that repeat themselves throughout of the series, while high values mean independence between the data, a low number of repeated patterns and randomness''. The statistical approximation is calculated with:
\begin{equation}
    \text{ApEn}(m, r, N) = \phi^m(r) - \phi^{m+1}(r)
\end{equation}
where $m$ is the length of the compared subsequences, $r$ is a tolerance filter to remove some small noise and $N$ is the length of the time series. 
\begin{equation}
\small
    \phi^m(r) = \frac{1}{N - m + 1} \sum\limits^{N-m+1}_{i=1} \log C^m_i(r)
\end{equation}
\begin{equation}
\small
    C^m_i(r) = \frac{1}{N-m+1} \sum\limits^{N-m+1}_{i=j} \delta^m_{i,j}(r)
\end{equation}
\begin{equation}
\small
\delta^m_{i,j}(r) = [\,d[\,\abs{x_m(j) - x_m(i)}\,] < r]
\end{equation}

where $x_m(i)$ is a vector of length $m$ starting at $i$ and $d$ is the maximal distance of the scalar components. However, some problems of ApEn are that the relative consistency is not guaranteed.  ApEn tends to overestimate the regularity. Further, to ensure that the logarithms stays finite, a self-counting component is included, but which introduces a bias for especially smaller time series \cite{delgado2019approximate}. 

\paragraph{Sample Entropy}
The \sampen \cite{richman2000physiological} is an alternative measure to the ApEn, which also calculates the randomness and similarity of patterns, while solving the three named problems of ApEn \cite{delgado2019approximate}. By not including the self-counting bias, it is mostly independent of the input length \cite{delgado2019approximate}. For this reasons, we consider both measurements. However, SampEn needs to find matching patterns to be defined and is further less reliable for small numbers of patterns, \eg~ \cite{richman2000physiological,delgado2019approximate}. The smaller SampEn is, the less complex the system is. The statistical approximation of SampEn is calculated as follows:

\begin{equation}
    \text{SampEn}(m, r, N) = - \ln[A^m(r) / B^m(r)]
\end{equation}
where again $m$ is the length of the compared subsequences, $r$ is a tolerance filter to remove some small noise and $N$ is the length of the time series.
\begin{equation}
\small
    B^m(r) = \frac{1}{N - m}\sum\limits^{N-m}_{i=1} B^m_i(r)
\end{equation}
\begin{equation}
\small
    B^m_i(r) = \frac{1}{N - m - 1}\sum\limits^{N-m}_{j=1;j\not=i} \delta^m_{i,j}(r)
\end{equation}
\begin{equation}
\small
    A^m(r) = \frac{1}{N - m}\sum\limits^{N-m}_{i=1} A^m_i(r)
\end{equation}
\begin{equation}
\small
    A^m_i(r) = \frac{1}{N - m - 1}\sum\limits^{N-m}_{j=1;j\not=i} \delta^{m+1}_{i,j}(r)
\end{equation}

\paragraph{Trend Shifts}
We introduced Trend Shifts (T. Shifts) in \cite{SA:21:local} to evaluate the complexity of a time series. The idea is that the less often the trend shifts (as in how often the slope changes), the less complex a time series is. The simplest case would be a line with 0 trend shifts. This makes it an easy-to-understand metric, which is somewhat similar to the CE, but at the same time focuses on different aspects of the complexity. As the underlying assumption is relatively simple, we will compare its correlation to other abstraction measures. We define it as:

\begin{equation}
    slop(x_{i-1}, x_{i}, steps) = \frac{x_i - x_{i-1}}{steps}
\end{equation}
\begin{equation}
    \text{Trend Shifts}(X) = \sum\limits_{i=2}^N \lvert (slop(x_{i-1}, x_{i}, 1) - slop(x_{i-2}, x_{i-1}, 1))\rvert \geq r
\end{equation}
where $x_i$ is the $i$-th element in $X$, $N$ the length of the time series and $r$ a regulation variable to not allow for too small direction changes, \eg in our case $r=0.001$.

\paragraph{Data Reduction}

In \cite{SA:21:local} we defined a precentral data reduction, \ie the number of data points that gets removed on average from a data sequence. We found that interpolating the removed data made the most sense, for interpretability and performance. In \cite{schwenke2021abstracting} we further found an average Pearson correlation of $r= -0.9528 \pm 0.0282$ between the data reduction to the complexity measures SampEn, ApEn, SvdEn and the Trend Shift. We define the data reduction as:

\begin{equation}
    \text{Data Reduction} = \frac{N - u}{N}
\end{equation}
domain
where $N$ is the length of the time series and $u$ the number of removed elements in this sequence. Later in section \ref{sec:metricDef} we discuss and formalize a few measurements based on some of those definitions, to better explore the quality of our explanations.

\subsubsection{XAI Metrics and Measurements}
\label{sec:metrics}
When defining an important explainability or interpretability metric, one big challenge is how to actually formalize them, because interpretability can be a quite relative concept \cite{carvalho2019machine, ruping2006learning}. Often, many concepts are rather domain or model specific, which makes it even harder to compare them \cite{ruping2006learning, robnik2018perturbation, carvalho2019machine}. Therefore, \citep{rudin2018please} argued that there can not be an all-purpose formalization for interpretability. Even if one only focuses on the time series domain, no interpretation metric exists to guarantee the right behaviour of a model \cite{rojat2021explainable}. \citep{carvalho2019machine} summarized a few quality properties of explanation methods from \citep{robnik2018perturbation}, on which we want to focus. However, because they are only verbally defined, it is unclear how to actually calculate, how to best apply them and in which use cases they are the most useful \cite{carvalho2019machine}. Here, we formalize some XAI metrics based on those quality measurements. It is important to keep in mind, that we focus in this work on the increase of the understandability and interpretability of the MHA, which is only a part of the Transformer model, but which therefore opens up multiple challenges, when comparing the performance of the model and the interpretation.

\paragraph{Fidelity:} Based on the description from \cite{carvalho2019machine}, we define fidelity to the original model as: 
\begin{equation}
\label{eq:modelFidelity}
F_m(M,I,D) = \frac{1}{l_D} \sum\limits_{X\in D} ( eq(M(X), I(X)))
\end{equation}

with $M$ as a function representing the original model and $I$ for the interpretation model, $D$ as input dataset, $l_D$ for the number of elements in $D$ and $eq(x,y)$ a binary function which is 1 if $x$ equals $y$, and 0 otherwise. A high $F_m$ is desired, so the interpretation model is as close as possible to the original model. Consequently, if the original model has a low accuracy, the interpretation model should also have a low one.

A different kind of fidelity -- because it is always the question of fidelity to what -- for salient features is defined by \citep{pope2019explainability} as: ``The difference in accuracy obtained by occluding all nodes with saliency value greater than 0.01 (on a scale 0 to 1)'', \ie checking if informative features are included in the less important features. However, because we are in the domain of time series data, often data points have shared information and thus removing this redundancy can lead to a misleading salient feature fidelity. Therefore, we do not look into this fidelity, but into the salient feature infidelity, where all low salient values are removed, to see if the important features for decision marking are still included. Because we use neural networks, we approximate a more accurate score by use the N-fold cross validation. However, this is exactly the approach we take to validate our \lasa approach from Section \ref{sec:lasa}, which is why we can just take the accuracy difference between a \lasa model and the original model, to receive the salient feature infidelity.

\paragraph{Consistency:} Neural networks themselves are often not consistent per run, which makes it hard to test the consistency of the interpretation. Because we mostly focus on the MHA and our method uses deterministic steps, we aim to measure the consistency of the Attention matrices inside the MHA. In Section \ref{sec:LAAM} we define a so-called \laam, which aggregates all Attention matrices of one trial into one matrix. Between two \laams we define the matrix euclidean distance $md$ as:

\begin{equation}
md(A, B): \sqrt{\sum\limits_{i,j \in \text{indices}(A)} (A_{i,j} - B_{i,j})^2}
\end{equation}

where $A$ and $B$ are two \laams of the same shape, and \textit{indices($A$)} provides a set of all possible index combinations of a two-dimensional matrix. To estimate some kind of relative consistency, we sampled three different test samples per class per dataset and calculated for each dataset three different distances:
\begin{itemize}
    \item \textit{OuterDistance}: First we have the average distance of a test sample $k$ in fold $f$, to the sample $k$ in every other fold, \ie the average distance of one sample between models.
    \item \textit{InnerFoldDistance}: Second, we have the average distance between a test sample $k$ of class $c_1$ in fold $f$, and other samples of another class in fold $f$, \ie the average distance between different classes in one model.
    \item \textit{InnerClassDistance}: Last but not least, we have the average distance between a test sample $k$ of class $c_1$ in fold $f$ to all other test samples of the same class $c_1$ in the same fold $f$, \ie the average distance between class members.
\end{itemize}

With those three measurements, we can estimate if we have a strong \laam inconsistency between different models. Due to the inconsistency of the neural networks, we already expect some general inconsistency between the \laams for the same input from different models. If the \textit{OuterDistance} is close to or higher than the \textit{InnerFoldDistance}, the \laam would be questionably inconsistent between models -- in consideration of what the $md$ measures, \ie the $md$ is quite simple and \eg does not measure the relative consistency between values in one \laam. However, if the \textit{OuterDistance} is lower or close to the \textit{InnerClassDistance}, we can talk about some form of inner class consistency, where the different \laams per class group in a similar area.

\paragraph{Stability:} Comparing the distance between \laams from different inputs is tricky, because for a different input at position $x$, different Attention values can arise -- which are additionally dependent on the whole input sequence --, thus it is hard to compare two different input values, when it is unclear how important they are for the classification task. However, while we evaluate the consistency, we can also compare the \textit{InnerFoldDistance} and the \textit{InnerClassDistance} to grasp the typical distance between classes, indicating the stability of \laams to some extent. On the other hand, our Global Coherence Representation (\gcr), which is a global construct, considers the Attention per symbol and position in deterministic steps and is thus always stable in the classification per constructed \gcr.

\paragraph{Certainty:} While \lasa (see Section \ref{sec:lasa}) is just an abstraction, the \gcr has a certainty measurement included, based on the ``best'' known class representation. We will analyse it with the train and test data, by checking the overall accuracy when removing all certainty values under a specific percentage based threshold. A good certainty would mean, that if one analysed the less certain options, one could examine in which samples and why the model struggles. When only looking at the ``certain'' (based on a relative threshold) trials, the interpretation should ideally achieve an accuracy of 100\% or if any wrongly classified trials are left, they should indicate grave errors in the model. If this is not the case, the quality of the certainty should be questioned. In case the certainty is not given by the model itself but an interpretation (as it is the case for the \gcr), the model fidelity is also an important factor to consider while interpreting the certainty. For the \gcr this could mean that the Attention values are suboptimal.

To analyse this property, we look into the most certain samples in the steps of $80\%$, $50\%$, $20\%$ and $10\%$ of the whole train and test set. A very good indication in favour of the certainty is if the accuracy rises for each step. Optimally this is always the case, but overall it is already good if the average accuracy per step rises for the selected certainty.

\paragraph{Importance:} As \citep{baan2019understanding} have shown, Attention is not equal to importance, but nevertheless Attention is somewhat interpretable and the Transformer model can handle multiple problems better due to the MHA; hence, we showed in \citep{SA:21:local} that it is possible to abstract local time series with Attention. Further, because Attention is suspected to highlight regardless of the output class, we want to reduce this local noise of unwanted classes in Attention by constructing the \gcr, aiming to receive more class dependent importances. Therefore, the better we approximate importance with Attention, the closer Attention should be to a good certainty score.

%% file: s4_methods.tex
\section{Methods}
\label{sec:methods}
\begin{figure}[ht!]
	\centering
	\includegraphics[width=0.98\columnwidth]{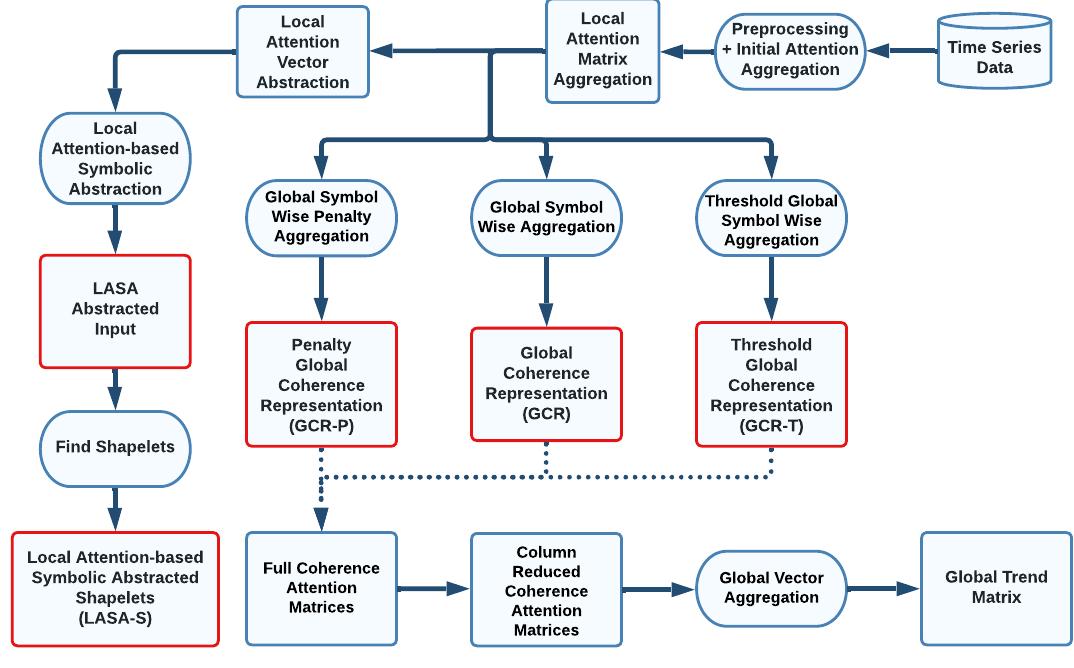}
	\caption{Overview of the relation between the different algorithms and representations and how they are connected. The round boxes represent methods and the edgy boxes stand for resulting data representations. The red border highlights the five different model types we look into.}
	\label{fig:overviewAll}
\end{figure}

In this section, we present the different methods proposed in \cite{SA:21:global,SA:21:local}, but formalize them in more depth and additionally introduce some further variations/extensions of those methods. An overview on how each method and representation relate can be found in Figure \ref{fig:overviewAll}. In the following, we will explain each method and resulting representation in more detail. The red borders in Figure \ref{fig:overviewAll} mark the five different types of models we look into; it is important to note that the global models have three different representation models. Therefore, in this work we look into eleven different types of models. The proposed aggregation methods are only a proposition and can be improved or interchanged easily, therefore our methods can be seen as a framework to create our proposed local and global representation with Attention. 

\subsection{Focus of Methods}

In this section, we want to clarify the different possible applications for each method. XAI is in general a quite complex topic with multiple open challenges. Due to the specialization of multiple methods, different types of interpretation approaches exist, each with different effects and ways of improving the interpretability. In our case, as can be seen in Figure \ref{fig:overviewAll}, we have two base methods, the \lasa and the \gcr approach. We look into different aggregation methods, using only the different Attention matrices inside the Model. This could be seen as a composition method \cite{li2021interpretable}, where the internal algorithm is still incomplete.

The \lasa approach is an abstraction approach, in the sense, that we remove all less attended information to see if it is possible to classify the current given problem. From a XAI perspective, we take learned information out of the model to abstract the input data and thus simplify the data to promote accessibility and understandability of the learned problem. From the model perspective, \lasa tells us which elements and hence information the model is focusing on, at least when excluding all other model weights; \ie we verify with an additional trained model that the highly attended elements in the internal state of the Transformer model (aggregated Attention) still contain enough information to classify the given problem, \cf \textit{RemOve And Retrain} (ROAR) \cite{hooker2019benchmark}.

The \gcr approach has two possible applications. It can be used for classification based on weighted occurrence in a symbol-to-symbol fashion; \ie we have a classification model that is fully interpretable, nicely visualisable and can classify a dataset based on from-to relations constructed from strength/importance assignments. We use the aggregated Attention weights to train multiple \gcr variations and verify if we can classify using this information. In other words, we train a new model with the internal information/state of a Transformer model, to verify that the Attention values provide well formatted information to solve the given classification task. The second option is not to approximate the solution for the classification problem, but use the \gcr to approximate the trained Transformer model, where the \gcr is acting as an interpretable proxy model \cite{li2021interpretable}. That means, the \gcr aims to approximate the missing link between Attention and the model output. With those two approaches and by only using Attention information, we create an interpretable and visualisable model to approximate a classification task via importance scores, as well as an interpretation method to approximate a better understanding of a given Transformer model.

\subsection{Initial Processing}

Each method we present here uses the in this section defined pre-processing pipeline before training the model. From the final trained model, Attention matrices can be extracted and aggregated into a \laam. Those two steps represent the initial process on which every representation is build.

\subsubsection{Pre-Processing}
\label{sec:prepro}

All input data is scaled to unit variance with the Sklearn \cite{sklearn_api} standard-scaler. Afterwards, the data is transformed into symbols using the SAX algorithm using a uniform distribution 
to calculate the symbol separations, which abstract the data into a discrete data format; thus already increasing the accessibility of the input data. Both steps are only fitted on the training data to exclude any test biases.
We look into a variety of number of symbols $S$ to analyse its impact, but focus on rather small numbers to further facilitate the interpretability. It is to note that by abstracting data into symbols, information is getting lost, thus selecting the right $S$ for the given classification task is crucial. By having symbolized data, we shift our application context more into the NLP context, which also opens up the possibility to use methods from NLP. Normally, an input embedding is trained for each symbol to improve the model's performance. However, here we use a simple evenly distributed mapping of the symbols to the interval [-1,1] according to the value relation of the symbols; \cf y-axis in Figure \ref{fig:sax}. In \citep{SA:21:local} we suggested using the simple mapping to keep the value relation information, rather than to approximate it, \eg the symbol-set \{$a$,$b$,$c$\} would be mapped to \{-1,0,1\}; \ie $a =$ ``largely negative values'', $b =$ ``values close to zero' and $c = $ ``largely positive values''.

\subsubsection{Initial Attention Aggregation}
\label{sec:LAAM}

For each symbolic input, we can extract $l \times h$ Attention matrices (also later referred to as local Attention matrices), where $h$ is the number of heads and $l$ the number of encoder layers inside the Transformer model. As other work has already shown \cite{pruthi2019learning,baan2019understanding,bhojanapalli2021leveraging}, Attention matrices include a lot of redundancy. Therefore, we aggregate all Attention matrices into one \acf{LAMA} on which all further calculations are based on.

\begin{definition}[Local Attention Matrix Aggregation]
The \acf{LAMA} is the matrix received after aggregating all or a subset of the Attention matrices inside a Transformer model in any specific way, with the goal to promote interpretability.
\end{definition}

Our proposed methods consist of two functions $f_{s1}(X)$ and $f_{s2}(X)$, with the goal to reduce the Attention heads and afterwards the encoder layers or vice versa; by using the $max$ or $sum$ operation. We hence define:

\begin{equation}
    A^{\mathit{L_{AMA}}} = f_{s2}(f_{s1}(T_A(M(X_{in}))))
\end{equation}

where $T_A$ is a function that extracts all emerging $l \times h$ many Attention matrices of size $n\times n$ from the Transformer model $M(X_{in})$, for a sequential input $X_{in}$. $T_A(M(X_{in})))$ therefore is a 4 dimensional matrix with shape $(l, h, n, n)$, which gets mapped to a matrix $A^{\mathit{L_{AMA}}}$ of shape $n\times n$:

$$T_A(M(X_{in})) = \begin{bmatrix}
A_1^1 & A_2^1 & \cdots & A_h^1\\
A_1^2 & A_2^2 & \cdots & A_h^2\\
\vdots & \vdots & \ddots  & \vdots\\
A_1^l & A_2^l & \cdots & A_h^l
\end{bmatrix}$$
with $l \times h$ many Attention matrices $A$ of size $n \times n$. $f_{s1}$ takes the 4 dimensional input and returns $h$ or $l$ (depending on the $f_{s1}$ variation) many $n\times n$ Attention matrices, effectively reducing one dimension. In our analysis, we look into four variations of $f_{s1}$, namely:

\begin{equation}
    f_{s1}^{lm}(T_A(M(X_{in}))) = \begin{bmatrix}
f^{max}(A^1_1, A^2_1, \dots, A^l_1) \\
f^{max}(A^1_2, A^2_2, \dots, A^l_2)\\
\vdots\\
f^{max}(A^1_h, A^2_h, \dots, A^l_h)
\end{bmatrix}
\end{equation}

\begin{equation}
    f_{s1}^{hm}(T_A(M(X_{in}))) = \begin{bmatrix}
f^{max}(A^1_1, A^1_2, \dots, A^1_h)\\
f^{max}(A^2_1, A^2_2, \dots, A^2_h)\\
\vdots\\
f^{max}(A^l_1, A^l_2, \dots, A^l_h )
\end{bmatrix}
\end{equation}

\begin{equation}
    f_{s1}^{lm}(T_A(M(X_{in}))) = \begin{bmatrix}
f^{sum}(A^1_1, A^2_1, \dots, A^l_1) \\
f^{sum}(A^1_2, A^2_2, \dots, A^l_2)\\
\vdots\\
f^{sum}(A^1_h, A^2_h, \dots, A^l_h)
\end{bmatrix}
\end{equation}

\begin{equation}
    f_{s1}^{hm}(T_A(M(X_{in}))) = \begin{bmatrix}
f^{sum}(A^1_1, A^1_2, \dots, A^1_h)\\
f^{sum}(A^2_1, A^2_2, \dots, A^2_h)\\
\vdots\\
f^{sum}(A^l_1, A^l_2, \dots, A^l_h )
\end{bmatrix}
\end{equation}

with functions:

\begin{equation}
    f^{max}(X) = \{\max\limits_{A \in X}(A_{ij}) \,\vert\, i,j = 1,\ldots, n\}
\end{equation}

\begin{equation}
    f^{sum}(X) = \{\sum\limits_{A \in X}(A_{ij}) \,\vert\, i,j = 1,\ldots, n\}
\end{equation}

with $A_{ij}$ being the element at position $i,j$ from the $n \times n$ attention matrix $A$. Note the order of the indices $l$ and $h$ in each function, indicating which dimension gets reduced. For $f_{s2}$ we look into two variations:

\begin{equation}
    f_{s2}^{m}(X) = f^{max}(X)
\end{equation}
\begin{equation}
    f_{s2}^{s}(X) = f^{sum}(X)
\end{equation}

\begin{figure}[ht!]
	\centering
	\includegraphics[width=0.98\columnwidth]{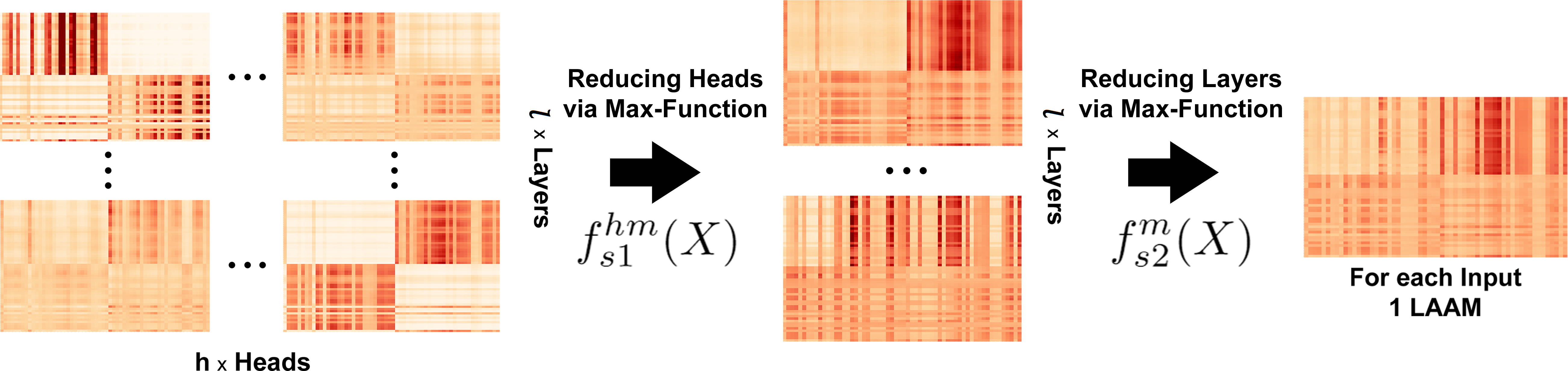}
	\caption{Example \laam aggregation process how n layers and h heads are reduced to one \laam using hl-max-max.}
	\label{fig:LAAMAggregation}
\end{figure}

In other words, we take the maximum or sum over the index of the heads or encoder layer in $f_{s1}$ and the maximum or sum over the index that is left (heads or encoder layer), resulting in eight variations to calculate the \laam. The average was not considered, due to the fact that the relative scale of the data is the same as for the sum, because for each matrix index the number of data points is the same. To differentiate the variations we introduce the following abbreviations: ``l'' stands for \emph{layer}, while ``h'' stands for the \emph{heads} and hence ``lh'' would denote the collapse of first the layers, and second the heads. Attached and separated by a ``-'' are the collapsing method(s), \ie \emph{max} or \emph{sum}. For example, ``hl-max-sum'' (abbreviated as ``hl-ms'') would be equivalent to $f_{s2}^s(f_{s1}^{hm}(T_A(M(X_{in}))))$. One example visualisation for hl-max-max can be seen in Figure \ref{fig:LAAMAggregation}. Important to note is that in our evaluations we only look into the $hl$ reduction, because it performed overall better in \cite{schwenke2021abstracting} and else we would have some time constraint problems with all our possible configurations, see Section \ref{sec:config}; \ie we look only in four out of eight possible configurations.\\

Multiple work on models with higher layer counts suggest to only use the last few layers for interpretation \cite{wang2020language,clark2019does,ramsauer2020hopfield}. This can be taken into account when constructing the \laam but we aggregate over all layers due to comparability and the rather low layer count we test on; but which is sufficient for most of the tasks we take on, as also indicated by \cite{wen2022transformers}. Further, we are aware that our chosen Attention aggregation methods are rather simple and do not justify the complex calculations of the follow-up layers of the model, but nevertheless we showed in \citep{SA:21:local,SA:21:global,schwenke2021abstracting} that even with simple aggregations, some value can be extracted from Attention. Therefore, our approaches can be seen as a framework, where the aggregation and of the Attention is interchangeable.

\subsection{Local Attention-based Symbolic Abstraction}
\label{sec:lasa}
Our first Method is \acf{LASA} \cite{SA:21:local} with the goal to use SAX and Attention as means to abstract the input data, \ie improving the accessibility of the data by reducing the overall data complexity. We do this by removing less attended symbolized data points -- with the assumption they are less important \cite{pruthi2019learning} -- of a singular (local) input to receive a local abstract input. We define \lasa as:

\begin{definition}[Local Attention-based Symbolic Abstraction]
\lasa is a process to abstract local sequential data by using a symbolic abstraction to reduce the value space complexity and using a human-in-the-loop focused treshold-based approach to reduce the amount of data points of interest inside the input sequence; based on the information provided by a \laam.
\end{definition}

In the following, we describe the process pipeline of \lasa, followed by explaining the analysed pipeline steps to finer define our method.

\subsubsection{Pipeline}

\begin{figure}[ht!]
	\centering
	\includegraphics[width=0.85\columnwidth]{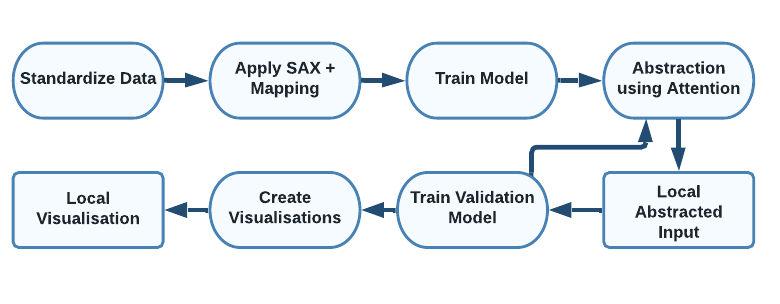}
	\caption{The process pipeline of our Local Attention-based Symbolic Abstraction approach.}
	\label{fig:localpipeline}
\end{figure}

Figure \ref{fig:localpipeline} shows the pipeline of our \lasa approach. The first two steps are about pre-processing the data as described in section \ref{sec:prepro}. The symbolized data is used to train a Transformer model, also to verify that the task is still trainable. Afterwards, one variation of the process in section \ref{sec:LAAM} is executed to extract a \laam out of the trained model -- for one given single input. Using a \laam and a set of thresholds, the abstraction steps in section \ref{sec:localAbstractionProcess} are executed. This is done for all the data, to receive a set of simpler data. This data can now be visualized, to verify the abstraction and improve the understanding of the underlying problem. To formally verify the process, we train another model, as described in section \ref{sec:liaVeri}. To explore and improve the abstraction further, a human-in-the-loop can now fine-tune the thresholds and repeat the presented process.

\subsubsection{Abstraction Steps}
\label{sec:localAbstractionProcess}

All abstractions are based on the \laam which is based on the to be abstracted local input. In the following, we propose one abstraction process, which is however interchangeable. The first step is to further reduce the \laam into a vector -- we name it \acf{LAVA} -- by using one out of two variations of function $f_{s3}(X)$:
\begin{equation}
    A^{\mathit{L_{AVA}}} = f_{s3}(f_{s2}(f_{s1}(T_A(M(X_{in})))))
\end{equation}
\begin{equation}
    f_{s3}^{m}(X) = \{\max\limits_{j=1}^n(x_{i,j}) \vert x_{i,j} \in X; i = 1,\ldots,n\}
\end{equation}
\begin{equation}
    f_{s3}^{s}(X) =\{\sum\limits_{j=1}^n(x_{i,j}) \vert x_{i,j} \in X; i = 1,\ldots,n\}
\end{equation}

and thus we introduce an extension to our abbreviation with an additional parameter, \eg `hl-max-sum-max'' (abbreviated as ``hl-msm'') would stand for using $f_{s3}^m(f_{s2}^s(f_{s1}^{hm}(T_A(M(X_{in})))))$. 

\begin{figure}[ht!]
    \centering
    \includegraphics{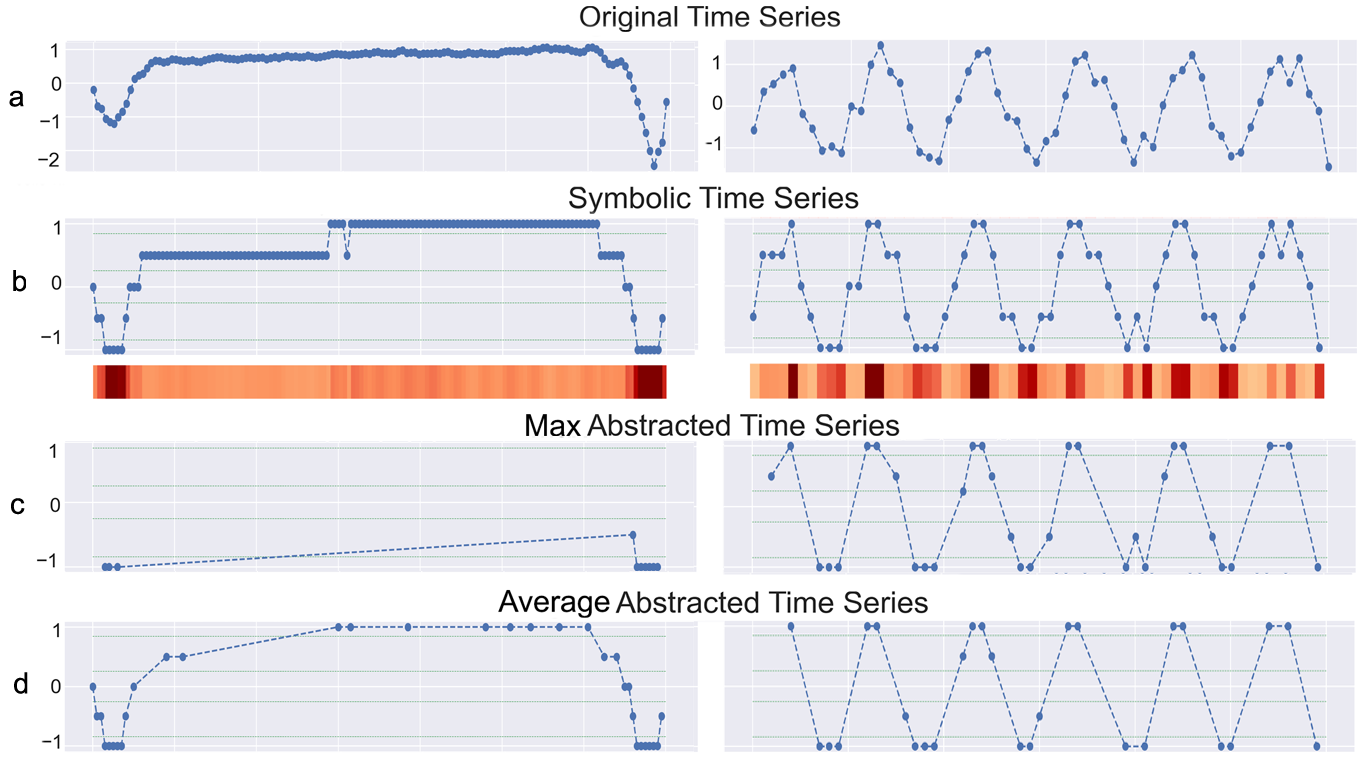}
    \caption{Examples how the \lasa process abstracts the original data into four less complex inputs, from two different datasets. It also highlights the effect of two different kinds of thresholds.}
    \label{fig:lasaExample}
\end{figure}

Often, only one cut-off is used to categorize the Attention weights into important or not important \cite{jain2019Attention, serrano2019Attention}. However, medium strong important data points can also hold partial information. Therefore, we introduce two thresholds $t_1$ and $t_2$ to differentiate between high, medium and low Attention values. We map all high attended values with Attention value $a > t_1$ as is into the abstraction. A sequence of medium strong attended values with Attention value $t_1 > a > t_2$ will be mapped to the centre position of this subsequence, with the median of the sequence as value. All values with an Attention value $a < t_2$ will be dropped. How to find good values for $t_1$ and $t_2$ is still unclear, and each problem needs its own optimization, thus a human-in-the-loop is needed to fine-tune the thresholds. In \citep{SA:21:local,schwenke2021abstracting}, we suggested two general guideline threshold sets, to either opt for keeping more accuracy or having a higher data reduction:

\begin{enumerate}
    \item \textbf{Average-based Thresholds (abt):} We select the average Attention value of the $A^{\mathit{L_{AVA}}}:$ $$\overline{A^{\mathit{L_{AVA}}}} = \frac{1}{n} \sum\limits^l_{i=1} A^{\mathit{L_{AVA}}}_i\,,$$ where $A^{\mathit{L_{AVA}}}_i$ is the $i$-th element inside the current \laav and $n$ is the length of the \laav vector. With this we define $$t_1^A = \frac{\overline{A^{\mathit{L_{AVA}}}} }{s_1}, t_2^A = \frac{\overline{A^{\mathit{L_{AVA}}}} }{s_2}\,,$$ where $s_1$ and $s_2$ are fine-tuning variables that explored in a human-in-the-loop approach. We analysed several parameters, \cf Section~\ref{sec:config}.
    \item \textbf{Maximum-based Thresholds (mbt):} With $$t_1^M = \frac{1}{s_1} max(\overline{A^{\mathit{L_{AVA}}}})\,, t_2^M = \frac{1}{s_2} max(\overline{A^{\mathit{L_{AVA}}}})\,.$$ we define maximum-based thresholds, again with parameters $s_1$ and $s_2$.
\end{enumerate}
The potential effects of the two different threshold are shown in Figure \ref{fig:lasaExample}, where four different \lasa abstractions can be seen from two different datasets.

In \citep{schwenke2021abstracting} we already looked into the effects of different aggregations on the accuracy and data abstraction, but for verification we add tests on more datasets and with more hyperparameters in this paper, in order to fine-tune the results and learn more about the effects of different parameters.

\subsubsection{Abstraction Validation}
\label{sec:liaVeri}
Due to the facts, that the leverage of inputs to outputs in neural networks are not fully understood and additionally that in this work we focus on the analysis of the Attention matrices of the MHA -- without considering the other layers of the Transformer model -- we focus on the interpretability of the trained classification task; by using a second model to verify that the abstraction still contains all necessary information for the trainings task, rather than masking inputs to the original model. This approach is also called ROAR \cite{hooker2019benchmark}. In \citep{SA:21:local} we tried to mask or to interpolate the removed data. Interpolating the data always provided the better results, which is why we only look into interpolating missing data for the verification model. All non-interpolatable data is masked, \ie the start and end of the input.

\subsubsection{Local Attention-based Symbolic Abstracted Shapelets}
\label{lasa-s}
Shapelets -- as we described in Section \ref{sec:shapelets} -- are one popular method to find shapes to maximize the distinguishability between classes. Because of their success, we want to use them to evaluate our approach. To do this, we calculate the Shapelets using Sktime \cite{sktime} on the original data and calculate their accuracy. Afterwards, we calculate the Shapelets on the abstracted data -- we call them \acf{LASA-S}. If the abstraction still contains all important classification information, the accuracy of the \lasas should be similar to the basic Shapelets. Furthermore, we calculate the complexity of the Shapelets and the \lasas to compare if our abstraction method also simplifies the found Shapelets, hence providing even more value to the interpretation.

\subsection{Global Coherence Representation}
\label{sec:gcr}
Compared to \lasa which focuses on one local input, the \acf{GCR} is an approach to visualize and classify the underlying problem in a more global and class based fashion. This means we put together the knowledge of multiple local inputs and aggregate them together to reduce the effect of outliers; showing which values per class at which positions are often clustered and what their typical coherence (\eg Attention) is. 
\begin{definition}[Global Coherence Representation]
The Global Coherence Representation is a matrix-like class representation which shows the class affiliation for each symbolic input per position, in form of a symbol-to-symbol coherence -- \eg Attention -- between all symbols in a vocabulary $V$ at each possible position of fixed length; Thus it shows for each possible class $c$ how strong each symbolic input contributes to the current class in coherence to all other inputs. This also includes all other representations, which are based on such an assumption. 
\end{definition}

Shapelets for example show a shape, with the aim to maximize the representativeness of a class to all other classes. However, Shapelets are limited to a one-dimensional representation and need multiple shapes to construct value based conditions, \eg value $x$ is only low if value $y$ is high, else $x$ is high. Our proposed Global Coherence Representation \cite{SA:21:global} offers a two-dimensional view per class to show the symbol-to-symbol relations in more detail and thus highlights which symbols are typically present at which position and under which other symbol's relations for the given class. By representing the typical class, the \gcr can also be used for classification. Despite that, Shapelets are still more focused on highlighting subsequences inside the input data, where the \gcr in this work for now still focuses on the whole input sequence. In the following four sections, we will present our pipeline to build the \gcr and three different variations of the \gcr, which build upon each other and get simpler by each step.

\subsubsection{Pipeline}
\begin{figure}[ht!]
	\centering
	\includegraphics[width=0.85\columnwidth]{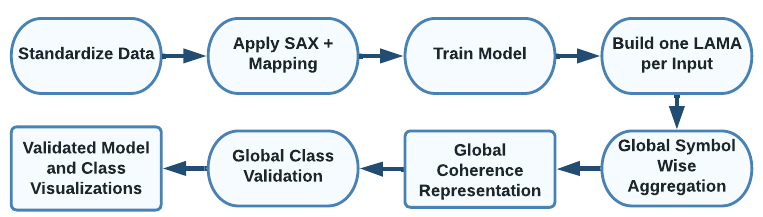}
	\caption{The process pipeline of our Global Coherence Representation.}
	\label{fig:globalpipeline}
\end{figure}

In Figure \ref{fig:globalpipeline} our pipeline for the \gcr can be seen. The first four steps are similar to our local approach, where we standardize the data, symbolify the input data, train a model based on the mapped symbol data and build \laams for each input. Afterwards we do a \acf{GSA} to aggregate the values inside the \laams based on their symbol, sequence position, and class into a \gcr. More precisely, we employ a \fcam per class, which will be described in the next section. The resulting \gcrs can now be visualised and validated on a test set, using the method described in section \ref{sec:globalVali}.

\subsubsection{Full Coherence Attention Matrices}

\begin{figure}[ht!]
	\centering
	\includegraphics[width=1\columnwidth]{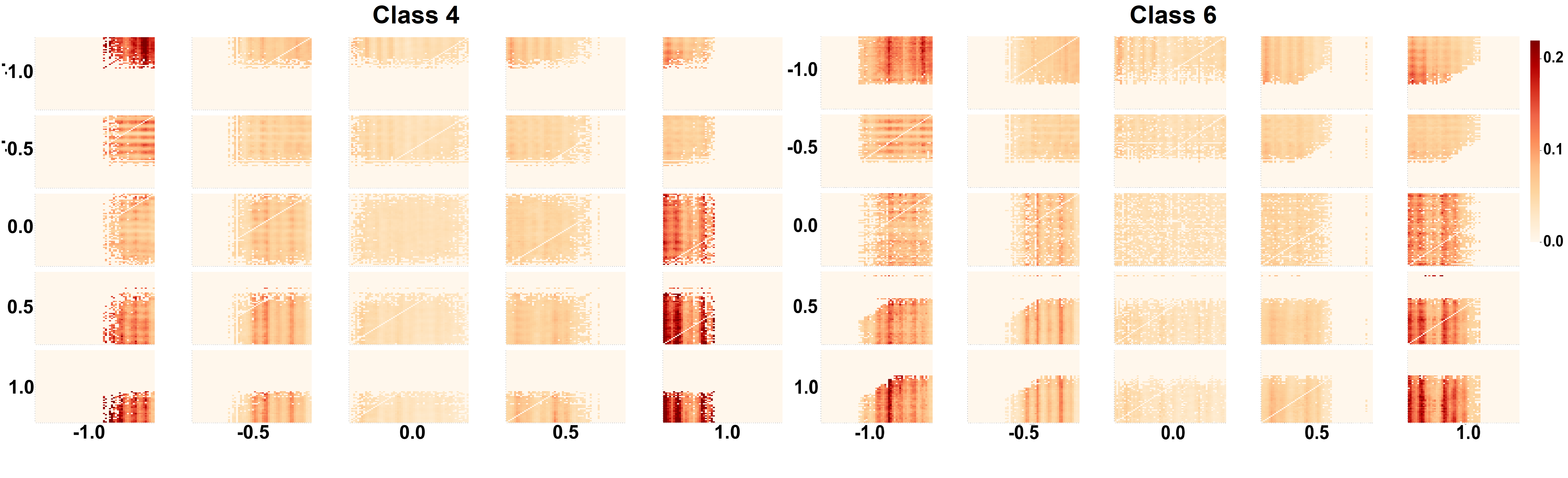}
	\caption{Two \fcams over the relative average \gswa and \textit{Max-Sum} \laam from the Synthetic dataset, w.r.t. class 4 (left) representing a slowly falling trend and class 6 (right) representing a sudden falling trend.}
	\label{fig:fullClassBoth}
\end{figure}

The first, most detailed \gcr is the \acf{FCAM}. Two examples for two different classes can be seen in Figure \ref{fig:fullClassBoth}. The \fcam contains a $v\times v$ matrix $F^c$ -- where $v$ is the size of the vocabulary and $c$ the class -- with each being a $n\times n$ matrix $P$, representing each symbol-to-symbol pair in the vocabulary $V$. Each matrix shows the aggregated Attention values for each two-dimensional position for this relation -- can be read like and works in the same sense as an Attention matrix from Section \ref{sec:encoderModel}. We hence define the $n\times n$ matrix $P^{ab-c} \in F^c$ as the matrix which coheres the relation between symbol $a\in V$ to symbol $b\in V$ in class $c$. To construct the \fcam we iterate through all \laams from the trainings dataset and put each aggregated Attention element $a_{ij} \in A^{\mathit{L_{AMA}}}_z$, which highlights symbol $a$ to symbol $b$, in the corresponding index position $P^{ab-c}_{ij}$, with $z$ being the iteration index over all training set \laams. 
We call this process \gswa and describe the principle for one out of two variants. For details on how to add one input to the global representation in more detail, we refer to Algorithm \ref{alg:fcam}. It is important to note that Algorithm \ref{alg:fcam} is executed for each \laam generated by each time series from the training set.

\begin{algorithm}
\small
\begin{algorithmic}[1]
    \Function{sum\_\gswa\_\fcam}{$F^{c},A^{\mathit{L_{AMA}}},X_{in}$}
        \For{$i = 0$ to len($A^{\mathit{L_{AMA}}}$)}
            \For{$j = 0$ to len($A^{\mathit{L_{AMA}}}[i]$)}
                \State $fromSymbol \gets X_{in}[i]$
                \State $toSymbol \gets X_{in}[j]$
                \State $P \gets F^{c} [fromSymbol][toSymbol]$
                \State $P[i][j] \gets P[i][j] + A^{\mathit{L_{AMA}}}[i][j]$
            \EndFor
        \EndFor
    \EndFunction
    \end{algorithmic}
    \caption{\small Global Symbol Wise Aggregation step to add one aggregated \laam $A^{\mathit{L_{AMA}}}$ from one specific symbol input sequence $X_{in}$ to a nested $v \times v$ dictionary map $F^{c}$ (representing the \fcam for class $c$) each including a $n \times n$ matrix for each possible sequence position. To complete the \fcam this method needs to be called for each input from the training set.}

    \label{alg:fcam}
\end{algorithm}

This \gswa variation will be referred to in the following as \emph{sum based \gswa}. This approach favours the statistical quantity of data per position and thus reduces the influence of the flat Attention values. The drawback is, however, that rare high attended positions can quickly become irrelevant. As alternative, we introduce the \emph{relative average (r. avg.) based \gswa}, which at the end additionally divides each position by the amount of data points added to it and hence this representation focuses more on the flat Attention values rather than the amount of data per index. This, however, can lead to cases, where an outlier's importance is exaggerated.

In \citep{SA:21:global} we showed that both \gswa variations can be favourable for different datasets. In the context of the huge number of runs, we further optimized the run time of our \gcr algorithms in general by parallelizing multiple operations, reducing time intensive operations and optimizing loop runs. The presented algorithms, later, are not the optimized versions, but rather aim to provide the general idea behind the algorithms. The functionality and the results of the implementations are not different, but more efficient. Regarding our results and implementation in \citep{SA:21:global}, we want to draw Attention to an implementation error in this publication, where only the outer symbols (\ie $1.0$ and $-1.0$) were considered for the classification and thus the results were only focused on the outer values. We will discuss this problem and the consequences of this later in Section \ref{sec:weaknesses}. 

\paragraph{Full Coherence Attention Matrices Example}

\begin{wrapfigure}{R}{0.58\columnwidth}
	\setlength{\abovecaptionskip}{5pt}
	\hfill
	\centering 
	\includegraphics[width=0.57\columnwidth]{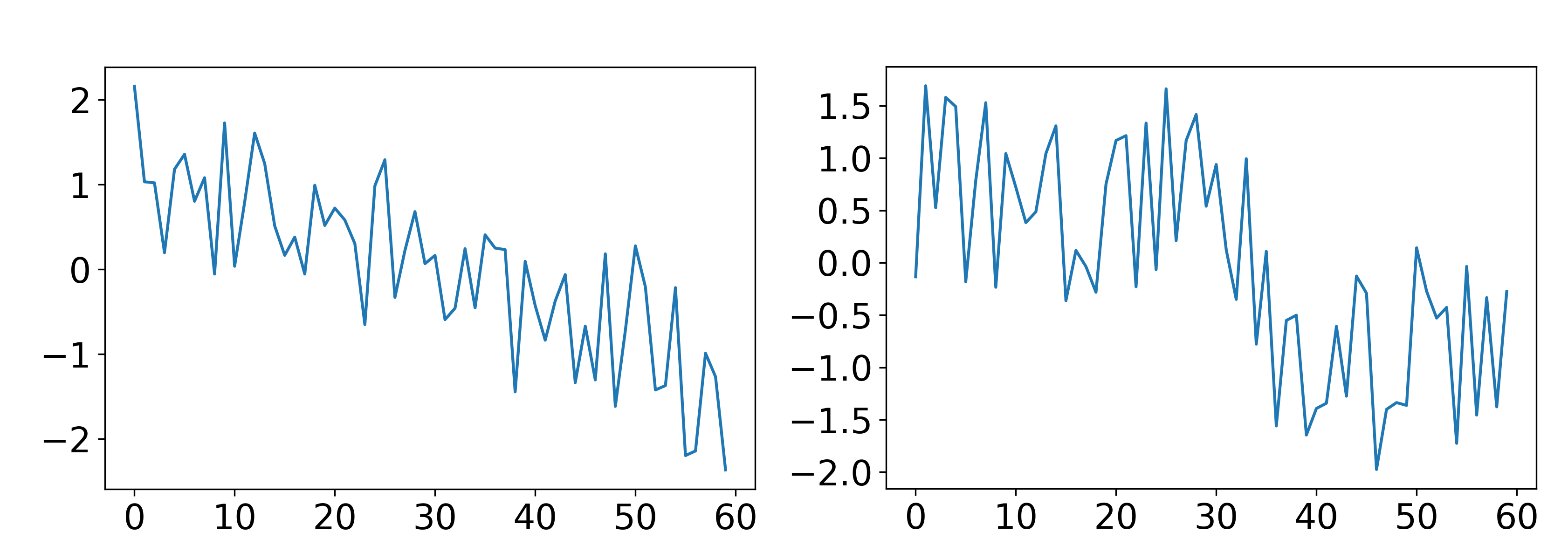}
	\caption{Example original time series for class 4 (left; slowly falling trend) and class 6 (right; sudden falling trend) from the Synthetic Control dataset.}
	\label{fig:gcrOriExample}
\end{wrapfigure}

Figure \ref{fig:fullClassBoth} (left) gives an illustrative example of two \fcams. At the top left matrix, one can see how the mapped symbol $-1.0$ highlights itself. In this matrix, the Attention is focused in the top right corner, showing that $-1.0$ symbols for this given class are only occurring at the end of the time series -- because the time series on both axes starts in the bottom left corner. On the other hand, the bottom right matrix shows how the mapped symbol $1.0$ highlights itself. Here, the Attention is focused in the bottom left corner, showing that $1.0$ symbols are typically at the beginning of the time series for this given class. All matrices in-between those two show a slowly drifting shift into the other corner, hence this \gcr's class shows a smooth trend fall. We can compare the left to the right \fcam from Figure \ref{fig:fullClassBoth} and see, that the Attention blocks of the right \fcam are bigger and thus indicate that the trend fall of the second class happens later and more sudden -- which is also the main difference between both classes. For a reference, two examples from each class can be seen in Figure \ref{fig:gcrOriExample}.

\subsubsection{Column Reduced Coherence Attention Matrices}
\begin{figure}[ht!]
	\centering
	\includegraphics[width=1\columnwidth]{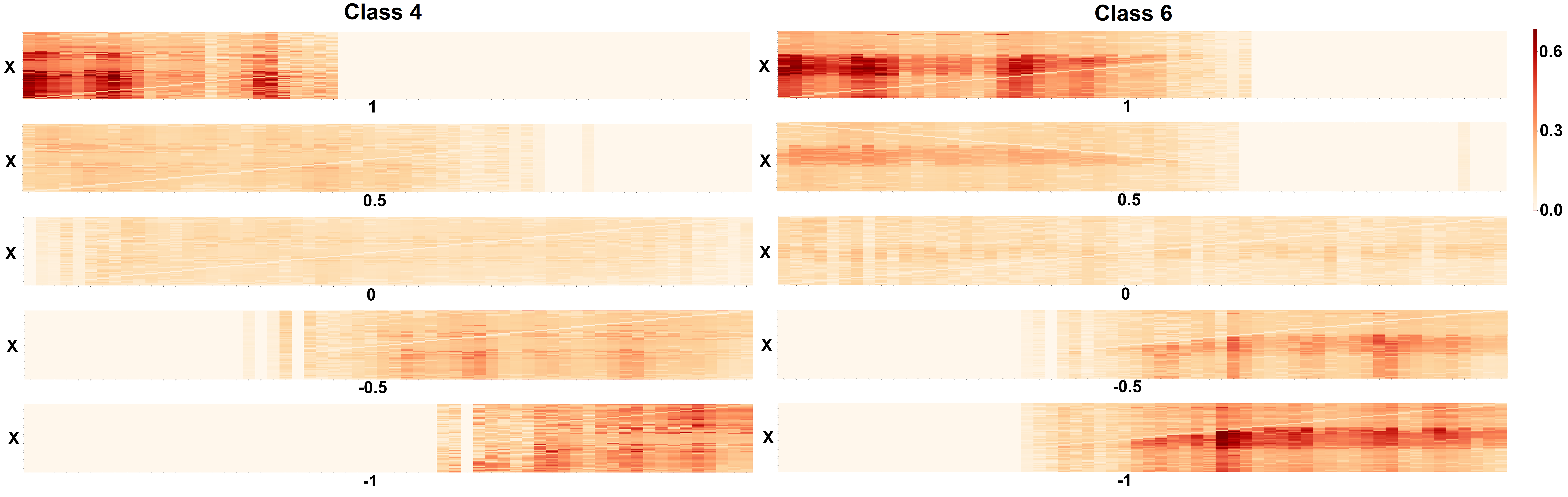}
	\caption{Two \crcams over the relative average \gswa and \textit{Max-Sum} \laam from the Synthetic dataset, w.r.t. class 4 (left) representing a slowly falling trend and class 6 (right) representing a sudden falling trend.}
	\label{fig:xClassBoth}
\end{figure}

The \gcr is quite detailed and can be hard to grasp. Therefore, we introduce a simplified variant, the \acf{CCAM}. Figure \ref{fig:xClassBoth} shows the same examples as in Figure \ref{fig:fullClassBoth} but as \crcam. It can be seen that the general class from the last example can now be better assessed. To construct the \crcam we take a \fcam and sum together all matrices in the \textit{from}-relation, thus resulting in $v$ matrices; showing how any symbol (in the figure represented as $x$) highlights on a specific symbol. Due to the fact, that we consider two variations for the \gswa to construct the \fcam, we have also two variations of the \crcam; one is based on the sum and the other on the relative average. For a better understanding of the principle, we provide Algorithm \ref{alg:crcam}; showing how to add the information of one \laam from one training data input into the \crcam, using the \emph{sum based \gswa} principle.

\begin{algorithm}
\small
\begin{algorithmic}[1]
\Function{sum\_\gswa\_crcam}{$C^{c},A^{\mathit{L_{AMA}}},X_{in}$}
    \For{$i = 0$ to len($A^{\mathit{L_{AMA}}}$)}
        \For{$j = 0$ to len($A^{\mathit{L_{AMA}}}[i]$)}
            \State $toSymbol \gets X_{in}[j]$
            \State $P \gets C^{c}[toSymbol]$
            \State $P[i][j] \gets P[i][j] + A^{\mathit{L_{AMA}}}[i][j]$
        \EndFor
    \EndFor
    \EndFunction
    
    \end{algorithmic}
    \caption{\small Process to add one aggregated \laam $A^{\mathit{L_{AMA}}}$ from one specific symbol input sequence $X_{in}$ to a $v$ sized dictionary map $C^{c}$ (representing a \crcam for class $c$) each including a $n \times n$ matrix for each possible sequence position. To complete the \crcam this method needs to be called for each input from the training set.}
    \label{alg:crcam}
\end{algorithm}

\subsubsection{Global Trend matrix}

\begin{figure}[ht!]
	\centering
	\includegraphics[width=1\columnwidth]{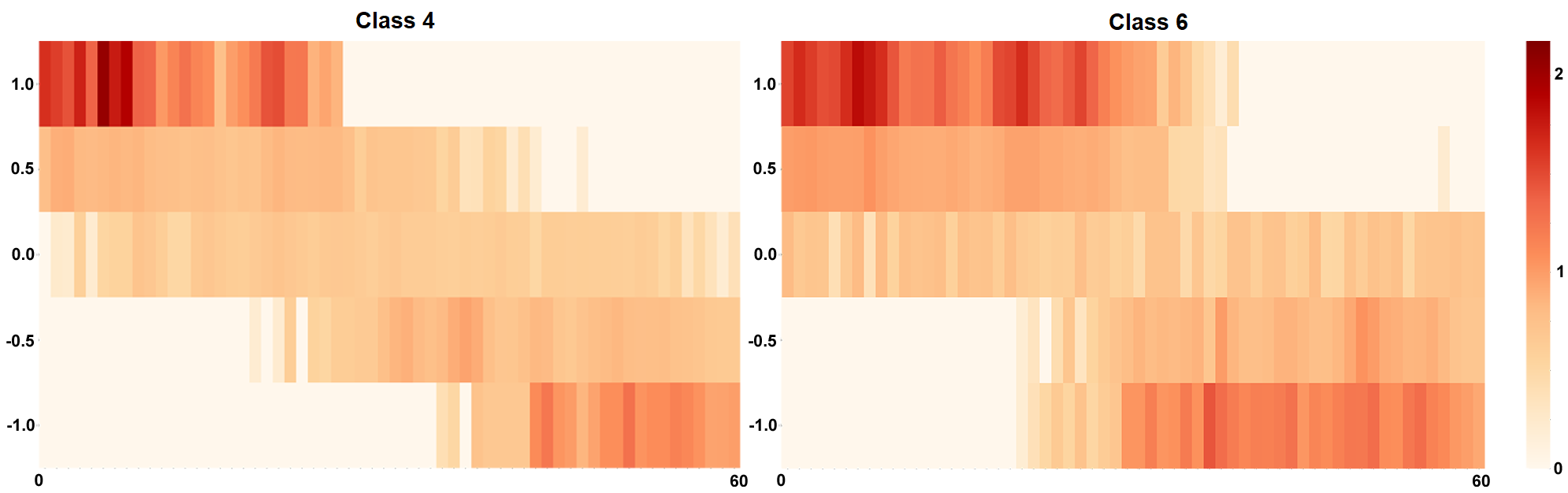}
	\caption{Two \gtms with maximum \gva, r. avg. \gswa and \textit{Sum-Sum} \laam from the Synthetic dataset, w.r.t. class 4 (left) representing a slowly falling trend and class 6 (right) representing a sudden falling trend.}
	\label{fig:minClassBoth}
\end{figure}

To make the general flow of a class even more clear, we reduce the \crcam even further by reducing the any-symbol of one symbol matrix into one vector/sequence. By putting the resulting symbol sequences in the related order, we receive one \acf{GTM}, \eg the two \gtms in Figure \ref{fig:minClassBoth}; showing where in the sequence each symbol is typically present and how strong attended it is. \Eg $P^{a-c}-i \in$ GTM shows how strong $a\in V$ is relevant for class $c$. For constructing multiple \gtm variations, we introduce three different \acf{GVA} methods. Keeping the two \fcam variations in mind, we end up with six \gtm variations we look into. We consider taking the maximum, median and average of the any-symbol relation as displayed in Algorithm \ref{alg:gtm}. We purposely use these rather simple aggregations, as they already achieve good results. However, due to the pipeline structure, it is also possible to replace them with more complex ones. 

\begin{algorithm}
\small
\begin{algorithmic}[1]
\Function{any\_gsv}{$C^{c}, G^{c}, f, V$}
    \For{$v$ in $V$}
        \State $G^{c}[v] \gets G^{c}[v] + f(C^{c}[v], axis=0)$
    \EndFor
    \EndFunction
    
    \end{algorithmic}
    \caption{\small Global Vector Aggregation process, to construct a \gtm $G^{c}$ of class $c$ out of a \crcam $C^{c}$, with any dimension reduction function $f$ -- \eg the maximum, median or average NumPy operation -- and a vocabulary set $V$. To complete the \gtm, this method needs to be called for each input from the training set.}
    \label{alg:gtm}
\end{algorithm}

\subsubsection{Global Class Validation}
\label{sec:globalVali}

Beside providing a great class distribution, the \gcr can also be used as a classifier. The idea behind the classification is that each symbol at each position has a statistical -- either based on occurrence (sum) or on the relative average Attention (r. avg.) -- affiliation with each class, while also depending on the different inputs to model dependencies. For the \fcam this means that each possible pair positions that occur in the time series is a part of a final class affiliation, \ie showing which pairs with which symbols are common for the class. To predict the class, we take one symbolized sequence $I$ and sum all related \gcr aggregated Attention values up to the sum score $V^c$, for each class $c \in C$. That means that for the \fcam we take all symbol pairs of the input and look up their values, resulting in $n^2 \times C$ lookups per $I$, where $C$ is the set of all classes. To relate each $V^c$ closer to the representing class, we divide each $V^c$ by the maximally reachable sum score $V^c_{max}$ per class $c$ to get the class membership score $S^c$ for Input $I$; \ie $V^c_{max}$ represents the sequence that has the strongest membership to the current class.

The biggest $S^c$ now indicates a percentage affiliation for class $c$ of the given input $I$. For the \crcam we also have $n^2 \times C$ lookups, but for the \gtm we only have $n \times C$ lookups. We want to note that because of the complexity to calculate $S^c_{max}$ for the \fcam, we only approximate it with the sum of the maximally reachable value per position -- without considering the effect one ``best'' symbol could have at another. Further, it is important to note is that when optimizing the \crcam classification, the $V^c_{max}$ of the \crcam is always the $V^c_{max}$ from the average based \gtm. That means the results are the same as the average \gtms, and thus we do not need to calculate the \crcam, but keep the descriptions as an explanation step.

Based on this, we define the general membership-function:
\begin{equation}S^c(X) = \frac{\sum\limits_{i=1}^n  P(x_i \vert c, i) }{ \sum\limits_{i=1}^n  \max\limits_{v \in V}(P(v \vert c, i))} \end{equation}\\
with length of the time series $n$, all classes $C$, vocabulary $V$, $x_i \in X$ being the $i$-th element in sequence $X$ and function $P$ providing the probability of the given conditions.
If we now assume that the GTM does approximate the probability function $P$ we receive the GTM membership-function:\\
\begin{equation}S^c_{GTM}(X) = \frac{\sum\limits_{i=1}^n  P^{x_i-c}_i}{\sum\limits_{i=1}^n \max\limits_{v \in V}(P^{v-c}_i)}\end{equation}\\
with $P^{v-c}_i \in $ GTM being our GTM approximation of $P(v \vert c, i)$.
We extend this to the general 2D membership-function:
\begin{equation}
S^c_{2D}(X) = \frac{ \sum\limits_i^n\sum\limits_j^n  P(x_i,y_j \vert c, i,j) }{\sum\limits_i^n\sum\limits_j^n  \max\limits_{u \in V} \max\limits_{v \in V}(P(v,u \vert c, i,j))}\end{equation}\\
And for the FCAM membership-function:
\begin{equation}S^c_{FCAM}(X) =  \frac{\sum\limits_i^n\sum\limits_j^n  P^{x_iy_j-c}_{ij} }{\sum\limits_i^n\sum\limits_j^n  \max\limits_{u \in V} \max\limits_{v \in V}(P^{uv-c}_{ij})}\end{equation}\\

To elucidate on how the classification works, the different \gcrs can be seen as reward distributions, where the further the values are from the peak at each position, the less the input is part of the class. This obviously assumes that the aggregated Attention values represent to some degree an affiliation, and the more correct this assumption is, the better the results.
In the Appendix \ref{ap:gcrAlgo} we present the different classification GCR algorithms \ref{alg:maxScoreFCAM}, \ref{alg:classifyFCAM} and \ref{alg:classifyGTM}.

\subsubsection{Threshold Global Coherence Representation}

To calculate the infidelity of the \gcr, we introduce the \acf{GCR-T}. The process to create the \gcr is straightforward; all values below a certain threshold are ignored. We call this modified \gswa the Thresholds \gswa (\gswa-T). The threshold is based on the average value of all \laams from the training set. With this, we can show that less attended values are less important for the classification, while also removing low level noise in the representation.

\subsubsection{Class Penalty Global Coherence Representation}

\begin{figure}[b]
	\centering
	\includegraphics[width=1\columnwidth]{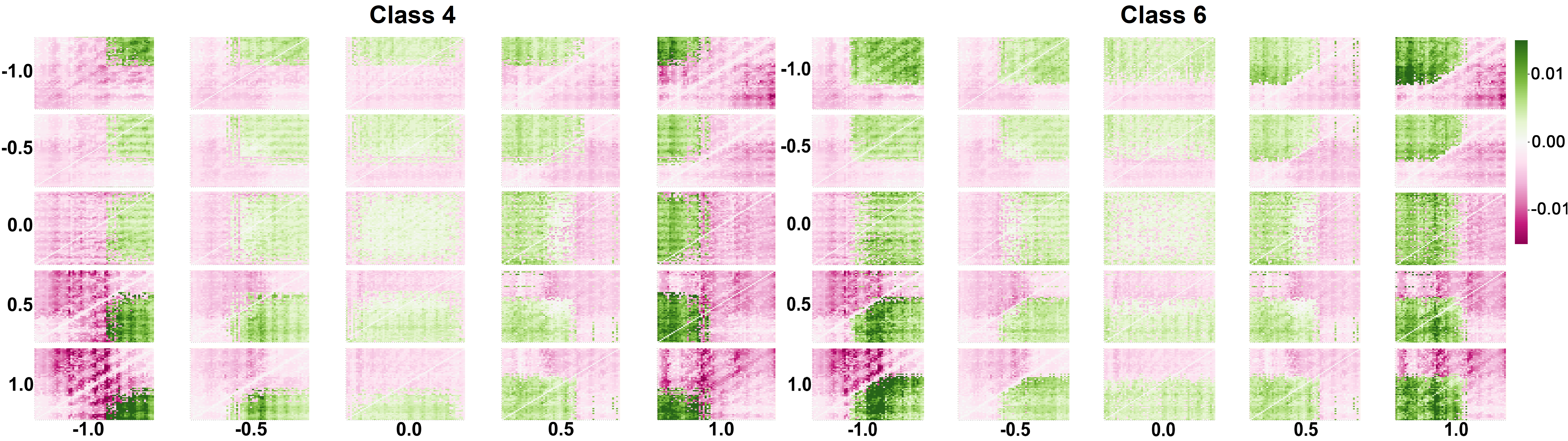}
	\caption{Two counting Penalty \fcams over the relative average \gswa and \textit{Max-Sum} \laam from the Synthetic dataset, w.r.t. class 4 (left) representing a slowly falling trend and class 6 (right) representing a sudden falling trend.}
	\label{fig:fullClassBothPenalty}
\end{figure}

\begin{figure}[ht!]
	\centering
	\includegraphics[width=1\columnwidth]{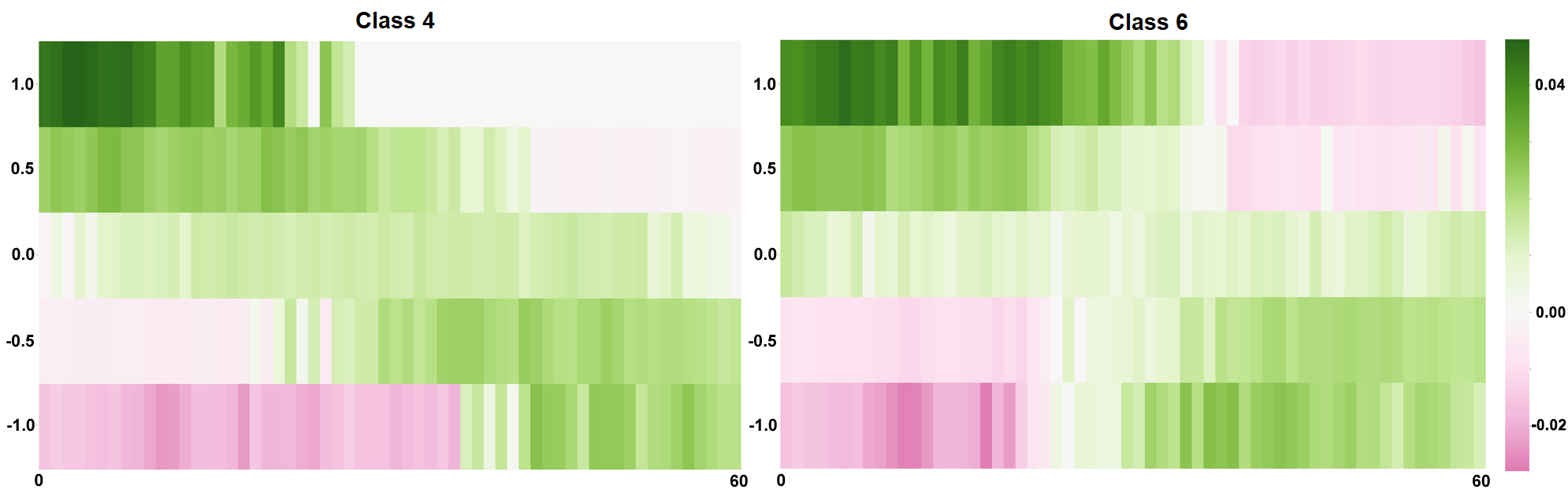}
	\caption{Two counting Penalty \gtms with average \gva, r. avg. \gswa and \textit{Max-Sum} \laam from the Synthetic dataset, w.r.t. class 4 (left) representing a slowly falling trend and class 6 (right) representing a sudden falling trend.}
	\label{fig:minClassBothPenalty}
\end{figure}

Compared to Shapelets the \gcr does only highlight the typical values of a class, regardless of how other classes look like. However, as we already discussed in \citep{SA:21:local} and \citep{SA:21:global}, the highlighting ability of Attention seem to focus on any important data point, regardless of the class. Which is why we introduce the \acf{GCR-P}. It reduces those class artefacts by introducing a class penalty, which penalizes one specific position for all other classes, if one Attention value is added to this specific position, \ie the \pgcr now focuses on the class uniquely highly attended positions. Hence, the representation now aims to maximize the unique class differences and gets penalized for unlikely positions. An example for two penalty-\fcams can be seen in Figures \ref{fig:fullClassBothPenalty} and for two penalty-\gtms in Figure \ref{fig:minClassBothPenalty}. Compared to the normal GCR, it can be seen that certain areas are penalized because they are more typical for other classes, and thus somewhat a uniqueness of an area can be seen.

\begin{algorithm}[H]
\scriptsize
\begin{algorithmic}[1]
    \Function{sum\_counter\_gswpa\_\fcam}{$F,A^{\mathit{L_{AMA}}},X_{in}, lCount, c, C$}
        \For{$i = 0$ to len($A^{\mathit{L_{AMA}}}$)}
            \For{$j = 0$ to len($A^{\mathit{L_{AMA}}}[i]$)}
                \State $fromSymbol \gets X_{in}[i]$
                \State $toSymbol \gets X_{in}[j]$
                \State $P \gets F^{c} [fromSymbol][toSymbol][i][j]$
                \State $sub \gets A^{\mathit{L_{AMA}}}[i][j] / lCount[c]$
                \State $v \gets \alpha \times  (\lvert C \rvert + 1) \times div$
                \State $P[i][j] \gets P[i][j] + v$
                \For{$d$ in $C$}
                    \State $dP \gets F^{d}[fromSymbol][toSymbol]$
                    \State $dp[i][j] \gets dp[i][j] - sub$
                \EndFor
            \EndFor
        \EndFor
    \EndFunction
    \end{algorithmic}
    \caption{\small Counting version of the Global Symbol Wise Penalty Aggregation step to add one aggregated \laam $A^{\mathit{L_{AMA}}}$, from one specific symbol input sequence $X$ of class $c$ to a nested $c \times v \times v$ dictionary map $F$ (including the \fcams for each class) each including a $n \times n$ matrix for each possible sequence position. $lCount$ is a dictionary including the count of instances per class and $C$ is a set of all possible classes. $\alpha$ is a scaling factor, in our case just 1. To complete the \fcam , this method needs to be called for each input from the training set.}

    \label{alg:fcamCOunting}
\end{algorithm}

We call this new aggregation step \acf{GSA-P}. We tried out two different kinds of penalties, one is based on a normalization using the number of classes and the number of instances per class (we refer to as ''counting'') and the other one is based on the entropy (we refer to as ''entropy'') of each class. Both options can be further fine-tuned by including a regularization factor for the reward or penalty. The counting option has a quite high penalty, while the entropy uses a rather small penalty. To show how the penalty works, we show the adjusted \fcam aggregation algorithm, in Algorithms \ref{alg:fcamCOunting} and \ref{alg:fcamEntropy}.

\begin{algorithm}
\scriptsize
\begin{algorithmic}[1]
    \Function{sum\_entropy\_gswpa\_\fcam}{$F,A^{\mathit{L_{AMA}}},X_{in}, lCount, tCount, c, C$}
        \For{$i = 0$ to len($A^{\mathit{L_{AMA}}}$)}
            \For{$j = 0$ to len($A^{\mathit{L_{AMA}}}[i]$)}
                \State $fromSymbol \gets X_{in}[i]$
                \State $toSymbol \gets X_{in}[j]$
                \State $P \gets F^{c} [fromSymbol][toSymbol][i][j]$
                \State $e = lCount[c] / tCount$
                \State $entropy = -(e * log(e))$
                \State $v \gets \alpha \times (\lvert C \rvert + 1) \times A^{\mathit{L_{AMA}}}[i][j]  /entropy$
                \State $P[i][j] \gets P[i][j] + v$
                \State $sub \gets  A^{\mathit{L_{AMA}}}[i][j] \times entropy$
                \For{$d$ in $C$}
                    \State $dP \gets F^{d}[fromSymbol][toSymbol]$
                    \State $dP[i][j] \gets dp[i][j] - sub$
                \EndFor
            \EndFor
        \EndFor
    \EndFunction
    \end{algorithmic}
    \caption{\small Entropy version of the Global Symbol Wise Penalty Aggregation step to add one aggregated \laam $A^{\mathit{L_{AMA}}}$, from one specific symbol input sequence $X$ of class $c$ to a nested $c \times v \times v$ dictionary map $F$ (including the \fcams for each class) each including a $n \times n$ matrix for each possible sequence position. $lCount$ is a dictionary including the count of instances per class and $C$ is a set of all possible classes and $tCount$ is the number of instances in the training set. $\alpha$ is a scaling factor, in our case just 1. To complete the \fcam , this method needs to be called for each input from the training set.}

    \label{alg:fcamEntropy}
\end{algorithm}


%% file: s5_results.tex
\section{Results}
\label{sec:results}
In this section, we display, compare and analyse the results for our approaches.
\subsection{Evaluation Data}

To evaluate our approaches, we look into all univariate datasets from the UCR UEA time series repository provided by the tslearn toolkit \cite{tslearn}. Due to some computational limitations, we only consider time series sequences of length 500 or smaller. Further, we limit our computation to one day per dataset per setting. At the end, we ended up with 170 successfully finished pipeline runs, over 37 unique datasets.

\subsection{Scope of Analysed Parameters}
\label{sec:config}
For our experiments, we test different hyperparameters to assess the influence of each parameter on the results. Due to the high experiment count, the number of parameter options is quite limited. We differentiate between static model parameters and grid searched parameters, for which each complete parameter-set a baseline model is trained. For the parameters, see Table \ref{tab:baseSettings}. The \textit{lowLayer} and \textit{highLayer} in the grid search differentiate further the possible combinations based on the Attention layer count. Therefore, we have 3 original baseline model parameter sets and 6 SAX baselines model parameter sets per dataset. To reference those configurations later on we introduce the abbreviation 2s-2l-8h, \ie 2 symbols, 2 layers and 8 heads. To select the different parameters, we did some manual sample testing on a few datasets while trying to optimize the general performance. For further parameters, we refer to our implementation linked in Section \ref{sec:introduction}.

The number of layers is decided to be rather low, because that is sufficient on most time series tasks, which is supported by the experiments in \cite{wen2022transformers}, showing that time series tasks perform best on a low layer count with 3-6 layers.

\begin{table}[ht]
\centering
\caption{Hyperparameter used to create the different Transformer models, differentiate between low and high layer count.}
\begin{tabular}{lc|c|ll|c|c}
\textbf{static:} & Variable   & Values &   \textbf{grid:} &           & Variable      & Values            \\
\hline
        & Error func.& MSE &        &           & symbol count     & {[}3,5{]}   \\
        & Optimizer  & Adam&        & lowLayer  &                  &             \\
        & Warmup steps & 10000&     &           & Attention layers & {[}2{]}     \\
        & folds      & 5   &        &           & head count       & {[}8, 16{]} \\
        & patience   & 70  &        &           & dff              & {[}8{]}     \\
        & batch size & 50  &        & highLayer &                  &             \\
        & epochs     & 500 &        &           & Attention layers & {[}5{]}     \\
        & dropout    & 0.3 &        &           & head count       & {[}6{]}     \\
        & dmodel     & 16  &        &           & dff              & {[}6{]}     
\end{tabular}
\label{tab:baseSettings}

\end{table}
\subsubsection{\lasa Parameters}
In Section \ref{sec:lasa} we introduced \lasa and the 16 possible combinations we considered so far, \eg ''hl-mmm''. However, due to our time limitation and the previous results from \cite{schwenke2021abstracting}, we only consider the ''hl'' reduction order. Additionally, we look into two maximum and two average based thresholds, to show their effects. They result in 32 possible settings per SAX baseline, indicated by Table \ref{tab:lasaParam}.

\begin{table}[ht]

\centering
\caption{Analysed \lasa hyperparameters to train the evaluation model.}
\begin{tabular}{c|c}
Variable                 & Values                          \\ \hline
combination steps 1-3    & {[}max, sum{]}                  \\
layer combinations       & {[}hl{]}                        \\
average based thresholds & {[}{[}1,1.2{]},{[}0.8,1.5{]}{]} \\
maximum based thresholds & {[}{[}2,3{]},{[}1.8,-1{]}{]}   
\end{tabular}
\label{tab:lasaParam}
\end{table}

\subsubsection{\gcr Parameters}
For the \gcr we only have two combination steps from the \laam (again excluding ''lh'') and thus we calculate 4 times each \gcr type, \ie resulting in 32 total \gcrs per SAX baseline. Further, we introduce two singular thresholds for the threshold \gcr and two penalty modes for the penalty \gcr. In total, we evaluate 160 different \gcr variants for each SAX baseline, respectively. The analysed parameters can be seen in Table \ref{tab:gcrParam}.

\begin{table}[ht]
\centering
\caption{Analysed hyperparameters for the different \gcr variations.}
\begin{tabular}{c|c}
Variable                 & Values                          \\ \hline
combination steps 1-2     & {[}max, sum{]}                    \\
layer combinations        & {[}hl{]}                          \\
avg. based \gcr thresholds & {[}1.3, 1.6, 1.0{]}               \\
penalty modes             & {[}entropy, counting{]}          
\end{tabular}
\label{tab:gcrParam}
\end{table}

\subsubsection{Shapelet Parameters}
For the Shapelet calculation, we consider the settings as shown in Table \ref{tab:shapletSettings}. The configurations resulted in one Shapelet model for each original and each SAX model and, additionally, a Shapelet model for each \lasa abstraction. For the classification, we trained a random forest model based on the found Shapelets. For our implementation, we use the following python package: Sktime \textit{ContractedShapeletTransform} \cite{sktime} to find Shapelets and Sklearn \cite{sklearn_api} for the Random Forest classifier.

\begin{table}[ht]
\centering
\caption{Settings for the Shapelet model. The 0.3 in Shapelet length means $\frac{1}{3}$ of the training data sequence length.}
\begin{tabular}{c|c}
Variable                 & Values                          \\ \hline
initial number of Shapelets per case & 5           \\
time constraint                      & 1           \\
minimal Shapelet length              & {[}2,0.3{]} \\
random forest estimators             & 100
\end{tabular}
\label{tab:shapletSettings}
\end{table}

\subsection{Experimental Summary}
To summarize the number of models per experiment fold run (1 out of 5 folds), we present Table \ref{tab:nrModelSummary}.

\begin{table}[h!]
\centering
\footnotesize
\caption{Summary of number of models per one fold in an experiment run.}
\begin{tabular}{lll}
1 Original model & 1 SAX model & 32 \lasa models \\
1 Original Shapelets model & 1 SAX Shapelets model & 32 \lasa Shapelets models \\
160 \gcr variations: &  &  \\ \hline
32  base \gcr models & 64 Penalty \gcr models & 64 Threshold \gcr models
\end{tabular}
\label{tab:nrModelSummary}
\end{table}

In total, we have 6 experiment runs per dataset, with 85 possible univariate datasets from the UCR UEA repository~\cite{bagnall2017great}. We exclude datasets with a sequence size longer than 500 and dataset configurations where the experiment pipeline did not finish in under a day. Thus, in total we had 1140 models per experiment with 170 finished individual experiment runs using 37 unique datasets.


\subsection{Baseline}
\label{sec:baseline}
For the baseline, we take the original data from each dataset and train a model for each setting with it -- referred to as \textit{Ori} in the results. By including the SAX algorithm (Section \ref{sec:sax}) in our pipeline, we abstract the data into symbols. Because we can't know which symbol count works best for which data beforehand, we train another baseline model for each mode setting on the symbolised data -- referred to as \textit{SAX} in the results -- to see if the symbol abstraction still contains all necessary data for training. Additionally, we also calculated a Shapelet baseline for each baseline as an additional state-of-the-art reference. To reduce the data-load (different datasets and different settings), we always provide the averaged results, including the standard deviation. For better comparison to the baseline, analysed models we want to compare will be provided as the average difference to the SAX model (baseline). Due to the huge amount of results, we will only show the most interesting ones, but added more detailed Tables in the Appendix. Per each inner (sub-)table for each scenario, we highlight the best performing values in bold. If no value is bold, the results can't be compared or the best value can be found in the Appendix, however, the shown values are most of the time quite close to the best.

Because neural networks can overfit easily and often need to be fine-tuned per problem, we differentiate in the results between \textit{all} results and \textit{good} results, with an SAX average n-fold cross validation baseline accuracy $>75\%$, \ie our selected model setting can train the model successfully. Typical problems can be that the model parameters are suboptimal and thus the model overfits or underfits, the learning rate steps are suboptimal and hence one or multiple folds do not converge. Because we do not differentiate for each dataset, multiple problems can lead to suboptimal results, which we however aim to analyse to some extent.

\begin{table}[h!]
\footnotesize
\centering
\setlength{\tabcolsep}{5pt}

\caption{Average performance baseline results of the Transformer model.}
\begin{tabular}{l|c|cccc}
Config & Data & Accuracy & Precision & Recall & F1  Score\\ \hline
All & \textit{Ori} & 0.6392 $\pm$ 0.1860 & 0.5771 $\pm$ 0.2402 & 0.6081 $\pm$ 0.1901 & 0.5593 $\pm$ 0.2292 \\
 & \textit{SAX} & 0.6262 $\pm$ 0.1752 & 0.5674 $\pm$ 0.2255 & 0.5880 $\pm$ 0.1841 & 0.5428 $\pm$ 0.2163 \\ \hline
 5s-2l-8h & \textit{Ori} & 0.6659 $\pm$ 0.1692 & 0.6057 $\pm$ 0.2183 & 0.6234 $\pm$ 0.1797 & 0.5836 $\pm$ 0.2129 \\
 & \textit{SAX} & 0.6569 $\pm$ 0.1664 & 0.5939 $\pm$ 0.2156 & 0.6103 $\pm$ 0.1761 & 0.5733 $\pm$ 0.2067 \\ \hline
\begin{tabular}[c]{@{}l@{}}SAX Acc. \\ $\geq 75\%$\end{tabular} & \textit{Ori} & \multicolumn{1}{l}{0.7879 $\pm$ 0.1767} & \multicolumn{1}{l}{0.7384 $\pm$ 0.2230} & \multicolumn{1}{l}{0.7519 $\pm$ 0.1968} & \multicolumn{1}{l}{0.7284 $\pm$ 0.2229} \\
(Good) & \textit{SAX} & \multicolumn{1}{l}{0.8453 $\pm$ 0.0770} & \multicolumn{1}{l}{0.8114 $\pm$ 0.1238} & \multicolumn{1}{l}{0.8017 $\pm$ 0.1318} & \multicolumn{1}{l}{0.7963 $\pm$ 0.1337}

\end{tabular}

\label{tab:baseline}
\end{table}

\begin{table}[h!]
\footnotesize
\centering
\caption{Average side information baseline results for the different neural network configurations. Good models have a SAX accuracy above 75\%.}
\begin{tabular}{l|cccc}
Config & \begin{tabular}[c]{@{}c@{}}SAX to Ori\\ Train Model Fidelity\end{tabular} & \begin{tabular}[c]{@{}c@{}}SAX to Ori\\ Test Model Fidelity\end{tabular} & Nr. of Data & Good Models \\ \hline
All & 0.7023 $\pm$ 0.2109 & 0.6593 $\pm$ 0.1972 & 170 & 38 \\ 
3s-2l-8h & 0.6838 $\pm$ 0.2047 & 0.6361 $\pm$ 0.1958 & 30 & 5 \\ 
3s-2l-16h & 0.7090 $\pm$ 0.1933 & 0.6568 $\pm$ 0.1782 & 32 & 5 \\ 
3s-5l-6h & 0.6691 $\pm$ 0.2316 & 0.6367 $\pm$ 0.2157 & 25 & 3 \\ 
5s-2l-8h & \textbf{0.7307 $\pm$ 0.2072} & 0.6816 $\pm$ 0.1979 & 31 & 10 \\ 
5s-2l-16h & 0.7234 $\pm$ 0.1940 & \textbf{0.6841 $\pm$ 0.1794} & 30 & 8 \\ 
5s-5l-6h & 0.6865 $\pm$ 0.2358 & 0.6551 $\pm$ 0.2170 & 22 & 7
\end{tabular}

\label{tab:baselineSide}
\end{table}

Table \ref{tab:baseline} shows the performance baseline for the Ori and SAX data and Table \ref{tab:baselineSide} shows some side information for each configuration. It is important to note that, while comparing the different results for the different model configurations, different datasets can be included in the average, because some configurations did not finish in the 24h limit and thus some datasets per configuration might be missing. The failure of many datasets to meet our criterion for "good" performance can be explained by the fact that neural networks typically require a lot of fine-tuning. Overall, 5s-2l-8h performed the best, which could be attributed to the higher amount of fine-tuning that was done on it. To evaluate each of our methods, we take each dataset performance result into relation to the baseline, \ie we show each performance metric as the difference to the respective SAX model. When comparing the Ori and SAX performance, it can be seen, that the SAX performance drops on average compared to the Ori performance, in all settings but for 5s-2l-16h. However, for all 5 symbols settings, the performance is always quite close, showing that a good number of symbols is an important approximation for the classification (see appendix Table \ref{tabA:baseline}). For better comparison, we introduced the last row in Table \ref{tab:baseline}, showing the average performance of the good models (SAX model acc. $\geq 75\%$). Interesting to note is that the Ori performance in this tape is lower on average.

When looking at the model fidelity of the SAX models to the Ori models in Table \ref{tab:splitFidelity}, we see that quite a few samples are typically classified differently, even though the data was only symbolised. This is even the case for the 5s-2l-16h setting, which outperformed the original model on average. Looking only at the good performing models (SAX accuracy $\geq 75\%$) in Table \ref{tab:splitFidelity}, the train model fidelity and test model fidelity increased a bit. Nonetheless, both models still show some differences in the classification. 
However, this is in line with results mentioned in \cite{kim2020puzzle,li2021interpretable}, where data augmentation can change the model interpretation, \ie our SAX data augmentation does not remove (much) important information, but shifts the features and their weights the model uses for classification.

\begin{table}[h!]
\centering
\footnotesize
\caption{Ori Model Fidelity of the SAX models split by symbol count and model performance.}
\begin{tabular}{c|cc}
 & \begin{tabular}[c]{@{}c@{}}Ori Train \\ Model Fidelity\end{tabular} & \begin{tabular}[c]{@{}c@{}}Ori Test \\ Model Fidelity\end{tabular} \\ \hline
All & \cellcolor[HTML]{C0C0C0} & \cellcolor[HTML]{C0C0C0} \\ \hline
SAX 5 & 0.7163 $\pm$ 0.2114 & 0.6755 $\pm$ 0.1972 \\
SAX 3 & 0.6888 $\pm$ 0.2094 & 0.6439 $\pm$ 0.1959 \\ \hline
Good & \cellcolor[HTML]{C0C0C0} & \cellcolor[HTML]{C0C0C0} \\ \hline
SAX 5 & 0.8316 $\pm$ 0.1695 & 0.7893 $\pm$ 0.1601 \\
SAX 3 & 0.7719 $\pm$ 0.1895 & 0.7220 $\pm$ 0.1615
\end{tabular}

\label{tab:splitFidelity}
\end{table}

\subsection{Shapelets Baseline}
Table \ref{tab:baselineShap} shows the performance baseline via the Shapelets algorithm, using the original (Ori) and symbolised (SAX) data. The data is based on the results for all 37 unique datasets that finished. SAX $n$ stands for the SAX symbol count and \textit{slen} is the minimal Shapelet length, either 0.3 of the maximum sequence length per dataset or just 2. Most of the time only \textit{slen} 0.3 is shown, because it performed better. As one can see, the performance of the Shapelet algorithm is on average over all datasets significantly better compared to the neural network, because it is more adapted for this kind of problem and does not need any fine-tuning. However, comparing the results on the good Shapelets models, to the last row of Table \ref{tab:baseline} we see that the SAX model's performance is quite close on average to the Shapelets results, while have a smaller standard deviation; showing that the Transformer can have comparable results to the Shapelets model if fine-tuned.

\begin{table}[h!]
\footnotesize
\centering
\setlength{\tabcolsep}{4pt}
\caption{Average performance baseline results for the different Shapelet configurations. SAX $n$ means we used SAX data with $n$ symbols and \textit{slen} is the set minimal Shapelet length.}
\begin{tabular}{c|c|cc|cc}
\multicolumn{1}{l|}{} & \multicolumn{1}{l|}{} & \multicolumn{1}{l}{All} & \multicolumn{1}{l|}{} & \multicolumn{1}{l}{Good} & \multicolumn{1}{l}{} \\
slen & Data & Acc. & F1 & Acc. & F1 \\ \hline
2 & \textit{Ori} & \textbf{0.7978 $\pm$ 0.1622} & \textbf{0.7572 $\pm$ 0.2026} & 0.8526 $\pm$ 0.1409 & 0.8071 $\pm$ 0.1764 \\
\multicolumn{1}{c|}{} & \textit{SAX 3} & \multicolumn{1}{c}{0.6901 $\pm$ 0.1707} & \multicolumn{1}{c|}{0.6327 $\pm$ 0.2177} & \multicolumn{1}{c}{0.8078 $\pm$ 0.1810} & \multicolumn{1}{c}{0.7957 $\pm$ 0.1891} \\
 & \textit{SAX 5} & 0.7216 $\pm$ 0.1700 & 0.6712 $\pm$ 0.2088 & 0.7967 $\pm$ 0.1652 & 0.7410 $\pm$ 0.1989 \\ \hline
0.3 & \textit{Ori} & 0.7895 $\pm$ 0.1540 & 0.7544 $\pm$ 0.1871 & \textbf{0.8558 $\pm$ 0.1291} & 0.8154 $\pm$ 0.1628 \\
\multicolumn{1}{c|}{} & \textit{SAX 3} & 0.7132 $\pm$ 0.1503 & 0.6680 $\pm$ 0.1961 & 0.8445 $\pm$ 0.1456 & \textbf{0.8369 $\pm$ 0.1475} \\
 & \textit{SAX 5} & \multicolumn{1}{c}{0.7382 $\pm$ 0.1559} & \multicolumn{1}{c|}{0.6953 $\pm$ 0.1912} & \multicolumn{1}{c}{0.8133 $\pm$ 0.1542} & \multicolumn{1}{c}{0.7596 $\pm$ 0.1929}
\end{tabular}
\label{tab:baselineShap}
\end{table}

In Table \ref{tab:fidelityXAI} the model fidelity of the Shapelets model to the Ori and SAX model is shown, to compare the similarity between those models. Comparing Table \ref{tab:splitFidelity} we can see that the SAX model's Ori model fidelity is always higher than the Shapelets one, but for the train fidelity for SAX 3 slen 0.3. This is even more so the case for the test model fidelity, \ie the principal on which the SAX model classifies is closer to the Ori model than the Shapelets model to the Ori model, which can also be seen by the clear train to test gap for the Shapelets model fidelities.

\begin{table}[h!]
\centering
\footnotesize
\setlength{\tabcolsep}{4pt}
\caption{Model fidelity of slen 0.3 Shapelets baselines, to the Ori and SAX model.}
\begin{tabular}{c|cccc}
\begin{tabular}[c]{@{}c@{}}Shapelets\\ slen 0.3\end{tabular} & \begin{tabular}[c]{@{}c@{}}Ori Train \\ Model Fidelity\end{tabular} & \begin{tabular}[c]{@{}c@{}}Ori Test \\ Model Fidelity\end{tabular} & \begin{tabular}[c]{@{}c@{}}SAX Train \\ Model Fidelity\end{tabular} & \begin{tabular}[c]{@{}c@{}}SAX Test \\ Model Fidelity\end{tabular} \\ \hline
All & \cellcolor[HTML]{C0C0C0} & \cellcolor[HTML]{C0C0C0} & \cellcolor[HTML]{C0C0C0} & \cellcolor[HTML]{C0C0C0} \\ \hline
Ori & 0.6998 $\pm$ 0.2398 & \textbf{0.6195 $\pm$ 0.2071} & 0.6742 $\pm$ 0.2170 & 0.6115 $\pm$ 0.1964 \\
SAX 5 & \textbf{0.7028 $\pm$ 0.2396} & 0.6061 $\pm$ 0.1982 &\textbf{ 0.6848 $\pm$ 0.2266} & \textbf{0.6268 $\pm$ 0.2064} \\
SAX 3 & 0.6829 $\pm$ 0.2294 & 0.5773 $\pm$ 0.1790 & 0.6813 $\pm$ 0.2222 & 0.6144 $\pm$ 0.2042 \\ \hline
Good & \cellcolor[HTML]{C0C0C0} & \cellcolor[HTML]{C0C0C0} & \cellcolor[HTML]{C0C0C0} & \cellcolor[HTML]{C0C0C0} \\ \hline
Ori & 0.8118 $\pm$ 0.2292 & \textbf{0.7223 $\pm$ 0.2193} & 0.8629 $\pm$ 0.1669 & 0.7799 $\pm$ 0.1870 \\
SAX 5 & 0.8023 $\pm$ 0.2368 & 0.7075 $\pm$ 0.2130 & 0.8544 $\pm$ 0.1911 & \textbf{ 0.7824 $\pm$ 0.1989} \\
SAX 3 &\textbf{ 0.8232 $\pm$ 0.2022} & 0.6649 $\pm$ 0.1804 & \textbf{0.8885 $\pm$ 0.0931} & 0.7446 $\pm$ 0.1742
\end{tabular}
\label{tab:fidelityXAI}
\end{table}

\subsection{Consistency}
As described in Section \ref{sec:metrics} we want to look into the consistency of the constructed \laams. The three averaged different distances can be seen in Table \ref{tab:consistency}. Looking at the overall standard deviations, we can see that the standard deviation for the \textit{OuterDistance} is always the smallest and thus indicating at least one form of consistency. Nonetheless, looking at the matrix euclidean distance $md$ we can see that for the all model results the \textit{OuterDistance} is, in all but the hl-mm combination, closer to the \textit{InnerClassDistance} than the \textit{InnerFoldDistance}; indicating inconsistent results. This however can be explained by the bad performing models, that can easily mislearn the different classes. Looking at the average performance of only the good models, the \textit{OuterDistance} is always closer to the \textit{InnerClassDistance} than the \textit{InnerFoldDistance}, in some cases even lower than the \textit{InnerFoldDistance}. This indicates at least some form of consistency for well-trained models -- considering that the $md$ is quite a simple comparison metric and that each fold has other training data.

\begin{table}[h!]
\centering
\footnotesize
\caption{Three different distances calculated the $md$ to estimate the consistency.}
\begin{tabular}{l|ccc}
 & OuterDistance & InnerClassDistance &  InnerFoldDistance\\ \hline
All & \cellcolor[HTML]{C0C0C0} & \cellcolor[HTML]{C0C0C0} & \cellcolor[HTML]{C0C0C0} \\ \hline
hl-mm & 0.5356 $\pm$ 0.4958 & 0.5234 $\pm$ 0.6758 & 0.6152 $\pm$ 0.7477 \\
hl-ms & 0.7423 $\pm$ 0.7079 & 0.6892 $\pm$ 0.9330 & 0.8152 $\pm$ 0.9989 \\
hl-sm & 1.7941 $\pm$ 1.6734 & 1.4874 $\pm$ 2.0728 & 1.7439 $\pm$ 2.4122 \\
hl-ss & 2.6052 $\pm$ 2.1707 & 2.1155 $\pm$ 2.7610 & 2.4725 $\pm$ 3.0567 \\ \hline
Good & \cellcolor[HTML]{C0C0C0} & \cellcolor[HTML]{C0C0C0} & \cellcolor[HTML]{C0C0C0} \\ \hline
hl-mm & 0.7209 $\pm$ 0.6128 & 0.7824 $\pm$ 0.9450 & 0.9353 $\pm$ 0.8881 \\
hl-ms & 1.0291 $\pm$ 0.9242 & 1.1313 $\pm$ 1.4549 & 1.3466 $\pm$ 1.3468 \\
hl-sm & 2.7119 $\pm$ 1.8319 & 2.5003 $\pm$ 2.8048 & 2.9253 $\pm$ 2.7388 \\
hl-ss & 3.8562 $\pm$ 2.3899 & 3.7490 $\pm$ 4.0038 & 4.3367 $\pm$ 3.7686
\end{tabular}
\label{tab:consistency}
\end{table}
\subsection{Local Input Abstraction Results}
In this section, we provide the results for the \lasa method from Section \ref{sec:lasa}. Additional information can be found in the appendix Tables \ref{tabA:lasaAR}-\ref{tabA:lasaShapAcc}.
\subsubsection{Performance} 

The most important performance features for the \lasa method are the accuracy and the data reduction. Based on those two, the thresholds need to be optimized; therefore, we present four different thresholds. Table \ref{tab:lasaAR} shows those two parameters for two (overall best and worst) possible \laav combinations for each threshold. The edge cases \textit{hl-mmm} and \textit{hl-sss} perform per combination most of the time the best or worst in certain areas, or are close to it. All not shown combinations are performance-wise in-between the shown ones. As one can see, using a good selection for both threshold parameters is quite crucial for the accuracy and data reduction results. The average threshold results reach a data reduction up to $92\%$, or an accuracy loss of minimally $0.0083\%$, showing that the human-in-the-loop process can be driven by individual preferences. Important to keep in mind is that typically each dataset needs its own threshold fine-tuning. For example, the max-based threshold gives a good example that the overall accuracy to reduction ratio is a lot worse than the average-based ones. However, when considering the standard deviation it indicates that some datasets had a better data reduction, thus the max-based thresholds are not well generally applied. The avg-based threshold had overall quite consistent results, but interestingly the performance of the accuracy and data reductions on the basis of the \laav configuration flips between both avg-based thresholds. This further shows how much influence a good threshold can have, but also shows that the analyses in \cite{schwenke2021abstracting} were not far-reaching enough. While the average threshold results perform on average over all datasets relatively well, the results for the good datasets, perform worse with a higher standard deviation. However, considering that we used a generalized threshold set, the results are relatively good, because we do not consider the individual form of each dataset.
\begin{table}[h!]
\centering
\footnotesize
\setlength{\tabcolsep}{2pt}
\caption{Relative Accuracy (to SAX model) with reduction results between different thresholds and combinations for the \lasa method.}
\begin{tabular}{l|cc|cc}
 & All (170) &  & Good (38) &  \\
Combi & Accuracy & Reduction & Accuracy & Reduction \\ \hline
\multicolumn{1}{l|}{\begin{tabular}[l]{@{}c@{}}Avg. {[}1, 1.2{]}\end{tabular}} & \cellcolor[HTML]{C0C0C0} & \cellcolor[HTML]{C0C0C0} & \cellcolor[HTML]{C0C0C0} & \cellcolor[HTML]{C0C0C0} \\ \hline
hl-mmm & -0.0364 $\pm$ 0.0754 & \textbf{0.5399 $\pm$ 0.0859} & -0.1108 $\pm$ 0.0989 & 0.5553 $\pm$ 0.0678 \\
hl-sss & -0.0258 $\pm$ 0.0616 & 0.4624 $\pm$ 0.0792 & \textbf{-0.0707 $\pm$ 0.0795} & 0.4828 $\pm$ 0.0614 \\ \hline
\multicolumn{1}{l|}{\begin{tabular}[l]{@{}c@{}}Avg. {[}0.8, 1.5{]}\end{tabular}} & \cellcolor[HTML]{C0C0C0} & \cellcolor[HTML]{C0C0C0} & \cellcolor[HTML]{C0C0C0} & \cellcolor[HTML]{C0C0C0} \\ \hline
hl-mmm & \textbf{-0.0474 $\pm$ 0.0660} & 0.7990 $\pm$ 0.1196 & -0.1015 $\pm$ 0.0779 & 0.7041 $\pm$ 0.0965 \\
hl-sss & -0.1264 $\pm$ 0.1439 & \textbf{0.9664 $\pm$ 0.0534} & -0.2659 $\pm$ 0.1657 & \textbf{0.9207 $\pm$ 0.0854} \\ \hline
\multicolumn{1}{l|}{\begin{tabular}[l]{@{}c@{}}Max {[}2, 3{]}\end{tabular}} &  \cellcolor[HTML]{C0C0C0} & \cellcolor[HTML]{C0C0C0} & \cellcolor[HTML]{C0C0C0} & \cellcolor[HTML]{C0C0C0} \\ \hline
hl-mmm & -0.0575 $\pm$ 0.1071 & \textbf{0.4076 $\pm$ 0.2721} & -0.1657 $\pm$ 0.1459 & \textbf{0.5766 $\pm$ 0.2533} \\
hl-sss & -0.0142 $\pm$ 0.0639 & 0.0497 $\pm$ 0.1509 & -0.0439 $\pm$ 0.0800 & 0.0422 $\pm$ 0.1137 \\ \hline
\multicolumn{1}{l|}{\begin{tabular}[l]{@{}c@{}}Max {[}1.8, -1{]}\end{tabular}} &  \cellcolor[HTML]{C0C0C0} & \cellcolor[HTML]{C0C0C0} & \cellcolor[HTML]{C0C0C0} & \cellcolor[HTML]{C0C0C0} \\ \hline
hl-mmm & -0.0699 $\pm$ 0.1024 & \textbf{0.4720 $\pm$ 0.2745} & -0.1784 $\pm$ 0.1163 & 0.6432 $\pm$ 0.2411 \\
hl-sss & \textbf{-0.0083 $\pm$ 0.0542} & 0.0619 $\pm$ 0.1729 & \textbf{-0.0224 $\pm$ 0.0613} & 0.0644 $\pm$ 0.1450
\end{tabular}

\label{tab:lasaAR}
\end{table}
\subsubsection{Performance Ranking}
Table \ref{tab:lasaRankings} shows the effect of the model configurations on the results per \laav configuration. We ranked the accuracy and data reduction of each \laav configuration per \lasa configuration into one average ranking score, which shows which \laam configuration has the best accuracy to data reduction ratio. The rankings in Table \ref{tab:lasaRankings} are the average rank between those both. As seen in Table \ref{tab:lasaRankings} the impact of the different model parameters rather low. In contrast, the thresholds have a huge impact on the performance of the \laav combinations; \eg the average ${[} 0.8, 1.5 {]}$ threshold flips the rankings. This shows that fine-tuning each problem with a good threshold is very important to maximize the accuracy and reduction ratio, including both parts of the threshold. It should also be noted that although the \laav combinations seem to have a tendency in the rankings, the inner ranking between the different datasets can vary depending on the data, \ie different problems may require different configurations. Additionally, neural networks can be quite inconsistent, \eg one fold did not convert or the data cannot be reduced in the range of what the threshold intents to do, which introduces a general inconsistency in the results.

\begin{table}[h!]
\centering
\footnotesize
\setlength{\tabcolsep}{5pt}
\caption{Average Ranking of accuracy and reduction ratio.}
\begin{tabular}{l|ccccccc}
Rankings & All & 3s-2l-8h & 3s-2l-16h & 3s-5l-8h & 5s-2l-8h & 5s-2l-16h & 5s-5l-8h \\ \hline
\begin{tabular}[c]{@{}l@{}}Avg. {[}1, 1.2{]}\end{tabular} &  \cellcolor[HTML]{C0C0C0}&  \cellcolor[HTML]{C0C0C0}&  \cellcolor[HTML]{C0C0C0}&  \cellcolor[HTML]{C0C0C0}&  \cellcolor[HTML]{C0C0C0}&  \cellcolor[HTML]{C0C0C0}&  \cellcolor[HTML]{C0C0C0}  \\ \hline
hl-mmm & \textbf{3.3765} & \textbf{3.5833} & \textbf{3.3594} & 3.4600 & \textbf{3.4355} & \textbf{3.2000} & \textbf{3.1818} \\
hl-sss & 5.5441 & 5.5333 & 5.5938 & 5.7000 & 5.3387 & 5.6333 & 5.4773 \\ \hline
\begin{tabular}[c]{@{}l@{}}Avg. {[}0.8, 1.5{]}\end{tabular}&  \cellcolor[HTML]{C0C0C0}&  \cellcolor[HTML]{C0C0C0}&  \cellcolor[HTML]{C0C0C0}&  \cellcolor[HTML]{C0C0C0}&  \cellcolor[HTML]{C0C0C0}&  \cellcolor[HTML]{C0C0C0}&  \cellcolor[HTML]{C0C0C0}  \\ \hline
hl-mmm & 5.4235 & 5.3167 & 5.7031 & 5.6800 & 5.3387 & 5.1167 & 5.4091 \\
hl-sss & \textbf{3.2029} & \textbf{3.2667} & \textbf{3.2031} & \textbf{2.6600} & \textbf{3.4516} & \textbf{3.4500} & \textbf{3.0455} \\ \hline
\begin{tabular}[c]{@{}l@{}}Max {[}2, 3{]}\end{tabular}&  \cellcolor[HTML]{C0C0C0}&  \cellcolor[HTML]{C0C0C0}&  \cellcolor[HTML]{C0C0C0}&  \cellcolor[HTML]{C0C0C0}&  \cellcolor[HTML]{C0C0C0}&  \cellcolor[HTML]{C0C0C0}&  \cellcolor[HTML]{C0C0C0}  \\ \hline
hl-mmm & \textbf{3.3882} & \textbf{3.3667} & \textbf{3.2969} & 3.2400 & \textbf{3.5000} & \textbf{3.5333} & \textbf{3.3636} \\
hl-sss & 4.6265 & 4.7667 & 4.5469 & 4.6200 & 4.6613 & 4.5500 & 4.6136 \\ \hline
\begin{tabular}[c]{@{}l@{}}Max {[}1.8, -1{]}\end{tabular} &  \cellcolor[HTML]{C0C0C0}&  \cellcolor[HTML]{C0C0C0}&  \cellcolor[HTML]{C0C0C0}&  \cellcolor[HTML]{C0C0C0}&  \cellcolor[HTML]{C0C0C0}&  \cellcolor[HTML]{C0C0C0}&  \cellcolor[HTML]{C0C0C0}  \\ \hline
hl-mmm & \textbf{3.4647} & \textbf{3.2667} & \textbf{3.5156} & \textbf{3.4200} & \textbf{3.3710} & \textbf{3.5333} & \textbf{3.7500} \\
hl-sss & 4.7882 & 5.0167 & 4.7500 & 4.7000 & 4.8387 & 4.6167 & 4.7955
\end{tabular}

\label{tab:lasaRankings}
\end{table}

\subsubsection{Explainability} 

Table \ref{tab:lasaXAI} shows the model fidelity for the \lasa models, compared to the Shapelet baseline. However, the model fidelity can decrease considerably even with small changes in accuracy. This could again be due to the data augmentation \cite{kim2020puzzle,li2021interpretable} that is the consequence of the data interpolation; \ie due to the change in the shape of the data, we highlight different or even introduce new points of interest for the Transformer model, leading to different results while maintaining a similar accuracy performance. This gets further highlighted by the models with very low data reduction, performing very similar to the SAX model, with some variation that could be expected due to the inconsistency in the learning process of a neural network. Nonetheless, the data reduction only partly influences the model fidelity, \cf \textit{Avg. {[}0.8, 1.5{]} hl-mmm} and \textit{Max {[}2,3{]} hl-mmm} which have a similar model fidelity but a huge gap in the data reduction. In the end, we still show that Attention can help to abstract the original data, similar to some form of ``importance''.

\begin{table}[h!]
\centering
\footnotesize
\caption{Average model fidelity from different \lasa configurations to the SAX model.}
\begin{tabular}{l|ccl}
\begin{tabular}[c]{@{}c@{}}Rankings\\ Good\end{tabular} & \begin{tabular}[c]{@{}c@{}}SAX Train\\ Fidelity\end{tabular} & \begin{tabular}[c]{@{}c@{}}SAX Test\\ Fidelity\end{tabular} \\ \hline
\begin{tabular}[c]{@{}l@{}}Baseline\end{tabular} &  \cellcolor[HTML]{C0C0C0}&  \cellcolor[HTML]{C0C0C0}  \\ \hline
Ori Shapelets slen 0.3 & 0.8629 $\pm$ 0.1669 & \textbf{ 0.7799 $\pm$ 0.1870}  \\
SAX Shapelets slen 0.3 & \textbf{0.8661 $\pm$ 0.1651} &  0.7695 $\pm$ 0.1916  \\ \hline
\begin{tabular}[c]{@{}l@{}}Avg. {[}1, 1.2{]}\end{tabular} &  \cellcolor[HTML]{C0C0C0}&  \cellcolor[HTML]{C0C0C0}  \\ \hline
hl-mmm & 0.8244 $\pm$ 0.1404 & 0.7684 $\pm$ 0.1317  \\
hl-sss & \textbf{0.8658 $\pm$ 0.1057} & \textbf{0.8158 $\pm$ 0.1009}  \\ \hline
\begin{tabular}[c]{@{}l@{}}Avg. {[}0.8, 1.5{]}\end{tabular} &  \cellcolor[HTML]{C0C0C0}&  \cellcolor[HTML]{C0C0C0}   \\ \hline
hl-mmm & \textbf{0.8265 $\pm$ 0.1534} & \textbf{0.7705 $\pm$ 0.1370}  \\
hl-sss & 0.6739 $\pm$ 0.1625 & 0.6209 $\pm$ 0.1630  \\ \hline
\begin{tabular}[c]{@{}l@{}}Max {[}2, 3{]}\end{tabular} &  \cellcolor[HTML]{C0C0C0}&  \cellcolor[HTML]{C0C0C0}   \\ \hline
hl-mmm & 0.7770 $\pm$ 0.1240 & 0.7170 $\pm$ 0.1342  \\
hl-sss & 0.9032 $\pm$ 0.1046 & 0.8704 $\pm$ 0.1050  \\ \hline
\begin{tabular}[c]{@{}l@{}}Max {[}1.8, -1{]}\end{tabular}&  \cellcolor[HTML]{C0C0C0} &  \cellcolor[HTML]{C0C0C0}   \\ \hline
hl-mmm & 0.7571 $\pm$ 0.1190 & 0.6996 $\pm$ 0.1367  \\
hl-sss & \textbf{0.9192 $\pm$ 0.0869} & \textbf{0.8768 $\pm$ 0.1015} 
\end{tabular}

\label{tab:lasaXAI}
\end{table}

Comparing the model fidelity for SAX and Shapelets in Table \ref{tab:lasaXAI}; the \lasa model varies a lot. For the Shaplets model, the train and test model fidelity deviate roughly $10\%$, while the \lasa models show a higher agreement between the train and test fidelity as well as a lower standard deviation. This indicates that some learned core features are shared between the SAX and the \lasa model.

\subsubsection{Complexity} 
In \cite{schwenke2021abstracting} our results suggested a strong correlation between the data reduction and different complexity metrics. However, in our tests with more datasets this is not the case, as can be seen in Table \ref{tab:complexityPearson}, where the average $r$ Pearson correlation between the percentage reduction in complexity to the data reduction over all datasets is shown. Here the connection between the reduction of the data and the complexity is less clear, even though on average there is a very slight positive correlation for most thresholds. Also, it should be noted, that a non-linear correlation could be included, because complexity does not necessarily need to be linearly distributed in the data, \eg removing the in-between points in a line would not reduce complexity, while increasing the data reduction measurement. Looking at Tables \ref{tab:complexOri} and \ref{tab:complexSax}, we can see the average percentage reduction from the SAX or Ori complexity of the different settings. While the reduction is not on-pair with the data reduction, at least the \textit{Avg. [0.8, 1.5]} threshold seems to reduce the complexity of multiple metrics. Especially the T. Shifts seem to be somewhat correlated to the data reduction and thus also often have a high reduction in general. Interestingly, for SvdEn and CE, the complexity has increased compared to the original input. This however can have multiple reasons, from what the metric views as complexity, to suboptimal noise producing thresholds. Further, per dataset it is unclear what the minimal complexity of a class is and hence it is hard to compare a percentage reduction. This can also be seen in the sometimes very high standard deviation, indicating different base complexities. This also means that it is unclear what the minimal complexity of a dataset is.

\begin{table}[h!]
\centering
\footnotesize
\setlength{\tabcolsep}{4pt}
\caption{Average Pearson $r$ value over all datasets between the reduction of the complexity of the \lasa model in percent (compared to the Original or the SAX model) and the data reduction of the \lasa models. The different combinations are merged together and the table only differentiates between the different thresholds.}
\label{tab:complexityPearson}
\begin{tabular}{c|ccccc}
Comp. & SvdEn & ApEn & SampEn & CE & T. Shifts \\ \hline
Ori &  \cellcolor[HTML]{C0C0C0}&  \cellcolor[HTML]{C0C0C0}&  \cellcolor[HTML]{C0C0C0}&  \cellcolor[HTML]{C0C0C0}&  \cellcolor[HTML]{C0C0C0}  \\\hline
\begin{tabular}[c]{@{}c@{}}Avg. \\ {[}1, 1.2{]}\end{tabular} & 0.1327 $\pm$ 0.3544 & 0.0288 $\pm$ 0.3101 & 0.0167 $\pm$ 0.3288 & 0.0912 $\pm$ 0.3688 & 0.1793 $\pm$ 0.3124 \\
\begin{tabular}[c]{@{}c@{}}Avg. \\ {[}0.8, 1.5{]}\end{tabular} & 0.2246 $\pm$ 0.4161 & 0.2362 $\pm$ 0.4198 & 0.1667 $\pm$ 0.4131 & 0.1015 $\pm$ 0.4450 & 0.3774 $\pm$ 0.3405 \\
\begin{tabular}[c]{@{}c@{}}Max \\ {[}2, 3{]}\end{tabular} & 0.3924 $\pm$ 0.3631 & 0.2450 $\pm$ 0.3166 & 0.1539 $\pm$ 0.3430 & 0.2514 $\pm$ 0.3875 & 0.4875 $\pm$ 0.2998 \\
\begin{tabular}[c]{@{}c@{}}Max \\ {[}1.8, -1{]}\end{tabular} & 0.4408 $\pm$ 0.3417 & 0.2064 $\pm$ 0.3302 & 0.1132 $\pm$ 0.3626 & 0.2927 $\pm$ 0.3933 & 0.4885 $\pm$ 0.2878 \\ \hline
SAX &  \cellcolor[HTML]{C0C0C0}&  \cellcolor[HTML]{C0C0C0}&  \cellcolor[HTML]{C0C0C0}&  \cellcolor[HTML]{C0C0C0}&  \cellcolor[HTML]{C0C0C0}  \\ \hline
\begin{tabular}[c]{@{}c@{}}Avg. \\ {[}1, 1.2{]}\end{tabular} & 0.0494 $\pm$ 0.3741 & -0.0121 $\pm$ 0.3326 & -0.0846 $\pm$ 0.3710 & -0.0077 $\pm$ 0.3574 & 0.1260 $\pm$ 0.3454 \\
\begin{tabular}[c]{@{}c@{}}Avg. \\ {[}0.8, 1.5{]}\end{tabular} & 0.2196 $\pm$ 0.4023 & 0.1926 $\pm$ 0.4666 & 0.1540 $\pm$ 0.4637 & 0.0847 $\pm$ 0.4009 & 0.3103 $\pm$ 0.3672 \\
\begin{tabular}[c]{@{}c@{}}Max \\ {[}2, 3{]}\end{tabular} & 0.3412 $\pm$ 0.5033 & 0.2699 $\pm$ 0.4855 & 0.1400 $\pm$ 0.5151 & 0.1321 $\pm$ 0.5561 & 0.6128 $\pm$ 0.3318 \\
\begin{tabular}[c]{@{}c@{}}Max \\ {[}1.8, -1{]}\end{tabular} & 0.4243 $\pm$ 0.4475 & 0.2164 $\pm$ 0.4987 & 0.0782 $\pm$ 0.5364 & 0.2012 $\pm$ 0.5129 & 0.6053 $\pm$ 0.3238
\end{tabular}
\end{table}

While the results are not as simple as in \cite{SA:21:local, schwenke2021abstracting}, the \textit{Avg. [0.8, 1.5]} threshold is on average nonetheless a good example that shows that the complexity of the data can be reduced, while the other thresholds show that keeping in mind what kind of data is getting reduced is also important.

\begin{table}[h!]
\centering
\tiny
\setlength{\tabcolsep}{3pt}
\caption{\lasa Complexity reduction compared to the Ori model complexity.}
\label{tab:complexOri}
\begin{tabular}{l|ccccc}
To Ori                                                             & SvdEn                & ApEn                 & SampEn              & CE                   & T. Shifts           \\ \hline
\begin{tabular}[c]{@{}c@{}}Avg.\\ {[}1.0,1.2{]}\end{tabular}&  \cellcolor[HTML]{C0C0C0}&  \cellcolor[HTML]{C0C0C0}&  \cellcolor[HTML]{C0C0C0}&  \cellcolor[HTML]{C0C0C0}&  \cellcolor[HTML]{C0C0C0}  \\  \hline
hl-mmm                                                       & -0.2232 $\pm$ 0.3597 & \textbf{0.2004 $\pm$ 0.3475}  & \textbf{0.4849 $\pm$ 0.2975} & -0.1804 $\pm$ 0.5775 & \textbf{0.7171 $\pm$ 0.2484} \\
hl-sss                                                       & -0.2519 $\pm$ 0.4207 & 0.0873 $\pm$ 0.3752  & 0.3898 $\pm$ 0.3222 & -0.1688 $\pm$ 0.6377 & 0.6670 $\pm$ 0.2436 \\ \hline
\begin{tabular}[c]{@{}c@{}}Avg.\\ {[}0.8,1.5{]}\end{tabular} & \cellcolor[HTML]{C0C0C0}&  \cellcolor[HTML]{C0C0C0}&  \cellcolor[HTML]{C0C0C0}&  \cellcolor[HTML]{C0C0C0}&  \cellcolor[HTML]{C0C0C0}  \\  \hline
hl-mmm                                                       & -0.0766 $\pm$ 0.4164 & 0.3044 $\pm$ 0.4013  & 0.3911 $\pm$ 0.5135 & -0.1357 $\pm$ 0.7047 & 0.8321 $\pm$ 0.1508 \\
hl-sss                                                       & \textbf{0.1770 $\pm$ 0.3333}  & \textbf{0.7106 $\pm$ 0.3117}  & \textbf{0.7687 $\pm$ 0.2822} & \textbf{0.0176 $\pm$ 0.6061}  & \textbf{0.9629 $\pm$ 0.0527} \\ \hline
\begin{tabular}[c]{@{}c@{}}Max.\\ {[}2,3{]}\end{tabular}     & \cellcolor[HTML]{C0C0C0}&  \cellcolor[HTML]{C0C0C0}&  \cellcolor[HTML]{C0C0C0}&  \cellcolor[HTML]{C0C0C0}&  \cellcolor[HTML]{C0C0C0}  \\  \hline
hl-mmm                                                       & \textbf{-0.2736 $\pm$ 0.4730} &\textbf{ 0.1541 $\pm$ 0.5552}  & \textbf{0.4739 $\pm$ 0.2996} & -0.2832 $\pm$ 0.6488 & \textbf{0.6858 $\pm$ 0.3010} \\
hl-sss                                                       & -0.4074 $\pm$ 0.4596 & -0.1849 $\pm$ 0.6970 & 0.3048 $\pm$ 0.4068 & -0.3095 $\pm$ 0.6211 & 0.4587 $\pm$ 0.3598 \\ \hline
\begin{tabular}[c]{@{}c@{}}Max\\ {[}1.8,-1{]}\end{tabular}   & \cellcolor[HTML]{C0C0C0}&  \cellcolor[HTML]{C0C0C0}&  \cellcolor[HTML]{C0C0C0}&  \cellcolor[HTML]{C0C0C0}&  \cellcolor[HTML]{C0C0C0}  \\  \hline
hl-mmm                                                       & \textbf{-0.2472 $\pm$ 0.4711} & \textbf{0.1142 $\pm$ 0.4926}  & \textbf{0.3715 $\pm$ 0.3775} & -0.2287 $\pm$ 0.6580 & \textbf{0.7073 $\pm$ 0.2768} \\
hl-sss                                                       & -0.3989 $\pm$ 0.4697 & -0.1785 $\pm$ 0.6909 & 0.3035 $\pm$ 0.4032 & -0.3022 $\pm$ 0.6286 & 0.4670 $\pm$ 0.3605
\end{tabular}
\end{table}

\begin{table}[h!]
\centering
\tiny
\setlength{\tabcolsep}{4pt}
\caption{\lasa Complexity reduction compared to the SAX model complexity.}
\label{tab:complexSax}
\begin{tabular}{l|ccccc}
To SAX                                                             & SvdEn               & ApEn                & SampEn               & CE                   & T. Shifts           \\ \hline
\begin{tabular}[c]{@{}c@{}}Avg.\\ {[}1.0,1.2{]}\end{tabular} &  \cellcolor[HTML]{C0C0C0}&  \cellcolor[HTML]{C0C0C0}&  \cellcolor[HTML]{C0C0C0}&  \cellcolor[HTML]{C0C0C0}&  \cellcolor[HTML]{C0C0C0}  \\  \hline               
hl-mmm                                                       & 0.1146 $\pm$ 0.1821 & \textbf{0.2379 $\pm$ 0.2568} & \textbf{0.0828 $\pm$ 0.7556}  & 0.0525 $\pm$ 0.3643  & \textbf{0.4325 $\pm$ 0.2433 }\\
hl-sss                                                       & 0.1061 $\pm$ 0.1455 & 0.1590 $\pm$ 0.2049 & -0.0958 $\pm$ 1.2945 & 0.0805 $\pm$ 0.2965  & 0.3610 $\pm$ 0.1900 \\ \hline
\begin{tabular}[c]{@{}c@{}}Avg.\\ {[}0.8,1.5{]}\end{tabular} &  \cellcolor[HTML]{C0C0C0}&  \cellcolor[HTML]{C0C0C0}&  \cellcolor[HTML]{C0C0C0}&  \cellcolor[HTML]{C0C0C0}&  \cellcolor[HTML]{C0C0C0}  \\  \hline
hl-mmm                                                       & 0.2324 $\pm$ 0.2020 & 0.3459 $\pm$ 0.3384 & -0.1056 $\pm$ 1.5060 & 0.1224 $\pm$ 0.4006  & 0.6747 $\pm$ 0.2155 \\
hl-sss                                                       & \textbf{0.4170 $\pm$ 0.1883} & \textbf{0.7226 $\pm$ 0.3103} & \textbf{0.5553 $\pm$ 0.7638}  & \textbf{0.2562 $\pm$ 0.3546}  & \textbf{0.9172 $\pm$ 0.1153} \\ \hline
\begin{tabular}[c]{@{}c@{}}Max.\\ {[}2,3{]}\end{tabular}     &   \cellcolor[HTML]{C0C0C0}&  \cellcolor[HTML]{C0C0C0}&  \cellcolor[HTML]{C0C0C0}&  \cellcolor[HTML]{C0C0C0}&  \cellcolor[HTML]{C0C0C0}  \\  \hline
hl-mmm                                                       & \textbf{0.1006 $\pm$ 0.1747} & \textbf{0.2449 $\pm$ 0.2778} & \textbf{0.0653 $\pm$ 0.7301}  & -0.0127 $\pm$ 0.3343 & \textbf{0.3895 $\pm$ 0.3525} \\
hl-sss                                                       & 0.0112 $\pm$ 0.0473 & 0.0146 $\pm$ 0.0912 & -0.0785 $\pm$ 0.7157 & -0.0004 $\pm$ 0.0626 & 0.0346 $\pm$ 0.1309 \\ \hline
\begin{tabular}[c]{@{}c@{}}Max\\ {[}1.8,-1{]}\end{tabular}   & \cellcolor[HTML]{C0C0C0}&  \cellcolor[HTML]{C0C0C0}&  \cellcolor[HTML]{C0C0C0}&  \cellcolor[HTML]{C0C0C0}&  \cellcolor[HTML]{C0C0C0}  \\  \hline
hl-mmm                                                       & \textbf{0.1233 $\pm$ 0.1686} & \textbf{0.1726 $\pm$ 0.2631} & -0.0757 $\pm$ 0.7022 & 0.0499 $\pm$ 0.3064  & \textbf{0.4199 $\pm$ 0.3448} \\
hl-sss                                                       & 0.0203 $\pm$ 0.0761 & 0.0027 $\pm$ 0.1199 & -0.0933 $\pm$ 0.6324 & 0.0131 $\pm$ 0.0707  & 0.0494 $\pm$ 0.1639
\end{tabular}
\end{table}

\subsubsection{\lasa Shapelets}
The idea of \lasas was that the more simple shapes of the \lasa abstracted input can further simplify the found Shapelets. However, the relative accuracy for the good models suffers very strongly from the \lasas abstraction, even for small reductions (see for details appendix Table \ref{tabA:lasaShapAcc}). To us, it is not fully clear why this is the case, \eg it could be due to a too small calculation restraint. While, the more powerful Transformer model can handle the re-training on the abstracted data with a similar performance, shows that the leftover information in the abstraction is in a format that is harder to handle for the Shapelets algorithm, while maintaining the information needed for a Transformer model.


\subsection{Global Coherence Representation Results}
In this section, we provide the results for the \gcr method from Section \ref{sec:gcr}. Additional information can be found in the appendix Tables \ref{tabA:gcrPerfGood}-\ref{tabA:gcrSaxFitCount}.

\subsubsection{Performance} 

We start by analysing each \gcr method's accuracy, for each \laam combination, only for the good performing models, as shown in Table \ref{tab:gcrKinds}. It can be seen that on average the sum \fcam performed the best with only an average accuracy reduction of around $5\%$. This and other average results are not much influenced by the \laam combination. However, the standard deviation of all \gcr variations is relatively high, \ie the sum \fcam is not always the best \gcr. For this reason, we only consider the best performing GCR per test instance in our further analyses. Table \ref{tab:gcrModelSplit} shows this merged \gcr performance for the good and for all datasets, further separated by the different model settings. For reference, we introduce Table \ref{tab:relativeShapelets} with the Shapelets performance relative to the SAX accuracy. The average performance of the \gcr improved considerably, especially for the hl-ss combination. Considering the results for all datasets, the average performance is even often positive. In combination with the standard deviation, this means that a \gcr sometimes performs better than the original model and for other classification problems the \gcr struggles. This is in line with our findings in \cite{SA:21:global}, where the \gcr can outperform the original model, if the original model does not converge well. 
The models with 5s-5l-6l performed on average the best, even outperforming the Shapelets for the good models and provide comparable results considering all datasets. Overall higher layer, symbol and header counts seem to have a positive effect on the accuracy. This effect is slightly more irregular for the good models, but this could be due to the irregular model count. A higher symbol count always leads to a considerable increase in computation time, thus it would be interesting for future research to see, what the positive limit of the other model parameters is. Even more so, considering that in our experience, too many layers leads to worse neural network models; This is also indicated by the results from \cite{wen2022transformers}, who found that lower layer counts are more preferable for time series data.

\begin{table}[h!]
\centering
\footnotesize
\setlength{\tabcolsep}{5pt}
\caption{Relative \gcr performance of the different \gcr methods for each \laam combination on the good models (accuracy $>75\%$) as compared to the SAX models.}
\begin{tabular}{l|cccc}
Good & hl-mm & hl-ms & hl-sm & hl-ss \\ \hline
\fcam & \cellcolor[HTML]{C0C0C0} & \cellcolor[HTML]{C0C0C0} & \cellcolor[HTML]{C0C0C0} & \cellcolor[HTML]{C0C0C0} \\ \hline
Sum &{-0.0497 $\pm$ 0.1340} & {-0.0484 $\pm$ 0.1357} & {-0.0442 $\pm$ 0.1361} & {-0.0434 $\pm$ 0.1360} \\
r. Avg. & -0.1619 $\pm$ 0.2590 & -0.1452 $\pm$ 0.2575 & -0.1293 $\pm$ 0.2603 & -0.1219 $\pm$ 0.2531 \\ \hline
\gtm & \cellcolor[HTML]{C0C0C0} & \cellcolor[HTML]{C0C0C0} & \cellcolor[HTML]{C0C0C0} & \cellcolor[HTML]{C0C0C0} \\ \hline
Max of sum & -0.1141 $\pm$ 0.1791 & -0.1056 $\pm$ 0.1792 & -0.1016 $\pm$ 0.1787 & -0.0970 $\pm$ 0.1782 \\
Max of r. avg. & -0.1556 $\pm$ 0.2137 & -0.1412 $\pm$ 0.2108 & -0.1322 $\pm$ 0.2125 & -0.1247 $\pm$ 0.2066 \\
Avg. of sum & -0.1046 $\pm$ 0.1794 & -0.0991 $\pm$ 0.1794 & -0.0931 $\pm$ 0.1786 & -0.0939 $\pm$ 0.1791 \\
Avg. of r. avg. & -0.1507 $\pm$ 0.2241 & -0.1390 $\pm$ 0.2186 & -0.1173 $\pm$ 0.2015 & -0.1109 $\pm$ 0.1944 \\
Med. of sum & -0.1020 $\pm$ 0.1793 & -0.0987 $\pm$ 0.1793 & -0.0922 $\pm$ 0.1793 & -0.0926 $\pm$ 0.1793 \\
Med. of r. avg. & -0.1528 $\pm$ 0.1918 & -0.1465 $\pm$ 0.1914 & -0.1391 $\pm$ 0.1962 & -0.1362 $\pm$ 0.1926
\end{tabular}
\label{tab:gcrKinds}
\end{table}

\begin{table}[h!]
\centering
\footnotesize
\setlength{\tabcolsep}{5pt}
\caption{Relative \gcr accuracy split between the different model configurations and \laam combination on the good models (accuracy $>75\%$) as compared to the SAX models.}
\begin{tabular}{l|cccc}
Good & hl-mm & hl-ms & hl-sm & hl-ss \\ \hline
All conf.& -0.0337 $\pm$ 0.1300 & -0.0305 $\pm$ 0.1320 & -0.0209 $\pm$ 0.1339 & -0.0172 $\pm$ 0.1340 \\
3s-2l-8h & -0.0715 $\pm$ 0.1296 & -0.0782 $\pm$ 0.1404 & -0.0720 $\pm$ 0.1569 & -0.0786 $\pm$ 0.1541 \\
3s-2l-16 & -0.0142 $\pm$ 0.1234 & -0.0191 $\pm$ 0.1307 & -0.0053 $\pm$ 0.1398 & -0.0080 $\pm$ 0.1434 \\
3s-5l-6h & -0.0320 $\pm$ 0.2308 & -0.0325 $\pm$ 0.2342 & -0.0374 $\pm$ 0.2369 & -0.0316 $\pm$ 0.2406 \\
5s-2l-8h & -0.0455 $\pm$ 0.1334 & -0.0395 $\pm$ 0.1324 & -0.0284 $\pm$ 0.1328 & -0.0233 $\pm$ 0.1311 \\
5s-2l-16 & -0.0346 $\pm$ 0.1017 & -0.0269 $\pm$ 0.0984 & -0.0102 $\pm$ 0.0845 & -0.0040 $\pm$ 0.0829 \\
5s-5l-6h & \textbf{-0.0036 $\pm$ 0.0800} & \textbf{0.0050 $\pm$ 0.0752 }& \textbf{0.0097 $\pm$ 0.0715} & \textbf{0.0196 $\pm$ 0.0647} \\ \hline
All &  hl-mm & hl-ms & hl-sm & hl-ss \\ \hline
All conf.& 0.0551 $\pm$ 0.1757 & 0.0554 $\pm$ 0.1757 & 0.0613 $\pm$ 0.1739 & 0.0630 $\pm$ 0.1734 \\
3s-2l-8h & 0.0380 $\pm$ 0.1701 & 0.0330 $\pm$ 0.1697 & 0.0419 $\pm$ 0.1714 & 0.0394 $\pm$ 0.1728 \\
3s-2l-16 & 0.0467 $\pm$ 0.1755 & 0.0424 $\pm$ 0.1773 & 0.0472 $\pm$ 0.1767 & 0.0483 $\pm$ 0.1760 \\
3s-5l-6h & 0.0691 $\pm$ 0.1860 & 0.0651 $\pm$ 0.1895 & 0.0677 $\pm$ 0.1872 & 0.0732 $\pm$ 0.1893 \\
5s-2l-8h & 0.0459 $\pm$ 0.1725 & 0.0505 $\pm$ 0.1718 & 0.0583 $\pm$ 0.1697 & 0.0600 $\pm$ 0.1688 \\
5s-2l-16 & 0.0568 $\pm$ 0.1673 & 0.0626 $\pm$ 0.1665 & 0.0697 $\pm$ 0.1610 & 0.0712 $\pm$ 0.1590 \\
5s-5l-6h & \textbf{0.0856 $\pm$ 0.1821} & \textbf{0.0910 $\pm$ 0.1756} & \textbf{0.0935 $\pm$ 0.1746} & \textbf{0.0981 $\pm$ 0.1698}
\end{tabular}

\label{tab:gcrModelSplit}
\end{table}

\begin{table}[h!]
\centering
\footnotesize
\setlength{\tabcolsep}{4pt}
\caption{Shapelets performance relative to SAX accuracy.}

\begin{tabular}{l|cc|cc}
Baseline & All & All slen 0.3 & Good &  Good slen 0.3 \\ \hline
Ori Shapelets & \textbf{0.1697 $\pm$ 0.1931} & 0.1636 $\pm$ 0.1873 & 0.0067 $\pm$ 0.1412 & \textbf{0.0084 $\pm$ 0.1348}\\
SAX Shapelets & 0.0876 $\pm$ 0.1808  & 0.0984 $\pm$ 0.1661 & -0.0441 $\pm$ 0.1485 & -0.0258 $\pm$ 0.1345
\end{tabular}

\label{tab:relativeShapelets}
\end{table}

\subsubsection{\gcr Selection}
As we already found out in \cite{SA:21:global}, selecting the correct type of \gcr depends heavily on the type of classification problem. The analyses in this work suggest the same; while the \fcam type seems to be overall the most successful, the best \gcr seems to be bound to the classification problem itself. More details on \gcr selection can be found in appendix Table \ref{tabA:grcCount}. Also, the different \laam combinations seem to influence the best performing \gcr type, hence selecting the right \gcr is not only influenced by the dataset itself.

\subsubsection{Explainability} 
In the following section, we provide and analyse the results on the explainability metrics of the \gcr models.
\paragraph{Certainty}
To see how well the certainty of the \gcr model works, we show the average certainty scores for the good models in Table \ref{tab:certTest}. It can be seen that the average accuracy rises regardless of the combination per certainty step. For the hl-ss \laam combination, the accuracy even always rises per step in $76.31\%$ of all train and in $60.53\%$ of all test cases for the good models. This trend can also be seen when looking at all models, however, the number of times the accuracy only rises for each certainty step for the hl-ss combination can only be seen in $61.76\%$ of the train and $41.17\%$ of the test cases. Nevertheless, even though the performance of more certain classification rises relatively a lot, the Attention or the model still seem to have some bigger learned misconceptions, because the accuracy for $10\%$ certainty did not reach $100\%$. Considering that the less good models have less well distributed Attention values and the fact that we only use Attention without the surrounding model weights, this not perfect certainty could be expected. Overall, the \gcr in combination with the Transformer Attention values seems to build into a relatively well working certainty score.

\begin{table}[h!]
\centering
\footnotesize
\caption{Effect of the \gcr certainty on the test accuracy, on the good models.}
\begin{tabular}{l|cccc}
Good & hl-mm & hl-ms & hl-sm & hl-ss \\ \hline
Test Cert. 100 & 0.8116 $\pm$ 0.1399 & 0.8148 $\pm$ 0.1454 & 0.8244 $\pm$ 0.1472 & 0.8281 $\pm$ 0.1488 \\
Test Cert. 80 & 0.8248 $\pm$ 0.1411 & 0.8299 $\pm$ 0.1459 & 0.8403 $\pm$ 0.1444 & 0.8439 $\pm$ 0.1472 \\
Test Cert. 50 & 0.8368 $\pm$ 0.1339 & 0.8429 $\pm$ 0.1347 & 0.8543 $\pm$ 0.1313 & 0.8612 $\pm$ 0.1351 \\
Test Cert. 20 & 0.8577 $\pm$ 0.1463 & 0.8584 $\pm$ 0.1470 & 0.8734 $\pm$ 0.1343 & 0.8872 $\pm$ 0.1303 \\
Test Cert.10 & 0.8692 $\pm$ 0.1545 & 0.8722 $\pm$ 0.1556 & 0.8885 $\pm$ 0.1342 & 0.9086 $\pm$ 0.1208
\end{tabular}
\label{tab:certTest}
\end{table}

\paragraph{Model Fidelity}
\label{sec:gcrModFidelity}

In the performance section of the \gcr, we verified the \gcr's classification abilities, showing that the \gcr can perform quite similar to the SAX model baseline or for bad performing models sometimes even better, while only using the internal model Attention values. Looking at the model fidelity in Table \ref{tab:gcrFidelity} for the good models, we can see that the fidelities of the GCR model are typically in a similar range to the Shapelet fidelity; \ie the \gcr does only partially learns in the same manner as the Transformer model. Nevertheless, the often higher test fidelity values (especially for the SAX model) with lower standard deviation show that the \gcr model uses more similar classification criteria than the Shapelets model, thus the learning approach of the \gcr approximates some core decision-making of the Transformer model. This gets more clear when looking at Table \ref{tab:GCRfidelityCert} where the overall train fidelities rise with the certainty.

A few special examples show that while the \gcr can perform well, the computational power and classification focus is not on pair with the neural network; \eg in Section \ref{sec:weaknesses} we will show a few class separating elements that can be challenging for the \gcr. Additionally, in this section, we only looked at the learning performance when trying to approximate the original labels. Later in this work in Section \ref{sec:gcrSaxModelFi}, we analyse the interpretability ability of the \gcr, when the \gcr is fit on the SAX model predictions.

\begin{table}[h!]
\centering
\footnotesize
\setlength{\tabcolsep}{2pt}
\caption{\gcr model fidelities for the good performing models, compared to the slen 0.3 Shapelet model fidelities.}
\begin{tabular}{l|cccc}
Good & \begin{tabular}[c]{@{}c@{}}Ori Train \\ Model Fidelity\end{tabular} & \begin{tabular}[c]{@{}c@{}}Ori Test \\ Model Fidelity\end{tabular} & \begin{tabular}[c]{@{}c@{}}SAX Train \\ Model Fidelity\end{tabular} & \begin{tabular}[c]{@{}c@{}}SAX Test \\ Model Fidelity\end{tabular} \\ \hline
Ori Shap. &0.8118 $\pm$ 0.2292  &  0.7223 $\pm$ 0.2193  &  0.8629 $\pm$ 0.1669   & 0.7799 $\pm$ 0.1870 \\
SAX Shap. & 0.8095 $\pm$ 0.2258  & 0.6929 $\pm$ 0.2034  & \textbf{ 0.8661 $\pm$ 0.1651}  &  0.7695 $\pm$ 0.1916  \\\hline
All conf.& 0.7497 $\pm$ 0.1958 & 0.7047 $\pm$ 0.1939 & 0.8259 $\pm$ 0.1536 & 0.7900 $\pm$ 0.1586 \\
5s-5l-6h & \textbf{0.8247 $\pm$ 0.1570} & \textbf{0.8118 $\pm$ 0.1099} & 0.8596 $\pm$ 0.0940 & \textbf{0.8381 $\pm$ 0.0635} \\

\end{tabular}

\label{tab:gcrFidelity}
\end{table}

\begin{table}[!htb]
\centering
\footnotesize
\caption{\gcr model fidelity based on the most certain instances for the good models.}
\begin{tabular}{c|cc}
Good & \begin{tabular}[c]{@{}c@{}}SAX Train\\ Model Fidelity\end{tabular} & \begin{tabular}[c]{@{}c@{}}SAX Test\\ Model Fidelity\end{tabular} \\ \hline
Cert. 100 & 0.8259 $\pm$ 0.1536 & 0.7900 $\pm$ 0.1586 \\
Cert. 80 & 0.8382 $\pm$ 0.1450 & 0.8111 $\pm$ 0.1588 \\
Cert. 50 & 0.8589 $\pm$ 0.1412 & 0.8403 $\pm$ 0.1560 \\
Cert. 20 & 0.8847 $\pm$ 0.1621 & 0.8773 $\pm$ 0.1690 \\
Cert. 10 & 0.8897 $\pm$ 0.1756 & 0.8898 $\pm$ 0.1798
\end{tabular}
\label{tab:GCRfidelityCert}
\end{table}

\subsubsection{Penalty \gcr}
The Penalty-\gcr is a subversion of the original \gcr. Overall the average accuracy per \pgcr (counting and entropy) gets a little worse, but by using a selective approach including the normal \gcr the results improve; \ie, we selected the best performing \gcr model per dataset. The performance results for this selective approach can be seen in Table \ref{tab:pgcrPerformanceComp}. Using a selective approach for the model fidelity, the model fidelity also improves, yet on average both \pgcrs have a better model fidelity, as seen in Table \ref{tab:fidelityPGCR}; \ie they better approximate the decision-making of the Transformer model. This improvement using the selective approach is also seeable for the certainty, as shown in Table \ref{tab:pgrcTestCert}. Here, even the accuracy rises per certainty step increases compared to Table \ref{tab:certTest}. Due to the adaptive nature of the penalty, the performance can be even further fine-tuned to match the current dataset for an even bigger improvement. This however would take too long for each dataset, which is why we used just one calculation per \pgcr here.

\begin{table}[h!]
\centering
\footnotesize
\setlength{\tabcolsep}{2pt}
\caption{Accuracy performance comparison between different \gcr variations and the Shapelets baseline. Also including the best model configuration: slen 0.3 for the Shapelets and 5s-5h-6l for \gcr.}
\begin{tabular}{l|cc|cc}
 & \multicolumn{1}{l}{All Configs} & \multicolumn{1}{l|}{} & \multicolumn{1}{l}{Best Config} & \multicolumn{1}{l}{} \\
Accuracy & All & Good & All & Good \\ \hline
Ori Shapelets & \textbf{0.1697 $\pm$ 0.1931} & \textbf{0.0067 $\pm$ 0.1412} & \textbf{0.1636 $\pm$ 0.1873} & 0.0084 $\pm$ 0.1348\\
SAX Shapelets & 0.0876 $\pm$ 0.1808 & -0.0441 $\pm$ 0.1485 & 0.0984 $\pm$ 0.1661 & -0.0258 $\pm$ 0.1345 \\ \hline
\gcr & 0.0587 $\pm$ 0.1747 & -0.0256 $\pm$ 0.1327 & 0.0920 $\pm$ 0.1756 & 0.0077 $\pm$ 0.0735 \\
Counting \gcr & 0.0479 $\pm$ 0.1599 & -0.0318 $\pm$ 0.1270 & 0.0881 $\pm$ 0.1732 & 0.0131 $\pm$ 0.0672 \\
Entropy \gcr & 0.0575 $\pm$ 0.1585 & -0.0146 $\pm$ 0.1206 & 0.0758 $\pm$ 0.1721 & -0.0138 $\pm$ 0.0932 \\
All \gcrs& 0.0806 $\pm$ 0.1634 & -0.0029 $\pm$ 0.1190 & 0.1069 $\pm$ 0.1716 & \textbf{0.0223 $\pm$ 0.0679}
\end{tabular}
\label{tab:pgcrPerformanceComp}
\end{table}

\begin{table}[h!]
\centering
\footnotesize
\setlength{\tabcolsep}{2pt}
\caption{Model fidelity for the good performing models differentiated between the \gcr variations.}
\begin{tabular}{l|cccc}
Good & \begin{tabular}[c]{@{}c@{}}Ori Train\\ Model Fidelity\end{tabular} & \begin{tabular}[c]{@{}c@{}}Ori Test\\ Model Fidelity\end{tabular} & \begin{tabular}[c]{@{}c@{}}SAX Train\\ Model Fidelity\end{tabular} & \begin{tabular}[c]{@{}c@{}}SAX Test\\ Model Fidelity\end{tabular} \\ \hline
Ori Shap. slen 0.3 &\textbf{0.8118 $\pm$ 0.2292}  &  0.7223 $\pm$ 0.2193  &  0.8629 $\pm$ 0.1669   & 0.7799 $\pm$ 0.1870 \\
SAX Shap. slen 0.3 & 0.8095 $\pm$ 0.2258  & 0.6929 $\pm$ 0.2034  &  0.8661 $\pm$ 0.1651  &  0.7695 $\pm$ 0.1916  \\ \hline
\gcr & \multicolumn{1}{c}{0.7497 $\pm$ 0.1958} & \multicolumn{1}{c}{0.7047 $\pm$ 0.1939} & \multicolumn{1}{c}{0.8259 $\pm$ 0.1536} & \multicolumn{1}{c}{0.7900 $\pm$ 0.1586} \\
Counting \gcr & 0.7638 $\pm$ 0.1674 & 0.7257 $\pm$ 0.1703 & 0.8449 $\pm$ 0.0941 & 0.8161 $\pm$ 0.1101 \\
Entropy \gcr & 0.7728 $\pm$ 0.1769 & 0.7326 $\pm$ 0.1784 & 0.8556 $\pm$ 0.1071 & 0.8241 $\pm$ 0.1154 \\
All \gcrs & 0.7915 $\pm$ 0.1717 & \textbf{0.7518 $\pm$ 0.1720} & \textbf{0.8763 $\pm$ 0.0920} & \textbf{0.8454 $\pm$ 0.1070}
\end{tabular}
\label{tab:fidelityPGCR}
\end{table}

\begin{table}[h!]
\centering
\footnotesize
\caption{Test certainty using a selective approach, including \gcr, Counting \pgcr and Entropy \pgcr.}
\begin{tabular}{l|cccc}
Good & hl-mm & hl-ms & hl-sm & hl-ss \\ \hline
Test Cert. 100 & 0.8363 $\pm$ 0.1205 & 0.8392 $\pm$ 0.1249 & 0.8461 $\pm$ 0.1301 & 0.8478 $\pm$ 0.1305 \\
Test Cert. 80 & 0.8593 $\pm$ 0.1199 & 0.8640 $\pm$ 0.1224 & 0.8683 $\pm$ 0.1259 & 0.8712 $\pm$ 0.1285 \\
Test Cert. 50 & 0.8769 $\pm$ 0.1105 & 0.8845 $\pm$ 0.1131 & 0.8922 $\pm$ 0.1176 & 0.8940 $\pm$ 0.1210 \\
Test Cert. 20 & 0.9069 $\pm$ 0.1004 & 0.9239 $\pm$ 0.0973 & 0.9248 $\pm$ 0.0949 & 0.9270 $\pm$ 0.0957 \\
Test Cert.10 & 0.9331 $\pm$ 0.0888 & 0.9444 $\pm$ 0.0887 & 0.9435 $\pm$ 0.0827 & 0.9470 $\pm$ 0.0750
\end{tabular}
\label{tab:pgrcTestCert}
\end{table}

The intention of using multiple \gcr methods is that it is unclear how to approximate the Attention interpretation best, so we use multiple possible interpretation techniques, each with a different focus.

\subsubsection{Threshold \gcr}
The idea of the Threshold \gcr (\tgcr) is that we can remove lower Attention values in a similar sense as in \lasa, to reduce the overall complexity of the model. It also serves to analyse the infidelity. Table \ref{tab:tgcrPerf} shows the accuracy development per \tgcr, where \textit{All \tgcr} is again a selective approach over all threshold values of the \tgcrs. The accuracy performance of each \tgcr shows that it is possible to remove lower Attention values, while keeping a similar \gcr performance. Sometimes the accuracy even increases, maybe due to noise in the lower Attention space; leading to an overall slightly better performance in the \textit{All \gcrs} row.

\begin{table}[h!]
\centering
\footnotesize
\setlength{\tabcolsep}{3pt}
\caption{Accuracy performance comparison between different \tgcr variations and the Shapelets baseline. Also including the best model configuration: slen 0.3 for the Shapelets and 5s-5h-6l for \gcr.}
\begin{tabular}{l|cc|cc}
 & \multicolumn{1}{l}{All Configs} & \multicolumn{1}{l|}{} & \multicolumn{1}{l}{Best Config} & \multicolumn{1}{l}{} \\
Accuracy & All & Good & All & Good \\ \hline
Ori Shapelets & \textbf{0.1697 $\pm$ 0.1931} & \textbf{0.0067 $\pm$ 0.1412} & \textbf{0.1636 $\pm$ 0.1873} & 0.0084 $\pm$ 0.1348\\
SAX Shapelets & 0.0876 $\pm$ 0.1808 & -0.0441 $\pm$ 0.1485 & 0.0984 $\pm$ 0.1661 & -0.0258 $\pm$ 0.1345 \\ \hline
\gcr t1.0 & 0.0472 $\pm$ 0.1739 & -0.0505 $\pm$ 0.1344 & 0.0788 $\pm$ 0.1874 & -0.0195 $\pm$ 0.0963 \\
\gcr t1.3 & 0.0585 $\pm$ 0.1741 & -0.0253 $\pm$ 0.1287 & 0.0901 $\pm$ 0.1774 & 0.0035 $\pm$ 0.0749 \\
\gcr t1.6 & 0.0585 $\pm$ 0.1748 & -0.0249 $\pm$ 0.1303 & 0.0912 $\pm$ 0.1765 & 0.0070 $\pm$ 0.0738 \\
All \tgcrs & 0.0718 $\pm$ 0.1660 & -0.0139 $\pm$ 0.1225 & 0.0965 $\pm$ 0.1754 & 0.0118 $\pm$ 0.0758 \\
All \gcrs & 0.0858 $\pm$ 0.1586 & 0.0023 $\pm$ 0.1138 & 0.1077 $\pm$ 0.1711 & \textbf{0.0229 $\pm$ 0.0679}
\end{tabular}

\label{tab:tgcrPerf}
\end{table}

Because we only added Attention higher than a specific threshold (infidelity), we showed in the scope of this work that most of the relevant classification information is included in the higher Attention values. We infer this from the fact that the overall performance did not decrease considerably and sometimes even increased. However, fluctuating results indicate that Attention holds only partially relevancy information. This can be even further observed in the \textit{All \tgcr} column, which shows that different thresholds classify different trials correctly; This suggests that low Attention values can still contain important classification information and that the level of information in Attention is not necessarily linear.

\subsection{\gcr for Interpretation}
\label{sec:gcrSaxModelFi}

In Section \ref{sec:gcrModFidelity} we showed the classification abilities of the \gcr if trained on the target labels; in this section we aim to approximate the Transformer model results by selecting the model predictions as target labels for the \gcr. To make a fair comparison, we do the same for the Shapelets model. Table \ref{tab:saxOptGCRFi} shows the results for different \gcr options and the Shapelets baseline. We observe that while the Shapelets model can better adapt to the training data, the \gcr model has an overall better approximation of how the Transformer model works; deducted by the test model fidelity results and the lower standard deviation. Looking at the certainties in Table \ref{tab:saxModelFitGCRCert}, we see again that the core understanding of the \gcr model is relatively high. We believe the results show that the \gcr has a subset of similar decision steps and thus can to a certain degree approximate an interpretation for the Transformer model.

\begin{table}[h!]
\centering
\footnotesize
\setlength{\tabcolsep}{3pt}
\caption{\gcr for interpretation compared to Shapelets for interpretation on SAX model fidelity.}
\begin{tabular}{l|cc|cc}
 & \multicolumn{1}{l}{All Models} & \multicolumn{1}{l|}{} & \multicolumn{1}{l}{Good Models} & \multicolumn{1}{l}{} \\
 & \begin{tabular}[c]{@{}c@{}}SAX Train\\ Model Fidelity\end{tabular} & \begin{tabular}[c]{@{}c@{}}SAX Test\\ Model Fidelity\end{tabular} & \begin{tabular}[c]{@{}c@{}}SAX Train\\ Model Fidelity\end{tabular} & \begin{tabular}[c]{@{}c@{}}SAX Test\\ Model Fidelity\end{tabular} \\ \hline
Ori Shapelets & \textbf{0.9572 $\pm$ 0.1990} & 0.7764 $\pm$ 0.1959 & 0.9972 $\pm$ 0.0172 & 0.8373 $\pm$ 0.1587 \\ 
SAX Shapelets & 0.9390 $\pm$ 0.2021 & 0.7868 $\pm$ 0.2031 & 0.9711 $\pm$ 0.0809 & 0.8212 $\pm$ 0.1705 \\
\begin{tabular}[c]{@{}l@{}}SAX Shapelets \\ slen 0.3\end{tabular} & 0.9523 $\pm$ 0.1989 & 0.8034 $\pm$ 0.1988 & \textbf{0.9975 $\pm$ 0.0076} & 0.8472 $\pm$ 0.1496 \\ \hline
\gcr & 0.9323 $\pm$ 0.0979 & 0.8618 $\pm$ 0.1102 & 0.9255 $\pm$ 0.1031 & 0.8614 $\pm$ 0.1193 \\
All \gcrs & 0.9456 $\pm$ 0.0804 & 0.8804 $\pm$ 0.0926 & 0.9388 $\pm$ 0.0855 & 0.8786 $\pm$ 0.1083 \\
\begin{tabular}[c]{@{}l@{}}All \gcrs\\ 3s-5l-6h\end{tabular} & 0.9544 $\pm$ 0.0767 & \textbf{0.8996 $\pm$ 0.0904} & 0.8856 $\pm$ 0.1496 & 0.7839 $\pm$ 0.1608 \\
\begin{tabular}[c]{@{}l@{}}All \gcrs\\ 5s-5l-6h\end{tabular} & 0.9523 $\pm$ 0.0770 & 0.8813 $\pm$ 0.0835 & 0.9595 $\pm$ 0.0382 & \textbf{0.8943 $\pm$ 0.0591} 
\end{tabular}

\label{tab:saxOptGCRFi}
\end{table}

\begin{table}[h!]
\centering
\footnotesize
\caption{Test certainty over all \gcr variations when interpreting the SAX model results.}
\begin{tabular}{l|cccc}
All & hl-mm & hl-ms & hl-sm & hl-ss \\ \hline
Test Cert. 100 & 0.8778 $\pm$ 0.0938 & 0.8809 $\pm$ 0.0910 & 0.8804 $\pm$ 0.0929 & 0.8823 $\pm$ 0.0925 \\
Test Cert. 80 & 0.9035 $\pm$ 0.0902 & 0.9071 $\pm$ 0.0871 & 0.9077 $\pm$ 0.0906 & 0.9093 $\pm$ 0.0907 \\
Test Cert. 50 & 0.9356 $\pm$ 0.0812 & 0.9391 $\pm$ 0.0795 & 0.9406 $\pm$ 0.0808 & 0.9419 $\pm$ 0.0813 \\
Test Cert. 20 & 0.9668 $\pm$ 0.0649 & 0.9713 $\pm$ 0.0570 & 0.9721 $\pm$ 0.0569 & 0.9720 $\pm$ 0.0587 \\
Test Cert.10 & 0.9778 $\pm$ 0.0526 & 0.9808 $\pm$ 0.0462 & 0.9825 $\pm$ 0.0437 & 0.9836 $\pm$ 0.0421
\end{tabular}
\label{tab:saxModelFitGCRCert}
\end{table}

\subsection{Special Cases}
In this section, we discuss some special cases like weaknesses and potential additional options for the \gcr.
\subsubsection{Weaknesses}
\label{sec:weaknesses}
Even with all current introduced options, the \gcr still is not as powerful as a neural network, but only approximate certain abilities. Therefore, certain types of classification tasks are challenging for the \gcr. In this section, we want to introduce a few of the current \gcr weaknesses we found. 

\begin{figure}[ht!]
	\centering
	\includegraphics[width=0.98\columnwidth]{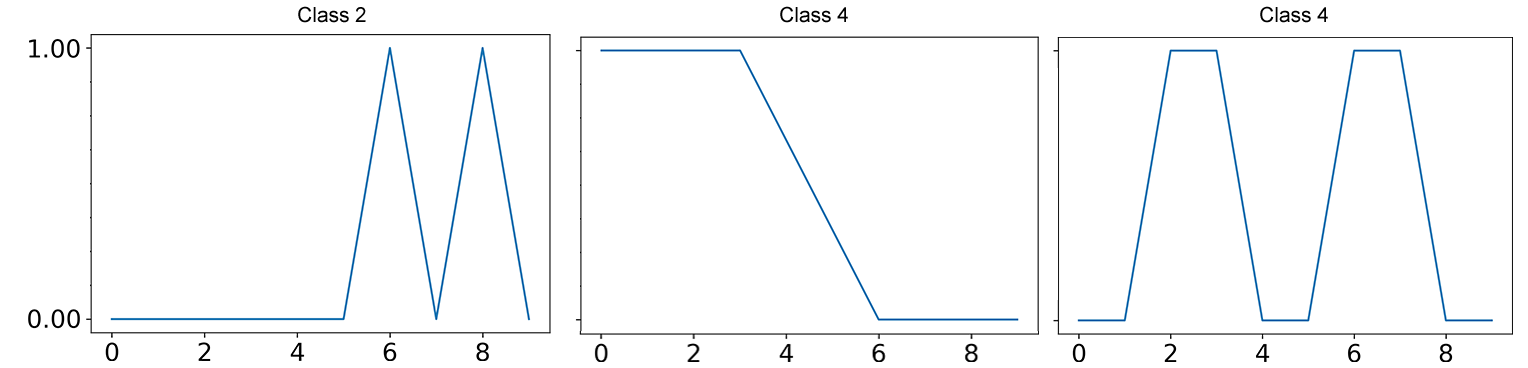}
	\caption{Example instances for the counting dataset.}
	\label{fig:countingExample}
\end{figure}

As an example, we create a simple dataset of length 10, where each position is either 0 or 1 and where the labels are the number of 1s in the sequence. Figure \ref{fig:countingExample} shows a few examples to further illustrate the dataset. We separate all possible combinations into 30/20/50\% (train/validation/test). Even though the Transformer model can easily achieve good accuracy results with over 98\% accuracy, the \gcr struggles to adapt to the given problem and on average over 5 folds only achieves an accuracy of 31\%. If all possible instances are included in the trainings set, the accuracy even drops to 25\%.

We introduce another toy dataset, that contains similar data, but the labels are binary and only show if more than four or less than five 1s exists in the sequence. The Transformer model has no problem learning this task. Looking at the \gcr, we also see an improvement in accuracy and reach up to 89.06\% over 5 folds with the initial split. The performance decreases to 79.49\% if all possible combinations are in the trainings set. Considering the reduction of the accuracy with more information, this indicates further that this form of task seems to be a challenge for which, however, an information bias can be useful.
If we only include trials with a count of exactly four or five and use all data for training, we surprisingly can reach an accuracy up to 63.42\%, indicating that even very close classes can at least be differentiated to a certain degree.

This behaviour give us some insight about current challenges for the \gcr. We would summarize them as: too similar classes, classes where the information does not lie in the sequence position or related position (thus the information can jump huge distances position wise in the sequence, \eg counting) and too large unknown areas between train and test set. This, as shown, includes counting datasets where the \gcr struggles to grasp very close classes and we theorize certain frequency based tasks could be a problem as well. Nonetheless, as long as the classes are not too detailed, as seen in the second dataset, the \gcr can still provide good results. We think those limitations can be overcome by either switching the symbol and position relation in the \gcr or by using a different classification approach; however, those adaptations still need to be researched and they would exceed the scope of this work.

\subsubsection{Debugging Global Inputs}
\begin{figure}[ht!]
	\centering
	\includegraphics[width=0.98\columnwidth]{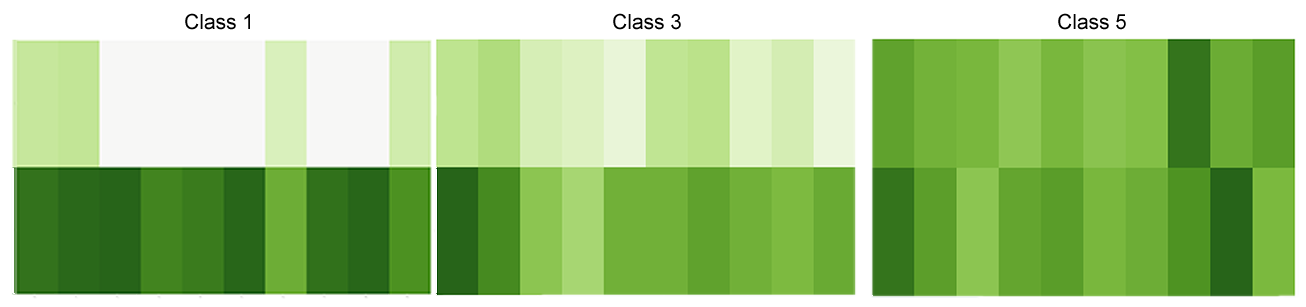}
	\caption{Example for three max \gtms, indicating multiple training data biases.}
	\label{fig:debugExample}
\end{figure}

One further usage for the \gcr is as a debugging assistant, \ie when comparing the \gcr with the expectations of a class it can provide starting points to look for possible errors; For example in Figure \ref{fig:minClassBoth} (left) we can see a low Attention phase in the top left corner, surrounded by high Attention values. This low Attention spot could indicate trials that are missing in the training set, but could be a part of the \gcr class. Hence, we can improve the training data of the model. Another, more specific example: Considering again the counting dataset with all 11 instances (number of 1s in the sequence). If we take only 30\% to train our model and visualise the max \gtm \gcr, we can see in Figure \ref{fig:debugExample} that certain areas were never added to the \gcr. For example, for class 5 we would expect that each position is equally valued, which is however not the case due to unknown data trials. This example shows that with the \gcr visualisation it is possible to compactly find and analyse missing parts in the training set which might cause trouble for the model or find parts which are not as strongly highlighted as one might expect. Hence, we conclude that the \gcr can be used as a debug assistant to improve the underlying model.

%% file: s2_related_work.tex
\section{Related Work}
\label{sec:relatedWork}
In this section, we give an overview over other research in the area to better position our methods in context.
\subsection{EXplainable AI on Time Series}

XAI on time series data is nothing new and many methods and model exists that can be used to handle the interpretability. \citep{rojat2021explainable} recently created a survey on XAI on time series data to give an overview. They gathered and compared multiple methods and differentiated them based on unified categories. We selected a few of those categories relevant for our methods to better specify our approaches and present them in Table \ref{tab:xaiposition}. Regarding the \textit{Methodology}, we have a combination of the Attention Mechanism with the SAX algorithm. As for the \textit{Scope}, we present a method for a global and a local approach. Furthermore, the \textit{Target Audience} is often an underrated aspect of a XAI method: Where the interest in the model explanation depends on the goal and understanding of the user. In our case, the local approach should help expert developers to simplify the data and thus better access it. On the other hand, the global approach has multiple detail grades, for different levels of expertise. However, due to the general complexity of time series data even simplified representations need a somewhat basic understanding, thus our approaches are rather developer focused.

\begin{table}[h!]
\centering
\caption{XAI category specification of the \lasa and \gcr method, based on \cite{rojat2021explainable}.}
\begin{tabular}{c|c|c|c|c}
\textbf{Model} &
  \textbf{\begin{tabular}[c]{@{}c@{}}Ante-hoc/\\ Post-hoc\end{tabular}} &
  \textbf{Methodology} &
  \textbf{Scope} &
  \textbf{Target Audience} \\ \hline
\lasa &
  Post-hoc &
  SAX + Attention &
  Local &
  Developer \\
\gcr &
  Post-hoc &
  SAX + Attention &
  Global &
  Developer
\end{tabular}

\label{tab:xaiposition}
\end{table}

\subsection{EXplainable AI on Transformers}

As we already hinted before, one major drawback of Transformer models --- which they share with general Deep Learning (black box) models --- are their still lacking interpretability and explainability \cite{pruthi2019learning,baan2019understanding,clark2019does}. Nevertheless, multiple approaches suggest using Attention to enhance the interpretability of the model or the underlying data. In the following sections, we briefly present how visualizations and interpretation are often presented in each application field; giving a few recent example approaches for model interpretation via Attention.

\subsubsection{Natural Language Processing}

\begin{figure}[h!]
	\centering
	\includegraphics[width=1.0\columnwidth]{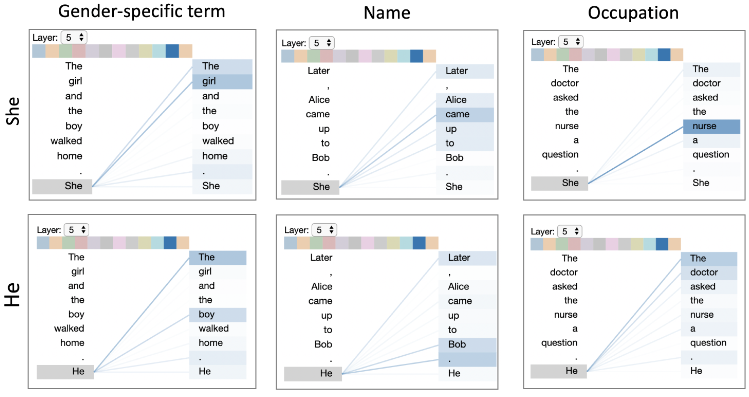}
	\caption{Attention visualization example from \citep{vig-2019-multiscale}, showing possible gender biases in the GPT-2 language model.}
	\label{fig:nlpexample}
\end{figure}

Most work up to now on Transformers was done in the context of Natural Language Processing (NLP), as their origin application area. Due to the two-dimensional relation-highlighting of Attention, multiple methods focus on visualizing the word-to-word relation in local sentences \cite{clark2019does,vig2019visualizing,vskrlj2020attviz}, showing how the meaning of the words affect each other, \eg Figure \ref{fig:nlpexample}; hence, its extracting semantic information for interpretation and explanation. For example, \citep{vig2019visualizing} and \citep{vskrlj2020attviz} provide each a visualization framework to highlight and explore word relations in different layers, based on Attention values. Besides those approaches, multiple others exist, which are summarized in \cite{bracsoveanu2020visualizing}.

\subsubsection{Computer Vision}

\begin{figure}[h!]
	\centering
	\includegraphics[width=0.43\columnwidth]{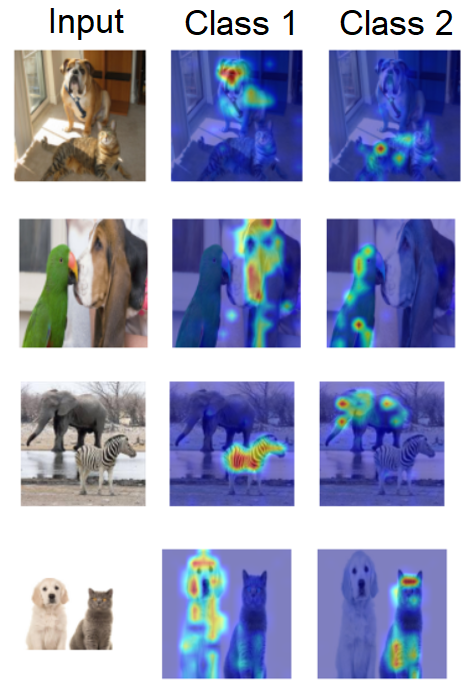}
	\caption{Example for Attention-based saliency maps using the LRP method from \citep{chefer2021transformer}.}
	\label{fig:AttentionMapDeit}
\end{figure}

Computer Vision (CV) often works with saliency maps or areas of interest to show the most important regions for classification \cite{buhrmester2019analysis} (\eg Figure \ref{fig:AttentionMapDeit}). This synergizes well with images and how humans focus on certain objects. However, it was shown that such representations for non-Transformer models can be misleading and are susceptible to adversarial attacks \cite{eykholt2018robust}. Transformers seem to suffer from a similar vulnerability \cite{mahmood2021robustness}, but can be more robust in some low-level noise cases \cite{shao2021adversarial}.

Considering the format of Transformer Attention, using Attention to construct saliency maps seems suitable as a mean to highlight important areas in images. Recent image processing Transformer models \cite{dosovitskiy2020image,carion2020end,caron2021emerging} apply such Attention-based saliency maps to highlight the most important feature areas to enable local interpretations. An example is illustrated in Figure \ref{fig:AttentionMapDeit}. 

\subsubsection{Time Series Data} 

Due to the memory complexity of Transformers and the unintuitivity of time series data in general, most Attention-based interpretation methods were to our best knowledge never applied on time series data inside a scientific publication. Even though, some model agonistic or general Attention-based approaches from \citep{bracsoveanu2020visualizing} could be theoretically used on time series data. The results from \cite{ismail2020benchmarking} suggest, that this is not that simple. Two examples \citep{lim2019temporal} and \citep{li2019enhancing}, which successfully applied Transformers on time series data, just used the average Attention for saliency maps to highlight locally high attended data points. For this lack of methods, we introduced our two XAI methods in this paper, to better utilize and understand Attention in the Transformer architecture.

\subsubsection{Attention Rollout}

How to aggregate and handle all Attention heads in different layers is a very recently emerged research topic. Work from \citep{abnar2020quantifying} aimed to detangle the Attention flow by setting the inputs and connections between the Attention layers into relation. Very recent research from \cite{chefer2021transformer} went a step further and detangled the information flow Attention by extending the Layer-Wise Relevance Propagation (LRP) algorithm. This even further improved the interpretability and highlighting ability from visual Transformers \cite{dosovitskiy2020image}, including a support for class related highlighting, as shown in Figure \ref{fig:AttentionMapDeit}.

While we look into more simple Attention aggregations, our analyses show how easy visible elements in Attention can already lead to successful local abstraction and global class representation, under different simple Attention aggregation strategies. Due to the generalizability of the approaches, aggregation strategies like from \cite{chefer2021transformer} can also be used to improve the results from our framework. This is however set for future work.

%% file: s6_discussion.tex
\section{Discussion and Outlook}
\label{sec:discussion}
In this section we will discuss the results of this work, suggest further interpretations and give directions for future research.
\subsection{\lasa}
In Table \ref{tab:lasaAR} we showed that \lasa can reduce on average over 96\%  of all datapoints in the input sequences with only loosing 12.64\% of accuracy on average (for all models). Different thresholds and \laam combinations can be an effective way to steer the ratio of data reduction and accuracy. The steering direction of the \laam combinations seems to be strongly influenced by the threshold. Therefore, selecting and fine-tuning the general threshold direction first seems to be the better first approach when applying \lasa. However, this is only a general guidance. When looking at the results in detail, it shows that selecting the right parameters for the abstraction depends highly on the dataset and needs to be fine-tuned by a human-in-the-loop.

\subsubsection{Explainability}
In this section, we further discuss the fidelity and complexity results of the \lasa method. 
\paragraph{Complexity}
While we found in Table \ref{tab:complexityPearson} that the different complexity metrics on average correlate only very weakly and at most medium strong, the correlation does vary relatively strongly, which indicates that other elements influence this correlation a lot; this is, however, in contrast to what we found in \cite{schwenke2021abstracting}. We suspect that the type and complexity of the initial data plays a major role in this; \eg if we compare many redundant datapoints that do not reduce the complexity on removal to, we have nearly no redundant datapoints where the minimal class complexity is nearly reached. Those examples prove to be also a challenge for interpreting the complexity reduction of the different complexity metrics in Tables \ref{tab:complexOri} and \ref{tab:complexSax}, as we do not know the maximal possible reduction per dataset. While we showed that it is also possible to reduce multiple complexity metrics with \lasa, multiple other conditions seem to be influencing the results, \eg that symbolizing the data can have a negative complexity effect for some metrics, as well as that a threshold can be suboptimal for the given dataset. This shows that more detailed research is needed, however, this could be a very time-consuming dataset specific approach, which is why this is not further analysed in this work. We conclude that reducing the complexity is possible with \lasa, however, multiple other dataset specific conditions need to be considered, which again highlights the need for a specific threshold optimization.

Another noteworthy irregularity we found indicates that when fine-tuning the abstraction for different classes of different general complexity for one specific dataset, each class has another optimal threshold. This suggests, again, that Attention does not seem to be class specific and can contain potentially important information for other classes. In theory, different classes would need different thresholds, which could be an extension of the \lasa approach for future work.

\paragraph{Fidelity}
When comparing the model fidelity of Tables \ref{tab:baselineSide}, \ref{tab:fidelityXAI} and \ref{tab:lasaXAI} with each other, we can see that \lasa can produce similar results as the base SAX model. However, heavy data augmentation seems to influence the predictions relatively strongly, as is suggested by \cite{kim2020puzzle,li2021interpretable} and was already seen in the model fidelity between SAX to the Ori model. Nonetheless, the \lasa model often captures some learned core decision elements of the Transformer model, seen by the low differences and lower variations between train and test model fidelity, compared to the Shapelets model fidelities. 

Another metric we analysed is the feature infidelity, which is already included in the approach of \lasa, but with a modified removal technique. Therefore, we handle the performance of \lasa as the feature infidelity, on which \lasa showed a very promising performance. Considering some irregularities in the complexity and the performance of different thresholds, this indicates that Attention has often at least a rough estimation of importance, in the sense that data reduction of less attended datapoints is possible while maintaining performance values.

\subsubsection{Shapelets Comparison}
Finding Shapelets in the \lasa reduced data proved to be not successful. This however shows that \lasa abstracts the data to a format that a Transformer can capture, while the less powerful Shapelets model can struggle with the new format of the data. If this heavily depends on a good threshold set or if it has any influence on the interpretability of the data is unclear and needs more research.

\subsection{\gcr}
In this work, we introduced two application variations and multiple structural variations and types of the GCR. In the following, we discuss them separately.

\subsubsection{\gcr as a Classifier}
Especially when using a selective approach, the \gcr shows relatively impressive results, even sometimes outperforming the original model, if it did not converge. 
This and our other analysis indicate that each individual classification task could need different types and variations of the \gcr, \ie each \gcr type and variation is a different approach to approximate the base model and has individual strong points. A further step could now be to find data criteria to better understand when, which \gcr type performs the best. While selecting the best \gcr option and variation depends on the dataset to learn, some types of tasks prove to be a challenge for our current set of \gcrs. Because we could manually differentiate the classes in the \gcr in the challenging datasets, we think that for those kinds of still challenging tasks, we can find alternative classification approaches and \gcr structures to achieve better performances. We would summarize the \gcr as a model that approximates a Transformer model with the help of human guidance.

\subsubsection{Selecting Combinations and Variants}
Looking at Table \ref{tab:gcrModelSplit} the results indicate that the \textit{hl-ss} \laam performs on average the best and that more symbols tend to perform better. Further, a higher layer $\times$ head count seem to also positively influence the \gcr. Due to the limited hyperparameter-search-space, however, it is unclear up to which point this improvement lasts. It would especially be interesting to further analyse up to which point the layers and headers do have a positive influence.

\subsubsection{Explainability}
To have a better grasp on the explainability, we looked into the certainty, model fidelity, and feature infidelity of the \gcr model. The certainty behaved very promising. While the overall performance rises with more certain elements, for the certainty of 10\% the model still contains a few misconceptions, which could be included due to multiple reasons; For example due to the limited classification ability of the \gcr or suboptimal Attention values.

The model fidelity of the \gcr in Table \ref{tab:fidelityPGCR}, compared to the model fidelity of Shapelets indicates, due to the smaller train-test gap of the \gcr, that the \gcr can successfully approximate certain core decisions of the Transformer model. Further, the feature infidelity, indicated by the \tgcr shows promising comparable performances. However, sometimes the \tgcr outperforms the \gcr, hinting on some noise included in Attention. This demonstrates similar to \lasa that Attention has a rough estimation of importance.

\subsubsection{\gcr for Interpretation}
If the \gcr tries to approximate the SAX model predictions, we received an even stronger indication that the \gcr can approximate multiple core decision mechanisms of the Transformer model, as hinted by the test results and the train-test gap in Table \ref{tab:saxOptGCRFi}. As a conclusion, the \gcr is well suited to approximate time series classification problems, while providing a helpful visualisation for interpretation.

\subsubsection{Weaknesses}
Even though the \gcr showed very promising results and included a big range of different variation, still a few classification tasks exists that can pose a problem; \eg data that is too close, has too large unknown areas in the training data or the case that classification information occurs in a wide range of positions. This was also shown by introducing the counting dataset. We argue, nonetheless, that because it is possible to distinguish the different counting classes in the \gcr visualizations, it is possible to introduce a modified classification step, which could handle this problem. This, however, is not in the scope of this work.

\subsection{Attention in Transformer Models}
While Attention in Transformer models is currently seen only as partially interpretable, we showed one way of using Attention in a human-in-the-loop approach to enhance the interpretation on a local or global level. Therefore, we argue that with a validation process per classification dataset, we can extract useful information from Attention; this information, however, can be somewhat noisy or might not be class specific, hence there is an interest for better Attention aggregation methods which enhance the quality of Attention, \eg for future work an Attention aggregation similar to \cite{chefer2021transformer} could be considered.

%% file: s7_conclusion_outlook.tex
\section{Conclusions \& Outlook}
\label{sec:conclusion}
We showed in exhaustive experimentation some application methods of Transformer Attention by using the \lasa and \gcr techniques to boost interpretability. \lasa showed the ability to find local informative features, based on similar core decisions to the original model, thus also simplifying the input to increase accessibility for humans. It enables us to include a human-in-the-loop to tradeoff accuracy with data reduction; with good thresholds, even maintaining similar accuracy to the baseline. A combination with Shapelets was sadly not successful. Further, we extended the \gcr approach to enable multiple attention aggregations into a global interpretable class approximation. The \gcr showed multiple interesting properties in regard to interpretability and showed a good approximation towards some core decisions of the model; especially if a per task approach is applied. In which case even on their own, the \gcrs showed on multiple task similar performance capabilities as the transformer model with some limitations we discussed. 

In conclusion, Attention is holding a lot of relevant information, however different tasks require different aggregations to exploit the optimal information usage. Therefore, we need to find better ways to extract this information. For future work, we want to apply \cite{chefer2021transformer} in combination with our methods to further fine-tune attention, as well as use sparse attention \cite{tay2020efficient} to reduce noise in attention.
\clearpage
\section*{Acronym List}
\begin{acronym}[all]
 \acro{ApEn}{Approximate Entropy}
 \acro{CCAM}{Column Reduced Coherence Attention Matrix}
 \acro{CE}{complexity estimation}
 \acro{CV}{Computer Vision}
 \acro{FCAM}{Full Coherence Attention Matrix}
 \acro{GCR}{Global Coherence Representation}
 \acro{GCR-P}{Penalty Global Coherence Representation}
 \acro{GCR-T}{Threshold Global Coherence Representation}
 \acro{GSA}{Global Symbol Wise Aggregation}
 \acro{GSA-P}{Global Symbol Wise Penalty Aggregation}
 \acro{GTM}{Global Trend Matrix}
 \acro{GVA}{Global Vector Aggregation}
 \acro{LAMA}{Local Attention Matrix Aggregation}
 \acro{LASA}{Local Attention-based Symbolic Abstraction}
 \acro{LASA-S}{Local Attention-based Symbolic Abstracted Shapelets}
 \acro{LAVA}{Local Attention Vector Abstraction}
 \acro{MHA}{Multi-Head Attention}
 \acro{NLP}{Natural Language Processing}
 \acro{SampEn}{Sample Entropy}
 \acro{SAX}{Symbolic Aggregate ApproXimation}
 \acro{SvdEn}{Singular Value Decomposition Entropy}
 \acro{XAI}{eXplainable Artificial Intelligence}

\end{acronym}
\clearpage

%% file: s8_appendix.tex
\begin{appendices}

\clearpage
\section{GCR Algorithms}
\label{ap:gcrAlgo}
\begin{algorithm}[H]

\begin{algorithmic}[1]
\small
\Function{calc\_\fcam\_Maxscores}{$F, C, V$}
    \State $maxScores \gets dict()$
    \For{$label$ in $C$}
        \State $maxes \gets [\text{ }]$
        \For{$v$ in $V$}
            \State $maxes.append(max(F[label][v].values(), axis=0) $
        \EndFor
        \State $maxScores[label] \gets  sum(max(maxes, axis=0))$
    \EndFor
    \State return $maxScores$
    \EndFunction
    
    \end{algorithmic}
    \caption{\small Method to calculate the approximate maximal score for a \fcam $F$, which includes all $F^c$, with $c \in C$, with $C$ being all possible classes. $V$ is vocabulary of the input.}
\label{alg:maxScoreFCAM}
\end{algorithm}
\eat{
\begin{algorithm}
\small
\begin{algorithmic}[1]
\Function{calc\_\gtm\_Maxscores}{$F, C, V$}
    \State $maxScores \gets dict()$
    \For{$label$ in $C$}
        \State $maxScores[label] \gets  sum(max(G[label].values(), axis=0))$
    \EndFor
    \State return $maxScores$
    \EndFunction
    
    \end{algorithmic}
    \caption{\small Method to calculate the approximate maximal score for a \gtm $G$, which includes all $G^c$, with $c \in C$, with $C$ being all possible classes. $V$ is vocabulary of the input.}
    \label{alg:maxScore\gtm}
\end{algorithm}
}
\begin{algorithm}[H]
\small
\begin{algorithmic}[1]
\Function{classify\_\fcam}{$F, C, n, V, X_{in}$}
    \State $labelScores \gets dict()$
    \State $maxScores \gets \text{calc\_\fcam\_Maxscores}(F, C, V)$
    \For{$c$ in $C$}
        \State $labelScores[c] \gets 0$
        \For{$i$ in $range(n)$}
            \For{$j$ in $range(n)$}
                \State $fromSymbol \gets X_{in}[i]$
                \State $toSymbol \gets X_{in}[j]$
                \State $value \gets F[c][fromSymbol][toSymbol][i][j]$
                \State $labelScores[c] \gets labelScores[c] + value$
            \EndFor
        \EndFor
        \State $labelScores[c] \gets labelScores[c]/maxScores[c]$
    \EndFor
    
    \State $highestC \gets labelScores.keys()[argmax(labelScores.values())]$
    \State $label \gets C[highestC]$
    \State return $label$
    \EndFunction
    
    \end{algorithmic}
    \caption{\small Method to classify one input sequence $X_{in}$ of length $n$ for all possible classes $C$, using a \fcam $F$. $V$ is the vocabulary of the input.}
    \label{alg:classifyFCAM}
\end{algorithm}

\begin{algorithm}[H]
\small
\begin{algorithmic}[1]
\Function{classify\_\gtm}{$G, C, n, V, X_{in}$}
    \State $labelScores \gets dict()$
    \State $maxScores \gets \text{calc\_\gtm\_Maxscores}(G, C, V)$
    \For{$c$ in $C$}
        \State $labelScores[c] \gets 0$
        \For{$i$ in $range(n)$}
                \State $toSymbol \gets X_{in}[i]$
                \State $value \gets G[c][toSymbol][i]$
                \State $labelScores[c] \gets labelScores[c] + value$
        \EndFor
        \State $labelScores[c] \gets labelScores[c]/maxScores[c]$
    \EndFor
    
    \State $highestC \gets labelScores.keys()[argmax(labelScores.values())]$
    \State $label \gets C[highestC]$
    \State return $label$
    \EndFunction
    
    \end{algorithmic}
    \caption{\small Method to classify one input sequence $X_{in}$ of length $n$ for all possible classes $C$, using a \fcam $F$. $V$ is the vocabulary of the input.}
    \label{alg:classifyGTM}
\end{algorithm}

\section{Baseline}
\begin{table}[h!]
\footnotesize
\centering
\setlength{\tabcolsep}{2pt}

\caption{Average performance baseline results for the different neural network configurations. While comparing, keep in mind, that different configurations can have different datasets as basis.}
\begin{tabular}{l|c|cccc}
Config & Data & Accuracy & Precision & Recall & F1  Score\\ \hline
All & \textit{Ori} & 0.6392 $\pm$ 0.1860 & 0.5771 $\pm$ 0.2402 & 0.6081 $\pm$ 0.1901 & 0.5593 $\pm$ 0.2292 \\
 & \textit{SAX} & 0.6262 $\pm$ 0.1752 & 0.5674 $\pm$ 0.2255 & 0.5880 $\pm$ 0.1841 & 0.5428 $\pm$ 0.2163 \\ \hline
3s-2l-8h & \textit{Ori} & 0.6441 $\pm$ 0.1742 & 0.5903 $\pm$ 0.2226 & 0.6129 $\pm$ 0.1787 & 0.5712 $\pm$ 0.2149 \\
 & \textit{SAX} & 0.6124 $\pm$ 0.1653 & 0.5538 $\pm$ 0.2152 & 0.5697 $\pm$ 0.1764 & 0.5251 $\pm$ 0.2060 \\ \hline
3s-2l-16h & \textit{Ori} & 0.6261 $\pm$ 0.2004 & 0.5650 $\pm$ 0.2613 & 0.5999 $\pm$ 0.2041 & 0.5470 $\pm$ 0.2464 \\
 & \textit{SAX} & 0.6106 $\pm$ 0.1701 & 0.5543 $\pm$ 0.2230 & 0.5760 $\pm$ 0.1833 & 0.5299 $\pm$ 0.2148 \\ \hline
3s-5l-6h & \textit{Ori} & 0.6139 $\pm$ 0.1869 & 0.5514 $\pm$ 0.2365 & 0.5898 $\pm$ 0.1866 & 0.5341 $\pm$ 0.2263 \\
 & \textit{SAX} & 0.5847 $\pm$ 0.1816 & 0.5320 $\pm$ 0.2395 & 0.5570 $\pm$ 0.1891 & 0.5010 $\pm$ 0.2225 \\ \hline
5s-2l-8h & \textit{Ori} & \textbf{0.6659 $\pm$ 0.1692} & \textbf{0.6057 $\pm$ 0.2183 }&\textbf{ 0.6234 $\pm$ 0.1797} & \textbf{0.5836 $\pm$ 0.2129} \\
 & \textit{SAX} & 0.6569 $\pm$ 0.1664 & 0.5939 $\pm$ 0.2156 & 0.6103 $\pm$ 0.1761 & 0.5733 $\pm$ 0.2067 \\ \hline
5s-2l-16h & \textit{Ori} & 0.6369 $\pm$ 0.1867 & 0.5642 $\pm$ 0.2445 & 0.6036 $\pm$ 0.1920 & 0.5489 $\pm$ 0.2295 \\
 & \textit{SAX} & 0.6540 $\pm$ 0.1705 & 0.5909 $\pm$ 0.2173 & 0.6163 $\pm$ 0.1788 & 0.5716 $\pm$ 0.2102 \\ \hline
5s-5l-6h & \textit{Ori} & 0.6460 $\pm$ 0.1948 & 0.5833 $\pm$ 0.2534 & 0.6193 $\pm$ 0.1966 & 0.5693 $\pm$ 0.2422 \\
 & \textit{SAX} & 0.6338 $\pm$ 0.1918 & 0.5758 $\pm$ 0.2423 & 0.5957 $\pm$ 0.1975 & 0.5510 $\pm$ 0.2339 \\ \hline
\begin{tabular}[c]{@{}l@{}}SAX Acc. \\ $\geq 75\%$\end{tabular} & \textit{Ori} & \multicolumn{1}{l}{0.7879 $\pm$ 0.1767} & \multicolumn{1}{l}{0.7384 $\pm$ 0.2230} & \multicolumn{1}{l}{0.7519 $\pm$ 0.1968} & \multicolumn{1}{l}{0.7284 $\pm$ 0.2229} \\
 & \textit{SAX} & \multicolumn{1}{l}{\textbf{0.8453 $\pm$ 0.0770}} & \multicolumn{1}{l}{\textbf{0.8114 $\pm$ 0.1238}} & \multicolumn{1}{l}{\textbf{0.8017 $\pm$ 0.1318}} & \multicolumn{1}{l}{\textbf{0.7963 $\pm$ 0.1337}}

\end{tabular}

\label{tabA:baseline}
\end{table}
\clearpage
\section{Shapelets Baseline}
\begin{table}[h!]
\footnotesize
\centering
\setlength{\tabcolsep}{4pt}
\caption{Average performance baseline results for the different Shapelet configurations. SAX $n$ means we used SAX data with $n$ symbols and \textit{slen} is the set minimal Shapelet length.}
\begin{tabular}{c|c|cccc}
slen & Data & Acc. & Prec. & Rec. & F1 \\ \hline
2 & \textit{Ori} & \textbf{0.7978 $\pm$ 0.1622} & \textbf{0.7728 $\pm$ 0.1944} & \textbf{0.7666 $\pm$ 0.1977} & \textbf{0.7572 $\pm$ 0.2026} \\
\multicolumn{1}{l|}{} & \textit{SAX 3} & \multicolumn{1}{l}{0.6901 $\pm$ 0.1707} & \multicolumn{1}{l}{0.6600 $\pm$ 0.2074} & \multicolumn{1}{l}{0.6520 $\pm$ 0.2100} & \multicolumn{1}{l}{0.6327 $\pm$ 0.2177} \\
 & \textit{SAX 5} & 0.7216 $\pm$ 0.1700 & 0.6940 $\pm$ 0.1995 & 0.6822 $\pm$ 0.2040 & 0.6712 $\pm$ 0.2088 \\ \hline
0.3 & \textit{Ori} & 0.7895 $\pm$ 0.1540 & 0.7727 $\pm$ 0.1748 & 0.7602 $\pm$ 0.1859 & 0.7544 $\pm$ 0.1871 \\
\multicolumn{1}{l|}{} & \textit{SAX 3} & 0.7132 $\pm$ 0.1503 & 0.6911 $\pm$ 0.1865 & 0.6784 $\pm$ 0.1937 & 0.6680 $\pm$ 0.1961 \\
 & \textit{SAX 5} & \multicolumn{1}{l}{0.7382 $\pm$ 0.1559} & \multicolumn{1}{l}{0.7146 $\pm$ 0.1804} & \multicolumn{1}{l}{0.7036 $\pm$ 0.1900} & \multicolumn{1}{l}{0.6953 $\pm$ 0.1912}
\end{tabular}
\label{tabA:baselineShap}
\end{table}

\begin{table}[h!]
\footnotesize
\centering
\setlength{\tabcolsep}{4pt}
\caption{Average performance baseline results for the different Shapelet configurations, considering only those datasets where the SAX model achieved an accuracy of $\geq 75\%$. SAX $n$ means we used SAX data with $n$ symbols and \textit{slen} is the set minimal Shapelet length.}
\begin{tabular}{c|c|cccc}
slen & Data & Acc. & Prec. & Rec. & F1 \\ \hline
2 & \textit{Ori} & 0.8526 $\pm$ 0.1409 & 0.8196 $\pm$ 0.1706 & 0.8176 $\pm$ 0.1728 & 0.8071 $\pm$ 0.1764 \\
\multicolumn{1}{l|}{} & \textit{SAX 3} & \multicolumn{1}{l}{0.8078 $\pm$ 0.1810} & \multicolumn{1}{l}{0.7990 $\pm$ 0.1852} & \multicolumn{1}{l}{0.8118 $\pm$ 0.1833} & \multicolumn{1}{l}{0.7957 $\pm$ 0.1891} \\
 & \textit{SAX 5} & 0.7967 $\pm$ 0.1652 & 0.7494 $\pm$ 0.1943 & 0.7463 $\pm$ 0.2002 & 0.7410 $\pm$ 0.1989 \\ \hline
0.3 & \textit{Ori} & \textbf{0.8558 $\pm$ 0.1291} & 0.8269 $\pm$ 0.1576 & 0.8231 $\pm$ 0.1619 & 0.8154 $\pm$ 0.1628 \\
\multicolumn{1}{l|}{} & \textit{SAX 3} & 0.8445 $\pm$ 0.1456 & \textbf{0.8361 $\pm$ 0.1498} & \textbf{0.8505 $\pm$ 0.1470} & \textbf{0.8369 $\pm$ 0.1475} \\
 & \textit{SAX 5} & \multicolumn{1}{l}{0.8133 $\pm$ 0.1542} & \multicolumn{1}{l}{0.7661 $\pm$ 0.1917} & \multicolumn{1}{l}{0.7682 $\pm$ 0.1946} & \multicolumn{1}{l}{0.7596 $\pm$ 0.1929}
\end{tabular}
\label{tabA:baselineShapGood}
\end{table}

\begin{table}[h!]
\centering
\footnotesize
\setlength{\tabcolsep}{4pt}
\caption{Model fidelity of the different Shapelets baselines, to the Ori and SAX model.}
\begin{tabular}{c|cccc}
Shapelets & \begin{tabular}[c]{@{}c@{}}Ori Train \\ Model Fidelity\end{tabular} & \begin{tabular}[c]{@{}c@{}}Ori Test \\ Model Fidelity\end{tabular} & \begin{tabular}[c]{@{}c@{}}SAX Train \\ Model Fidelity\end{tabular} & \begin{tabular}[c]{@{}c@{}}SAX Test \\ Model Fidelity\end{tabular} \\ \hline
\begin{tabular}[c]{@{}c@{}}All\\ slen 2\end{tabular} & \cellcolor[HTML]{C0C0C0} & \cellcolor[HTML]{C0C0C0} & \cellcolor[HTML]{C0C0C0} & \cellcolor[HTML]{C0C0C0} \\ \hline
Ori & 0.6968 $\pm$ 0.2376 & 0.6171 $\pm$ 0.2065 & 0.6720 $\pm$ 0.2150 & 0.6108 $\pm$ 0.1970 \\
SAX 5 & 0.6913 $\pm$ 0.2339 & 0.5912 $\pm$ 0.1873 & 0.6841 $\pm$ 0.2244 & 0.6181 $\pm$ 0.2020 \\
SAX 3 & 0.6548 $\pm$ 0.2043 & 0.5571 $\pm$ 0.1656 & 0.6627 $\pm$ 0.2105 & 0.5989 $\pm$ 0.2010 \\ \hline
\begin{tabular}[c]{@{}c@{}}All\\ slen 0.3\end{tabular} & \cellcolor[HTML]{C0C0C0} & \cellcolor[HTML]{C0C0C0} & \cellcolor[HTML]{C0C0C0} & \cellcolor[HTML]{C0C0C0} \\ \hline
Ori & 0.6998 $\pm$ 0.2398 & \textbf{0.6195 $\pm$ 0.2071} & 0.6742 $\pm$ 0.2170 & 0.6115 $\pm$ 0.1964 \\
SAX 5 & \textbf{0.7028 $\pm$ 0.2396} & 0.6061 $\pm$ 0.1982 & \textbf{0.6848 $\pm$ 0.2266} &\textbf{ 0.6268 $\pm$ 0.2064} \\
SAX 3 & 0.6829 $\pm$ 0.2294 & 0.5773 $\pm$ 0.1790 & 0.6813 $\pm$ 0.2222 & 0.6144 $\pm$ 0.2042 \\ \hline
\begin{tabular}[c]{@{}c@{}}Good\\ slen 2\end{tabular} & \cellcolor[HTML]{C0C0C0} & \cellcolor[HTML]{C0C0C0} & \cellcolor[HTML]{C0C0C0} & \cellcolor[HTML]{C0C0C0} \\ \hline
Ori & 0.8024 $\pm$ 0.2253 & 0.7120 $\pm$ 0.2255 & 0.8535 $\pm$ 0.1652 & 0.7702 $\pm$ 0.1951 \\
SAX 5 & 0.7971 $\pm$ 0.2326 & 0.6793 $\pm$ 0.2089 & 0.8530 $\pm$ 0.1891 & 0.7587 $\pm$ 0.2079 \\
SAX 3 & 0.7350 $\pm$ 0.1568 & 0.6032 $\pm$ 0.1631 & 0.8138 $\pm$ 0.1175 & 0.7023 $\pm$ 0.1961 \\ \hline
\begin{tabular}[c]{@{}c@{}}Good\\ slen 0.3\end{tabular} & \cellcolor[HTML]{C0C0C0} & \cellcolor[HTML]{C0C0C0} & \cellcolor[HTML]{C0C0C0} & \cellcolor[HTML]{C0C0C0} \\ \hline
Ori & 0.8118 $\pm$ 0.2292 & \textbf{0.7223 $\pm$ 0.2193} & 0.8629 $\pm$ 0.1669 & 0.7799 $\pm$ 0.1870 \\
SAX 5 & 0.8023 $\pm$ 0.2368 & 0.7075 $\pm$ 0.2130 & 0.8544 $\pm$ 0.1911 & \textbf{0.7824 $\pm$ 0.1989} \\
SAX 3 & \textbf{0.8232 $\pm$ 0.2022} & 0.6649 $\pm$ 0.1804 & \textbf{0.8885 $\pm$ 0.0931} & 0.7446 $\pm$ 0.1742
\end{tabular}
\label{tabA:fidelityXAI}
\end{table}

\clearpage
\section{\lasa Performance}\label{secA1}

\begin{table}[h!]
\centering
\footnotesize
\setlength{\tabcolsep}{2pt}
\caption{Relative Accuracy (to SAX model) with reduction results between different thresholds and combinations for the \lasa method.}
\begin{tabular}{l|cc|cc}
 & All (170) &  & Good (38) &  \\
Combi & Accuracy & Reduction & Accuracy & Reduction \\ \hline
\multicolumn{1}{c|}{\begin{tabular}[c]{@{}c@{}}Avg.\\ {[}1, 1.2{]}\end{tabular}} & \cellcolor[HTML]{C0C0C0} &\cellcolor[HTML]{C0C0C0}  & \cellcolor[HTML]{C0C0C0} &  \cellcolor[HTML]{C0C0C0} \\\hline
hl-mmm & -0.0364 $\pm$ 0.0754 & \textbf{0.5399 $\pm$ 0.0859} & -0.1108 $\pm$ 0.0989 & 0.5553 $\pm$ 0.0678 \\
hl-mms & -0.0360 $\pm$ 0.0739 & 0.5032 $\pm$ 0.0862 & -0.1031 $\pm$ 0.1028 & 0.5025 $\pm$ 0.0580 \\
hl-msm & -0.0355 $\pm$ 0.0811 & 0.5106 $\pm$ 0.0885 & -0.1161 $\pm$ 0.1134 & 0.5106 $\pm$ 0.0750 \\
hl-mss & -0.0364 $\pm$ 0.0731 & 0.4802 $\pm$ 0.0908 & -0.1055 $\pm$ 0.1004 & 0.4708 $\pm$ 0.0719 \\
hl-smm & -0.0393 $\pm$ 0.0654 & 0.5292 $\pm$ 0.0878 & -0.0976 $\pm$ 0.0837 & \textbf{0.5483 $\pm$ 0.0633} \\
hl-sms & \textbf{-0.0237 $\pm$ 0.0585} & 0.4966 $\pm$ 0.0837 & -0.0713 $\pm$ 0.0684 & 0.5081 $\pm$ 0.0606 \\
hl-ssm & -0.0333 $\pm$ 0.0692 & 0.5027 $\pm$ 0.0914 & -0.0876 $\pm$ 0.0978 & 0.5137 $\pm$ 0.0755 \\
hl-sss & -0.0258 $\pm$ 0.0616 & 0.4624 $\pm$ 0.0792 & \textbf{-0.0707 $\pm$ 0.0795} & 0.4828 $\pm$ 0.0614 \\ \hline
\multicolumn{1}{c|}{\begin{tabular}[c]{@{}c@{}}Avg.\\ {[}0.8, 1.5{]}\end{tabular}} & \cellcolor[HTML]{C0C0C0} &\cellcolor[HTML]{C0C0C0}  & \cellcolor[HTML]{C0C0C0} &  \cellcolor[HTML]{C0C0C0} \\\hline
hl-mmm & \textbf{-0.0474 $\pm$ 0.0660} & 0.7990 $\pm$ 0.1196 & -0.1015 $\pm$ 0.0779 & 0.7041 $\pm$ 0.0965 \\
hl-mms & -0.0685 $\pm$ 0.0818 & 0.8587 $\pm$ 0.1201 & \textbf{-0.1412 $\pm$ 0.1034} & 0.7555 $\pm$ 0.1262 \\
hl-msm & -0.0712 $\pm$ 0.0963 & 0.8532 $\pm$ 0.1239 & -0.1574 $\pm$ 0.1267 & 0.7532 $\pm$ 0.1278 \\
hl-mss & -0.0776 $\pm$ 0.0898 & 0.9057 $\pm$ 0.1048 & -0.1440 $\pm$ 0.1144 & 0.8173 $\pm$ 0.1330 \\
hl-smm & -0.0856 $\pm$ 0.0961 & 0.8929 $\pm$ 0.0969 & -0.1673 $\pm$ 0.1175 & 0.8050 $\pm$ 0.1084 \\
hl-sms & -0.1044 $\pm$ 0.1107 & 0.9500 $\pm$ 0.0650 & -0.2086 $\pm$ 0.1303 & 0.8868 $\pm$ 0.0929 \\
hl-ssm & -0.1026 $\pm$ 0.0983 & 0.9392 $\pm$ 0.0745 & -0.1966 $\pm$ 0.1060 & 0.8713 $\pm$ 0.1026 \\
hl-sss & -0.1264 $\pm$ 0.1439 & \textbf{0.9664 $\pm$ 0.0534} & -0.2659 $\pm$ 0.1657 & \textbf{0.9207 $\pm$ 0.0854} \\ \hline
\multicolumn{1}{c|}{\begin{tabular}[c]{@{}c@{}}Max\\ {[}2, 3{]}\end{tabular}} & \cellcolor[HTML]{C0C0C0} &\cellcolor[HTML]{C0C0C0}  & \cellcolor[HTML]{C0C0C0} &  \cellcolor[HTML]{C0C0C0}\\\hline
hl-mmm & -0.0575 $\pm$ 0.1071 & \textbf{0.4076 $\pm$ 0.2721} & -0.1657 $\pm$ 0.1459 & \textbf{0.5766 $\pm$ 0.2533} \\
hl-mms & -0.0376 $\pm$ 0.0823 & 0.2266 $\pm$ 0.2465 & -0.1011 $\pm$ 0.1138 & 0.3554 $\pm$ 0.2325 \\
hl-msm & -0.0502 $\pm$ 0.0844 & 0.2655 $\pm$ 0.2690 & -0.1230 $\pm$ 0.1154 & 0.4061 $\pm$ 0.2864 \\
hl-mss & -0.0203 $\pm$ 0.0730 & 0.1345 $\pm$ 0.2170 & -0.0645 $\pm$ 0.1019 & 0.2034 $\pm$ 0.2020 \\
hl-smm & -0.0438 $\pm$ 0.0951 & 0.2164 $\pm$ 0.2612 & -0.1301 $\pm$ 0.1354 & 0.3831 $\pm$ 0.2777 \\
hl-sms & \textbf{-0.0085 $\pm$ 0.0575} & 0.0800 $\pm$ 0.1893 & \textbf{-0.0227 $\pm$ 0.0622} & 0.0921 $\pm$ 0.1472 \\
hl-ssm & -0.0198 $\pm$ 0.0630 & 0.1160 $\pm$ 0.2266 & -0.0578 $\pm$ 0.0880 & 0.2057 $\pm$ 0.2441 \\
hl-sss & -0.0142 $\pm$ 0.0639 & 0.0497 $\pm$ 0.1509 & -0.0439 $\pm$ 0.0800 & 0.0422 $\pm$ 0.1137 \\ \hline
\multicolumn{1}{c|}{\begin{tabular}[c]{@{}c@{}}Max\\ {[}1.8, -1{]}\end{tabular}} &  \cellcolor[HTML]{C0C0C0} &\cellcolor[HTML]{C0C0C0}  & \cellcolor[HTML]{C0C0C0} &  \cellcolor[HTML]{C0C0C0}  \\\hline
hl-mmm & -0.0699 $\pm$ 0.1024 & \textbf{0.4720 $\pm$ 0.2745} & -0.1784 $\pm$ 0.1163 & 0.6432 $\pm$ 0.2411 \\
hl-mms & -0.0432 $\pm$ 0.0837 & 0.2821 $\pm$ 0.2640 & -0.1213 $\pm$ 0.1024 & 0.4457 $\pm$ 0.2537 \\
hl-msm & -0.0486 $\pm$ 0.0920 & 0.3154 $\pm$ 0.2846 & -0.1341 $\pm$ 0.1197 & \textbf{0.4795 $\pm$ 0.2948} \\
hl-mss & -0.0143 $\pm$ 0.0668 & 0.1700 $\pm$ 0.2372 & -0.0550 $\pm$ 0.0676 & 0.2721 $\pm$ 0.2356 \\
hl-smm & -0.0518 $\pm$ 0.0857 & 0.2672 $\pm$ 0.2765 & -0.1317 $\pm$ 0.1093 & 0.4556 $\pm$ 0.2927 \\
hl-sms & -0.0130 $\pm$ 0.0553 & 0.1059 $\pm$ 0.2092 & -0.0439 $\pm$ 0.0599 & 0.1635 $\pm$ 0.1855 \\
hl-ssm & -0.0347 $\pm$ 0.0754 & 0.1438 $\pm$ 0.2457 & -0.0886 $\pm$ 0.1008 & 0.2689 $\pm$ 0.2814 \\
hl-sss & \textbf{-0.0083 $\pm$ 0.0542} & 0.0619 $\pm$ 0.1729 & \textbf{-0.0224 $\pm$ 0.0613} & 0.0644 $\pm$ 0.1450
\end{tabular}

\label{tabA:lasaAR}
\end{table}

\begin{table}[h!]
\centering
\footnotesize
\caption{Average Ranking of accuracy and reduction ratio.}
\begin{tabular}{l|ccccccc}
Rankings & All & 3s-2l-8h & 3s-2l-16h & 3s-5l-8h & 5s-2l-8h & 5s-2l-16h & 5s-5l-8h \\ \hline
\begin{tabular}[c]{@{}l@{}}Avg.\\ {[}1, 1.2{]}\end{tabular} &  \cellcolor[HTML]{C0C0C0}& \cellcolor[HTML]{C0C0C0}& \cellcolor[HTML]{C0C0C0}& \cellcolor[HTML]{C0C0C0}& \cellcolor[HTML]{C0C0C0}& \cellcolor[HTML]{C0C0C0}& \cellcolor[HTML]{C0C0C0}  \\\hline
hl-mmm & \textbf{3.3765} & \textbf{3.5833} & \textbf{3.3594} & 3.4600 & \textbf{3.4355} & \textbf{3.2000} & \textbf{3.1818} \\
hl-mms & 4.6147 & 4.8167 & 5.0156 & 3.8000 & 4.6129 & 4.9667 & 4.2045 \\
hl-msm & 4.2588 & 4.4000 & 4.0781 & 4.3400 & 4.3871 & 4.1667 & 4.1818 \\
hl-mss & 5.1324 & 5.4167 & 5.0938 & 5.2800 & 4.8226 & 5.2000 & 4.9773 \\
hl-smm & 3.7588 & 3.7333 & 3.6406 & \textbf{3.2800} & 4.2258 & 3.7167 & 3.9091 \\
hl-sms & 4.4059 & 4.2500 & 4.4062 & 4.4400 & 4.6613 & 4.3000 & 4.3636 \\
hl-ssm & 4.5941 & 4.0500 & 4.4531 & 5.3400 & 4.3065 & 4.4000 & 5.3636 \\
hl-sss & 5.5441 & 5.5333 & 5.5938 & 5.7000 & 5.3387 & 5.6333 & 5.4773 \\ \hline
\begin{tabular}[c]{@{}l@{}}Avg.\\ {[}0.8, 1.5{]}\end{tabular} & \cellcolor[HTML]{C0C0C0}& \cellcolor[HTML]{C0C0C0}& \cellcolor[HTML]{C0C0C0}& \cellcolor[HTML]{C0C0C0}& \cellcolor[HTML]{C0C0C0}& \cellcolor[HTML]{C0C0C0}& \cellcolor[HTML]{C0C0C0} \\\hline
hl-mmm & 5.4235 & 5.3167 & 5.7031 & 5.6800 & 5.3387 & 5.1167 & 5.4091 \\
hl-mms & 4.8265 & 5.0500 & 4.9219 & 4.9600 & 4.6452 & 4.4167 & 5.0455 \\
hl-msm & 4.9588 & 5.5833 & 4.9219 & 4.5400 & 5.0484 & 5.0167 & 4.4318 \\
hl-mss & 3.7912 & 3.8500 & 3.7656 & 3.5600 & 3.9839 & 3.8333 & 3.6818 \\
hl-smm & 4.4853 & 4.4333 & 4.2969 & 4.5600 & 4.4032 & 4.4833 & 4.8636 \\
hl-sms & 3.5147 & 3.4167 & 3.4531 & 3.6600 & 3.4677 & 3.5833 & 3.5455 \\
hl-ssm & 3.6706 & 3.6167 & 3.6250 & 3.2000 & 3.9355 & 4.0000 & 3.5227 \\
hl-sss & \textbf{3.2029} & \textbf{3.2667} & \textbf{3.2031} & \textbf{2.6600} & \textbf{3.4516} & \textbf{3.4500} & \textbf{3.0455} \\ \hline
\begin{tabular}[c]{@{}l@{}}Max\\ {[}2, 3{]}\end{tabular} & \cellcolor[HTML]{C0C0C0}& \cellcolor[HTML]{C0C0C0}& \cellcolor[HTML]{C0C0C0}& \cellcolor[HTML]{C0C0C0}& \cellcolor[HTML]{C0C0C0}& \cellcolor[HTML]{C0C0C0}& \cellcolor[HTML]{C0C0C0}  \\\hline
hl-mmm & \textbf{3.3882} & \textbf{3.3667} & \textbf{3.2969} & 3.2400 & \textbf{3.5000} & \textbf{3.5333} & \textbf{3.3636} \\
hl-mms & 3.8353 & 3.9500 & 4.0156 & \textbf{3.1800} & 3.8548 & 4.1167 & 3.7500 \\
hl-msm & 3.9618 & 3.9500 & 3.7656 & 4.5200 & 3.7258 & 3.8667 & 4.0909 \\
hl-mss & 4.1000 & 4.3667 & 4.4062 & 3.8200 & 4.1774 & 3.8167 & 3.8864 \\
hl-smm & 4.0882 & 4.2500 & 4.2812 & 3.7600 & 3.9516 & 4.1667 & 4.0455 \\
hl-sms & 4.3735 & 4.5833 & 4.3125 & 4.5600 & 4.5806 & 4.2333 & 3.8636 \\
hl-ssm & 4.3912 & 4.2667 & 4.4375 & 4.3400 & 4.4194 & 4.4500 & 4.4318 \\
hl-sss & 4.6265 & 4.7667 & 4.5469 & 4.6200 & 4.6613 & 4.5500 & 4.6136 \\ \hline
\begin{tabular}[c]{@{}l@{}}Max\\ {[}1.8, -1{]}\end{tabular} &  \cellcolor[HTML]{C0C0C0}& \cellcolor[HTML]{C0C0C0}& \cellcolor[HTML]{C0C0C0}& \cellcolor[HTML]{C0C0C0}& \cellcolor[HTML]{C0C0C0}& \cellcolor[HTML]{C0C0C0}& \cellcolor[HTML]{C0C0C0}  \\\hline
hl-mmm & \textbf{3.4647} & \textbf{3.2667} & \textbf{3.5156} & \textbf{3.4200} & \textbf{3.3710} & \textbf{3.5333} & \textbf{3.7500} \\
hl-mms & 3.8971 & 3.9500 & 4.0469 & 3.4800 & 3.9516 & 4.1000 & 3.7273 \\
hl-msm & 3.7853 & 3.7333 & 3.8281 & 4.0400 & 3.4839 & 3.7333 & 4.0000 \\
hl-mss & 4.0353 & 4.0500 & 4.0156 & 3.8000 & 4.1613 & 4.1167 & 4.0227 \\
hl-smm & 4.1353 & 4.1333 & 4.2344 & 4.0000 & 4.2742 & 4.0500 & 4.0682 \\
hl-sms & 4.4853 & 4.8500 & 4.4531 & 4.1600 & 4.6613 & 4.4667 & 4.1818 \\
hl-ssm & 4.7029 & 4.9167 & 4.7344 & 4.8200 & 4.7742 & 4.4000 & 4.5455 \\
hl-sss & 4.7882 & 5.0167 & 4.7500 & 4.7000 & 4.8387 & 4.6167 & 4.7955
\end{tabular}

\label{tabA:lasaRankings}
\end{table}

\begin{table}[h!]
\centering
\footnotesize
\setlength{\tabcolsep}{2pt}
\caption{Average performance of the \lasa approach, relative to the base SAX model, for only the good models.}
\begin{tabular}{l|cccc}
\begin{tabular}[c]{@{}l@{}}Threshold\\ Combi\end{tabular} & Accuracy & Precision & Recall & F1 \\ \hline
\begin{tabular}[c]{@{}l@{}}Avg.\\ {[}1, 1.2{]}\end{tabular} & \cellcolor[HTML]{C0C0C0} & \cellcolor[HTML]{C0C0C0} & \cellcolor[HTML]{C0C0C0} & \cellcolor[HTML]{C0C0C0} \\
hl-mmm & -0.1108 $\pm$ 0.0989 & -0.1510 $\pm$ 0.1208 & -0.1532 $\pm$ 0.1032 & -0.1713 $\pm$ 0.1142 \\
hl-mms & -0.1031 $\pm$ 0.1028 & -0.1409 $\pm$ 0.1209 & -0.1406 $\pm$ 0.1068 & -0.1596 $\pm$ 0.1154 \\
hl-msm & -0.1161 $\pm$ 0.1134 & -0.1499 $\pm$ 0.1288 & -0.1548 $\pm$ 0.1113 & -0.1733 $\pm$ 0.1212 \\
hl-mss & -0.1055 $\pm$ 0.1004 & -0.1394 $\pm$ 0.1265 & -0.1423 $\pm$ 0.1060 & -0.1622 $\pm$ 0.1178 \\
hl-smm & -0.0976 $\pm$ 0.0837 & -0.1375 $\pm$ 0.1063 & -0.1420 $\pm$ 0.0933 & -0.1589 $\pm$ 0.1022 \\
hl-sms & -0.0713 $\pm$ 0.0684 & -0.1062 $\pm$ 0.1011 & -0.1117 $\pm$ 0.0942 & -0.1259 $\pm$ 0.0996 \\
hl-ssm & -0.0876 $\pm$ 0.0978 & -0.1227 $\pm$ 0.1148 & -0.1305 $\pm$ 0.1068 & -0.1450 $\pm$ 0.1115 \\
hl-sss & \textbf{-0.0707 $\pm$ 0.0795} &\textbf{ -0.0993 $\pm$ 0.1020} & \textbf{-0.1084 $\pm$ 0.0996} & \textbf{-0.1230 $\pm$ 0.1065} \\ \hline
\begin{tabular}[c]{@{}l@{}}Avg.\\ {[}0.8, 1.5{]}\end{tabular} & \cellcolor[HTML]{C0C0C0} & \cellcolor[HTML]{C0C0C0} & \cellcolor[HTML]{C0C0C0} & \cellcolor[HTML]{C0C0C0}  \\\hline
hl-mmm & \textbf{-0.1015 $\pm$ 0.0779} & \textbf{-0.1351 $\pm$ 0.1130} & \textbf{-0.1473 $\pm$ 0.0971} & \textbf{-0.1681 $\pm$ 0.1135} \\
hl-mms & -0.1412 $\pm$ 0.1034 & -0.1825 $\pm$ 0.1342 & -0.1812 $\pm$ 0.1045 & -0.2097 $\pm$ 0.1256 \\
hl-msm & -0.1574 $\pm$ 0.1267 & -0.1987 $\pm$ 0.1439 & -0.1946 $\pm$ 0.1163 & -0.2241 $\pm$ 0.1380 \\
hl-mss & -0.1440 $\pm$ 0.1144 & -0.2031 $\pm$ 0.1330 & -0.1947 $\pm$ 0.1045 & -0.2232 $\pm$ 0.1238 \\
hl-smm & -0.1673 $\pm$ 0.1175 & -0.2229 $\pm$ 0.1530 & -0.2092 $\pm$ 0.1176 & -0.2419 $\pm$ 0.1413 \\
hl-sms & -0.2086 $\pm$ 0.1303 & -0.2848 $\pm$ 0.1653 & -0.2497 $\pm$ 0.1236 & -0.2978 $\pm$ 0.1521 \\
hl-ssm & -0.1966 $\pm$ 0.1060 & -0.2586 $\pm$ 0.1407 & -0.2361 $\pm$ 0.1129 & -0.2801 $\pm$ 0.1351 \\
hl-sss & -0.2659 $\pm$ 0.1657 & -0.3551 $\pm$ 0.1878 & -0.3044 $\pm$ 0.1479 & -0.3655 $\pm$ 0.1770 \\ \hline
\begin{tabular}[c]{@{}l@{}}Max\\ {[}2, 3{]}\end{tabular} &  \cellcolor[HTML]{C0C0C0} & \cellcolor[HTML]{C0C0C0} & \cellcolor[HTML]{C0C0C0} & \cellcolor[HTML]{C0C0C0} \\\hline
hl-mmm & -0.1657 $\pm$ 0.1459 & -0.2058 $\pm$ 0.1517 & -0.2003 $\pm$ 0.1403 & -0.2237 $\pm$ 0.1544 \\
hl-mms & -0.1011 $\pm$ 0.1138 & -0.1404 $\pm$ 0.1366 & -0.1399 $\pm$ 0.1174 & -0.1574 $\pm$ 0.1267 \\
hl-msm & -0.1230 $\pm$ 0.1154 & -0.1575 $\pm$ 0.1202 & -0.1571 $\pm$ 0.1116 & -0.1793 $\pm$ 0.1188 \\
hl-mss & -0.0645 $\pm$ 0.1019 & -0.1000 $\pm$ 0.1239 & -0.1038 $\pm$ 0.1145 & -0.1167 $\pm$ 0.1232 \\
hl-smm & -0.1301 $\pm$ 0.1354 & -0.1628 $\pm$ 0.1468 & -0.1658 $\pm$ 0.1307 & -0.1864 $\pm$ 0.1429 \\
hl-sms & \textbf{-0.0227 $\pm$ 0.0622} & \textbf{-0.0605 $\pm$ 0.1093} & \textbf{-0.0642 $\pm$ 0.0980} & \textbf{-0.0710 $\pm$ 0.1052} \\
hl-ssm & -0.0578 $\pm$ 0.0880 & -0.0982 $\pm$ 0.1120 & -0.0978 $\pm$ 0.0981 & -0.1095 $\pm$ 0.1061 \\
hl-sss & -0.0439 $\pm$ 0.0800 & -0.0829 $\pm$ 0.1146 & -0.0834 $\pm$ 0.0979 & -0.0943 $\pm$ 0.1077 \\ \hline
\begin{tabular}[c]{@{}l@{}}Max\\ {[}1.8, -1{]}\end{tabular} & \cellcolor[HTML]{C0C0C0} & \cellcolor[HTML]{C0C0C0} & \cellcolor[HTML]{C0C0C0} & \cellcolor[HTML]{C0C0C0} \\\hline
hl-mmm & -0.1784 $\pm$ 0.1163 & -0.2252 $\pm$ 0.1310 & -0.2126 $\pm$ 0.1178 & -0.2479 $\pm$ 0.1297 \\
hl-mms & -0.1213 $\pm$ 0.1024 & -0.1547 $\pm$ 0.1157 & -0.1548 $\pm$ 0.1063 & -0.1747 $\pm$ 0.1158 \\
hl-msm & -0.1341 $\pm$ 0.1197 & -0.1771 $\pm$ 0.1265 & -0.1741 $\pm$ 0.1127 & -0.1933 $\pm$ 0.1269 \\
hl-mss & -0.0550 $\pm$ 0.0676 & -0.0937 $\pm$ 0.0915 & -0.0983 $\pm$ 0.0893 & -0.1066 $\pm$ 0.0901 \\
hl-smm & -0.1317 $\pm$ 0.1093 & -0.1808 $\pm$ 0.1294 & -0.1721 $\pm$ 0.1180 & -0.1965 $\pm$ 0.1288 \\
hl-sms & -0.0439 $\pm$ 0.0599 & -0.0770 $\pm$ 0.0954 & -0.0832 $\pm$ 0.0916 & -0.0929 $\pm$ 0.0970 \\
hl-ssm & -0.0886 $\pm$ 0.1008 & -0.1277 $\pm$ 0.1189 & -0.1305 $\pm$ 0.1053 & -0.1461 $\pm$ 0.1183 \\
hl-sss &\textbf{ -0.0224 $\pm$ 0.0613} & \textbf{-0.0491 $\pm$ 0.0917} & \textbf{-0.0596 $\pm$ 0.0944} & \textbf{-0.0672 $\pm$ 0.0983}
\end{tabular}

\label{tab:perfLasaGood}
\end{table}

\begin{table}[h!]
\centering
\footnotesize
\caption{Average model fidelity from different \lasa configurations to the SAX model}
\begin{tabular}{l|ccl}
\begin{tabular}[c]{@{}c@{}}Rankings\\ Good\end{tabular} & \begin{tabular}[c]{@{}c@{}}SAX Train\\ Fidelity\end{tabular} & \begin{tabular}[c]{@{}c@{}}SAX Test\\ Fidelity\end{tabular} \\ \hline
\begin{tabular}[c]{@{}l@{}}Avg.\\ {[}1, 1.2{]}\end{tabular} &\cellcolor[HTML]{C0C0C0}&\cellcolor[HTML]{C0C0C0}  \\\hline
hl-mmm & 0.8244 $\pm$ 0.1404 & 0.7684 $\pm$ 0.1317  \\
hl-mms & 0.8379 $\pm$ 0.1163 & 0.7877 $\pm$ 0.1111  \\
hl-msm & 0.8194 $\pm$ 0.1282 & 0.7716 $\pm$ 0.1232  \\
hl-mss & 0.8279 $\pm$ 0.1489 & 0.7776 $\pm$ 0.1419  \\
hl-smm & 0.8449 $\pm$ 0.1330 & 0.7981 $\pm$ 0.1221  \\
hl-sms & 0.8648 $\pm$ 0.1103 & 0.8116 $\pm$ 0.1101  \\
hl-ssm & 0.8452 $\pm$ 0.1220 & 0.7978 $\pm$ 0.1228  \\
hl-sss & \textbf{0.8658 $\pm$ 0.1057} & \textbf{0.8158 $\pm$ 0.1009}  \\ \hline
\begin{tabular}[c]{@{}l@{}}Avg.\\ {[}0.8, 1.5{]}\end{tabular} &\cellcolor[HTML]{C0C0C0}  &\cellcolor[HTML]{C0C0C0}   \\\hline
hl-mmm & \textbf{0.8265 $\pm$ 0.1534} & \textbf{0.7705 $\pm$ 0.1370}  \\
hl-mms & 0.8045 $\pm$ 0.1329 & 0.7430 $\pm$ 0.1337  \\
hl-msm & 0.7897 $\pm$ 0.1486 & 0.7327 $\pm$ 0.1438  \\
hl-mss & 0.7938 $\pm$ 0.1219 & 0.7416 $\pm$ 0.1240  \\
hl-smm & 0.7795 $\pm$ 0.1328 & 0.7194 $\pm$ 0.1289  \\
hl-sms & 0.7200 $\pm$ 0.1526 & 0.6678 $\pm$ 0.1514  \\
hl-ssm & 0.7506 $\pm$ 0.1206 & 0.6905 $\pm$ 0.1244  \\
hl-sss & 0.6739 $\pm$ 0.1625 & 0.6209 $\pm$ 0.1630  \\\hline
\begin{tabular}[c]{@{}l@{}}Max\\ {[}2, 3{]}\end{tabular} &\cellcolor[HTML]{C0C0C0} &\cellcolor[HTML]{C0C0C0}   \\\hline
hl-mmm & 0.7770 $\pm$ 0.1240 & 0.7170 $\pm$ 0.1342  \\
hl-mms & 0.8409 $\pm$ 0.1421 & 0.7847 $\pm$ 0.1440  \\
hl-msm & 0.8189 $\pm$ 0.1452 & 0.7638 $\pm$ 0.1505  \\
hl-mss & 0.8611 $\pm$ 0.1237 & 0.8207 $\pm$ 0.1283  \\
hl-smm & 0.8218 $\pm$ 0.1465 & 0.7684 $\pm$ 0.1514  \\
hl-sms & \textbf{0.9099 $\pm$ 0.1052} & \textbf{0.8766 $\pm$ 0.1108}  \\
hl-ssm & 0.8808 $\pm$ 0.1214 & 0.8327 $\pm$ 0.1400  \\
hl-sss & 0.9032 $\pm$ 0.1046 & 0.8704 $\pm$ 0.1050  \\\hline
\begin{tabular}[c]{@{}l@{}}Max\\ {[}1.8, -1{]}\end{tabular} &\cellcolor[HTML]{C0C0C0}  &\cellcolor[HTML]{C0C0C0}   \\\hline
hl-mmm & 0.7571 $\pm$ 0.1190 & 0.6996 $\pm$ 0.1367  \\
hl-mms & 0.8408 $\pm$ 0.0942 & 0.7696 $\pm$ 0.1110  \\
hl-msm & 0.8064 $\pm$ 0.1320 & 0.7488 $\pm$ 0.1488  \\
hl-mss & 0.8932 $\pm$ 0.0779 & 0.8459 $\pm$ 0.1042  \\
hl-smm & 0.8025 $\pm$ 0.1227 & 0.7455 $\pm$ 0.1437  \\
hl-sms & 0.9033 $\pm$ 0.0800 & 0.8582 $\pm$ 0.0920  \\
hl-ssm & 0.8555 $\pm$ 0.1333 & 0.8029 $\pm$ 0.1470  \\
hl-sss & \textbf{0.9192 $\pm$ 0.0869} & \textbf{0.8768 $\pm$ 0.1015} 
\end{tabular}

\label{tabA:lasaXAI}
\end{table}

\clearpage
\section{\lasa complexity}

\begin{table}[htp]
\centering
\tiny
\setlength{\tabcolsep}{2pt}
\caption{\lasa Complexity reduction compared to the SAX model complexity.}
\label{tabA:complexSax}
\begin{tabular}{l|ccccc}
to sax                                                             & SvdEn               & ApEn                & SampEn               & CE                   & T. Shifts           \\ \hline
\begin{tabular}[c]{@{}c@{}}Avg.\\ {[}1.0,1.2{]}\end{tabular} & \cellcolor[HTML]{C0C0C0}&  \cellcolor[HTML]{C0C0C0}&  \cellcolor[HTML]{C0C0C0}&  \cellcolor[HTML]{C0C0C0}&  \cellcolor[HTML]{C0C0C0}  \\  \hline
hl-mmm                                                       & 0.1146 $\pm$ 0.1821 & \textbf{0.2379 $\pm$ 0.2568} & \textbf{0.0828 $\pm$ 0.7556}  & 0.0525 $\pm$ 0.3643  & \textbf{0.4325 $\pm$ 0.2433 }\\
hl-mms                                                       & 0.1134 $\pm$ 0.1656 & 0.1968 $\pm$ 0.2343 & 0.0399 $\pm$ 0.7930  & 0.0742 $\pm$ 0.3389  & 0.3871 $\pm$ 0.2117 \\
hl-msm                                                       & 0.1195 $\pm$ 0.1695 & 0.2081 $\pm$ 0.2428 & 0.0440 $\pm$ 0.7687  & 0.0817 $\pm$ 0.3367  & 0.3948 $\pm$ 0.2247 \\
hl-mss                                                       & \textbf{0.1239 $\pm$ 0.1548} & 0.1746 $\pm$ 0.2324 & -0.0385 $\pm$ 0.9581 & \textbf{0.1038 $\pm$ 0.3099}  & 0.3622 $\pm$ 0.1996 \\
hl-smm                                                       & 0.1131 $\pm$ 0.1592 & 0.1804 $\pm$ 0.2480 & -0.0514 $\pm$ 0.9879 & 0.0700 $\pm$ 0.3387  & 0.3791 $\pm$ 0.2290 \\
hl-sms                                                       & 0.1068 $\pm$ 0.1488 & 0.1534 $\pm$ 0.2174 & -0.1381 $\pm$ 1.2408 & 0.0759 $\pm$ 0.3113  & 0.3600 $\pm$ 0.1864 \\
hl-ssm                                                       & 0.1112 $\pm$ 0.1471 & 0.1548 $\pm$ 0.2339 & -0.1310 $\pm$ 1.1763 & 0.0819 $\pm$ 0.3127  & 0.3567 $\pm$ 0.2028 \\
hl-sss                                                       & 0.1061 $\pm$ 0.1455 & 0.1590 $\pm$ 0.2049 & -0.0958 $\pm$ 1.2945 & 0.0805 $\pm$ 0.2965  & 0.3610 $\pm$ 0.1900 \\ \hline
\begin{tabular}[c]{@{}c@{}}Avg.\\ {[}0.8,1.5{]}\end{tabular} &  \cellcolor[HTML]{C0C0C0}&  \cellcolor[HTML]{C0C0C0}&  \cellcolor[HTML]{C0C0C0}&  \cellcolor[HTML]{C0C0C0}&  \cellcolor[HTML]{C0C0C0}  \\  \hline
hl-mmm                                                       & 0.2324 $\pm$ 0.2020 & 0.3459 $\pm$ 0.3384 & -0.1056 $\pm$ 1.5060 & 0.1224 $\pm$ 0.4006  & 0.6747 $\pm$ 0.2155 \\
hl-mms                                                       & 0.3055 $\pm$ 0.2324 & 0.4707 $\pm$ 0.3866 & 0.0679 $\pm$ 1.4606  & 0.1756 $\pm$ 0.3969  & 0.7530 $\pm$ 0.2160 \\
hl-msm                                                       & 0.2956 $\pm$ 0.2204 & 0.4559 $\pm$ 0.3772 & 0.0117 $\pm$ 1.6538  & 0.1618 $\pm$ 0.3942  & 0.7526 $\pm$ 0.2089 \\
hl-mss                                                       & 0.3482 $\pm$ 0.2239 & 0.5610 $\pm$ 0.3862 & 0.2237 $\pm$ 1.3758  & 0.2034 $\pm$ 0.3893  & 0.8225 $\pm$ 0.1872 \\
hl-smm                                                       & 0.3311 $\pm$ 0.2197 & 0.5164 $\pm$ 0.3755 & 0.1706 $\pm$ 1.3359  & 0.1846 $\pm$ 0.3933  & 0.8101 $\pm$ 0.1742 \\
hl-sms                                                       & 0.3906 $\pm$ 0.2050 & 0.6596 $\pm$ 0.3361 & 0.4438 $\pm$ 1.0401  & 0.2238 $\pm$ 0.3776  & 0.8907 $\pm$ 0.1356 \\
hl-ssm                                                       & 0.3799 $\pm$ 0.2069 & 0.6283 $\pm$ 0.3426 & 0.4138 $\pm$ 0.8266  & 0.2236 $\pm$ 0.3783  & 0.8777 $\pm$ 0.1331 \\
hl-sss                                                       & \textbf{0.4170 $\pm$ 0.1883} & \textbf{0.7226 $\pm$ 0.3103} & \textbf{0.5553 $\pm$ 0.7638}  & \textbf{0.2562 $\pm$ 0.3546}  & \textbf{0.9172 $\pm$ 0.1153} \\ \hline
\begin{tabular}[c]{@{}c@{}}Max.\\ {[}2,3{]}\end{tabular}     &   \cellcolor[HTML]{C0C0C0}&  \cellcolor[HTML]{C0C0C0}&  \cellcolor[HTML]{C0C0C0}&  \cellcolor[HTML]{C0C0C0}&  \cellcolor[HTML]{C0C0C0}  \\  \hline
hl-mmm                                                       & \textbf{0.1006 $\pm$ 0.1747} & \textbf{0.2449 $\pm$ 0.2778} & \textbf{0.0653 $\pm$ 0.7301}  & -0.0127 $\pm$ 0.3343 & \textbf{0.3895 $\pm$ 0.3525} \\
hl-mms                                                       & 0.0589 $\pm$ 0.1205 & 0.1264 $\pm$ 0.1996 & -0.0310 $\pm$ 0.6921 & 0.0073 $\pm$ 0.2443  & 0.2188 $\pm$ 0.2758 \\
hl-msm                                                       & 0.0827 $\pm$ 0.1557 & 0.1615 $\pm$ 0.2310 & 0.0035 $\pm$ 0.6740  &\textbf{ 0.0226 $\pm$ 0.2685}  & 0.2642 $\pm$ 0.3151 \\
hl-mss                                                       & 0.0384 $\pm$ 0.0900 & 0.0662 $\pm$ 0.1551 & -0.0535 $\pm$ 0.6854 & 0.0087 $\pm$ 0.1665  & 0.1257 $\pm$ 0.2227 \\
hl-smm                                                       & 0.0659 $\pm$ 0.1471 & 0.1166 $\pm$ 0.2173 & -0.0338 $\pm$ 0.6866 & 0.0132 $\pm$ 0.2514  & 0.2079 $\pm$ 0.3055 \\
hl-sms                                                       & 0.0178 $\pm$ 0.0609 & 0.0310 $\pm$ 0.1261 & -0.0684 $\pm$ 0.5834 & -0.0063 $\pm$ 0.1097 & 0.0640 $\pm$ 0.1820 \\
hl-ssm                                                       & 0.0354 $\pm$ 0.1015 & 0.0538 $\pm$ 0.1569 & -0.0675 $\pm$ 0.6327 & 0.0052 $\pm$ 0.1691  & 0.1082 $\pm$ 0.2391 \\
hl-sss                                                       & 0.0112 $\pm$ 0.0473 & 0.0146 $\pm$ 0.0912 & -0.0785 $\pm$ 0.7157 & -0.0004 $\pm$ 0.0626 & 0.0346 $\pm$ 0.1309 \\ \hline
\begin{tabular}[c]{@{}c@{}}Max\\ {[}1.8,-1{]}\end{tabular}   & \cellcolor[HTML]{C0C0C0}&  \cellcolor[HTML]{C0C0C0}&  \cellcolor[HTML]{C0C0C0}&  \cellcolor[HTML]{C0C0C0}&  \cellcolor[HTML]{C0C0C0}  \\  \hline
hl-mmm                                                       & \textbf{0.1233 $\pm$ 0.1686} & \textbf{0.1726 $\pm$ 0.2631} & -0.0757 $\pm$ 0.7022 & 0.0499 $\pm$ 0.3064  & \textbf{0.4199 $\pm$ 0.3448} \\
hl-mms                                                       & 0.0840 $\pm$ 0.1568 & 0.1062 $\pm$ 0.2106 & -0.0841 $\pm$ 0.6629 & 0.0445 $\pm$ 0.2628  & 0.2731 $\pm$ 0.2969 \\
hl-msm                                                       & 0.0989 $\pm$ 0.1665 & 0.1228 $\pm$ 0.2235 & \textbf{-0.0718 $\pm$ 0.6468} & 0.0567 $\pm$ 0.2575  & 0.3000 $\pm$ 0.3232 \\
hl-mss                                                       & 0.0595 $\pm$ 0.1266 & 0.0590 $\pm$ 0.1750 & -0.0742 $\pm$ 0.5778 & 0.0432 $\pm$ 0.1820  & 0.1691 $\pm$ 0.2542 \\
hl-smm                                                       & 0.0911 $\pm$ 0.1664 & 0.0935 $\pm$ 0.2100 & -0.0804 $\pm$ 0.6451 & \textbf{0.0610 $\pm$ 0.2305}  & 0.2514 $\pm$ 0.3178 \\
hl-sms                                                       & 0.0337 $\pm$ 0.0988 & 0.0184 $\pm$ 0.1444 & -0.0950 $\pm$ 0.5944 & 0.0211 $\pm$ 0.1115  & 0.0912 $\pm$ 0.2152 \\
hl-ssm                                                       & 0.0572 $\pm$ 0.1399 & 0.0417 $\pm$ 0.1702 & -0.0869 $\pm$ 0.5880 & 0.0472 $\pm$ 0.1609  & 0.1382 $\pm$ 0.2653 \\
hl-sss                                                       & 0.0203 $\pm$ 0.0761 & 0.0027 $\pm$ 0.1199 & -0.0933 $\pm$ 0.6324 & 0.0131 $\pm$ 0.0707  & 0.0494 $\pm$ 0.1639
\end{tabular}
\end{table}

\begin{table}[htp]
\centering
\tiny
\setlength{\tabcolsep}{2pt}
\caption{\lasa Complexity reduction compared to the Ori model complexity.}
\label{tabA:complexOri}
\begin{tabular}{l|ccccc}
to ori                                                             & SvdEn                & ApEn                 & SampEn              & CE                   & T. Shifts           \\ \hline
\begin{tabular}[c]{@{}c@{}}Avg.\\ {[}1.0,1.2{]}\end{tabular} &\cellcolor[HTML]{C0C0C0}&  \cellcolor[HTML]{C0C0C0}&  \cellcolor[HTML]{C0C0C0}&  \cellcolor[HTML]{C0C0C0}&  \cellcolor[HTML]{C0C0C0}  \\  \hline
hl-mmm                                                       & -0.2232 $\pm$ 0.3597 & \textbf{0.2004 $\pm$ 0.3475}  & \textbf{0.4849 $\pm$ 0.2975} & -0.1804 $\pm$ 0.5775 & \textbf{0.7171 $\pm$ 0.2484} \\
hl-mms                                                       & -0.2228 $\pm$ 0.3365 & 0.1519 $\pm$ 0.3309  & 0.4460 $\pm$ 0.2984 & -0.1414 $\pm$ 0.5222 & 0.6876 $\pm$ 0.2463 \\
hl-msm                                                       & -0.2140 $\pm$ 0.3414 & 0.1647 $\pm$ 0.3361  & 0.4534 $\pm$ 0.3021 & -0.1365 $\pm$ 0.5300 & 0.6952 $\pm$ 0.2506 \\
hl-mss                                                       & \textbf{-0.2081 $\pm$ 0.3284} & 0.1275 $\pm$ 0.3294  & 0.4210 $\pm$ 0.3049 & \textbf{-0.1055 $\pm$ 0.5085} & 0.6763 $\pm$ 0.2303 \\
hl-smm                                                       & -0.2328 $\pm$ 0.3844 & 0.1369 $\pm$ 0.3396  & 0.3999 $\pm$ 0.3544 & -0.1673 $\pm$ 0.6135 & 0.6881 $\pm$ 0.2467 \\
hl-sms                                                       & -0.2473 $\pm$ 0.4056 & 0.0932 $\pm$ 0.3520  & 0.3712 $\pm$ 0.3524 & -0.1660 $\pm$ 0.6152 & 0.6695 $\pm$ 0.2351 \\
hl-ssm                                                       & -0.2341 $\pm$ 0.3603 & 0.1094 $\pm$ 0.3318  & 0.3838 $\pm$ 0.3397 & -0.1464 $\pm$ 0.5466 & 0.6750 $\pm$ 0.2284 \\
hl-sss                                                       & -0.2519 $\pm$ 0.4207 & 0.0873 $\pm$ 0.3752  & 0.3898 $\pm$ 0.3222 & -0.1688 $\pm$ 0.6377 & 0.6670 $\pm$ 0.2436 \\ \hline
\begin{tabular}[c]{@{}c@{}}Avg.\\ {[}0.8,1.5{]}\end{tabular} & \cellcolor[HTML]{C0C0C0}&  \cellcolor[HTML]{C0C0C0}&  \cellcolor[HTML]{C0C0C0}&  \cellcolor[HTML]{C0C0C0}&  \cellcolor[HTML]{C0C0C0}  \\  \hline
hl-mmm                                                       & -0.0766 $\pm$ 0.4164 & 0.3044 $\pm$ 0.4013  & 0.3911 $\pm$ 0.5135 & -0.1357 $\pm$ 0.7047 & 0.8321 $\pm$ 0.1508 \\
hl-mms                                                       & 0.0253 $\pm$ 0.4081  & 0.4186 $\pm$ 0.4367  & 0.4745 $\pm$ 0.5296 & -0.0729 $\pm$ 0.6608 & 0.8734 $\pm$ 0.1231 \\
hl-msm                                                       & 0.0083 $\pm$ 0.4133  & 0.4058 $\pm$ 0.4282  & 0.4668 $\pm$ 0.5189 & -0.0984 $\pm$ 0.6953 & 0.8727 $\pm$ 0.1214 \\
hl-mss                                                       & 0.0854 $\pm$ 0.3707  & 0.5204 $\pm$ 0.4118  & 0.5752 $\pm$ 0.4689 & -0.0434 $\pm$ 0.6452 & 0.9091 $\pm$ 0.1046 \\
hl-smm                                                       & 0.0585 $\pm$ 0.3948  & 0.4682 $\pm$ 0.4216  & 0.5211 $\pm$ 0.5096 & -0.0694 $\pm$ 0.6674 & 0.9024 $\pm$ 0.1003 \\
hl-sms                                                       & 0.1428 $\pm$ 0.3471  & 0.6362 $\pm$ 0.3488  & 0.7036 $\pm$ 0.3493 & -0.0223 $\pm$ 0.6297 & 0.9487 $\pm$ 0.0647 \\
hl-ssm                                                       & 0.1271 $\pm$ 0.3561  & 0.5981 $\pm$ 0.3671  & 0.6576 $\pm$ 0.4030 & -0.0217 $\pm$ 0.6346 & 0.9390 $\pm$ 0.0738 \\
hl-sss                                                       & \textbf{0.1770 $\pm$ 0.3333}  & \textbf{0.7106 $\pm$ 0.3117}  & \textbf{0.7687 $\pm$ 0.2822} & \textbf{0.0176 $\pm$ 0.6061}  & \textbf{0.9629 $\pm$ 0.0527} \\ \hline
\begin{tabular}[c]{@{}c@{}}Max.\\ {[}2,3{]}\end{tabular}     & \cellcolor[HTML]{C0C0C0}&  \cellcolor[HTML]{C0C0C0}&  \cellcolor[HTML]{C0C0C0}&  \cellcolor[HTML]{C0C0C0}&  \cellcolor[HTML]{C0C0C0}  \\  \hline
hl-mmm                                                       & \textbf{-0.2736 $\pm$ 0.4730} &\textbf{ 0.1541 $\pm$ 0.5552}  & \textbf{0.4739 $\pm$ 0.2996} & -0.2832 $\pm$ 0.6488 & \textbf{0.6858 $\pm$ 0.3010} \\
hl-mms                                                       & -0.3378 $\pm$ 0.4679 & -0.0116 $\pm$ 0.5978 & 0.3835 $\pm$ 0.3370 & -0.2717 $\pm$ 0.6177 & 0.5749 $\pm$ 0.3367 \\
hl-msm                                                       & -0.2952 $\pm$ 0.4553 & 0.0371 $\pm$ 0.5934  & 0.4069 $\pm$ 0.3455 & \textbf{-0.2348 $\pm$ 0.5857} & 0.6074 $\pm$ 0.3282 \\
hl-mss                                                       & -0.3692 $\pm$ 0.4694 & -0.1077 $\pm$ 0.6559 & 0.3375 $\pm$ 0.3842 & -0.2834 $\pm$ 0.6205 & 0.5157 $\pm$ 0.3460 \\
hl-smm                                                       & -0.3263 $\pm$ 0.4756 & -0.0344 $\pm$ 0.6449 & 0.3702 $\pm$ 0.3841 & -0.2635 $\pm$ 0.6090 & 0.5704 $\pm$ 0.3496 \\
hl-sms                                                       & -0.3999 $\pm$ 0.4651 & -0.1594 $\pm$ 0.6832 & 0.3153 $\pm$ 0.4055 & -0.3116 $\pm$ 0.6206 & 0.4750 $\pm$ 0.3597 \\
hl-ssm                                                       & -0.3737 $\pm$ 0.4767 & -0.1286 $\pm$ 0.6852 & 0.3336 $\pm$ 0.3953 & -0.2906 $\pm$ 0.6277 & 0.5051 $\pm$ 0.3564 \\
hl-sss                                                       & -0.4074 $\pm$ 0.4596 & -0.1849 $\pm$ 0.6970 & 0.3048 $\pm$ 0.4068 & -0.3095 $\pm$ 0.6211 & 0.4587 $\pm$ 0.3598 \\ \hline
\begin{tabular}[c]{@{}c@{}}Max\\ {[}1.8,-1{]}\end{tabular}   & \cellcolor[HTML]{C0C0C0}&  \cellcolor[HTML]{C0C0C0}&  \cellcolor[HTML]{C0C0C0}&  \cellcolor[HTML]{C0C0C0}&  \cellcolor[HTML]{C0C0C0}  \\  \hline
hl-mmm                                                       & \textbf{-0.2472 $\pm$ 0.4711} & \textbf{0.1142 $\pm$ 0.4926}  & \textbf{0.3715 $\pm$ 0.3775} & -0.2287 $\pm$ 0.6580 & \textbf{0.7073 $\pm$ 0.2768} \\
hl-mms                                                       & -0.3022 $\pm$ 0.4720 & 0.0121 $\pm$ 0.5464  & 0.3671 $\pm$ 0.3254 & -0.2357 $\pm$ 0.6267 & 0.6106 $\pm$ 0.3264 \\
hl-msm                                                       & -0.2758 $\pm$ 0.4569 & 0.0307 $\pm$ 0.5533  & 0.3643 $\pm$ 0.3532 & \textbf{-0.2118 $\pm$ 0.6004} & 0.6334 $\pm$ 0.3135 \\
hl-mss                                                       & -0.3427 $\pm$ 0.4839 & -0.0752 $\pm$ 0.6162 & 0.3440 $\pm$ 0.3515 & -0.2525 $\pm$ 0.6357 & 0.5452 $\pm$ 0.3431 \\
hl-smm                                                       & -0.2917 $\pm$ 0.4721 & -0.0235 $\pm$ 0.6032 & 0.3465 $\pm$ 0.3747 & -0.2193 $\pm$ 0.6089 & 0.6039 $\pm$ 0.3248 \\
hl-sms                                                       & -0.3825 $\pm$ 0.4791 & -0.1461 $\pm$ 0.6711 & 0.3148 $\pm$ 0.3909 & -0.2936 $\pm$ 0.6349 & 0.4948 $\pm$ 0.3567 \\
hl-ssm                                                       & -0.3483 $\pm$ 0.4975 & -0.1128 $\pm$ 0.6698 & 0.3280 $\pm$ 0.3769 & -0.2573 $\pm$ 0.6470 & 0.5270 $\pm$ 0.3545 \\
hl-sss                                                       & -0.3989 $\pm$ 0.4697 & -0.1785 $\pm$ 0.6909 & 0.3035 $\pm$ 0.4032 & -0.3022 $\pm$ 0.6286 & 0.4670 $\pm$ 0.3605
\end{tabular}
\end{table}

\FloatBarrier
\section{Shapelet \lasa}

\begin{table}[h!]
\centering
\footnotesize
\setlength{\tabcolsep}{2pt}
\caption{Relative accuracy to the good SAX models for the \lasa Shapelets model with slen 2.}
\begin{tabular}{l|cccc}
Good & Avg. {[}1,1.2{]} & Avg. {[}0.8,1.5{]} & Max. {[}2,3{]} & Max. {[}1.8,-1{]} \\ \hline
hl-mmm & -0.3186 $\pm$ 0.2284 & \textbf{-0.3534 $\pm$ 0.1275} & -0.3594 $\pm$ 0.2503 & -0.3929 $\pm$ 0.2510 \\
hl-mms & -0.2602 $\pm$ 0.1959 & -0.4235 $\pm$ 0.1562 & -0.1913 $\pm$ 0.2100 & -0.2951 $\pm$ 0.2465 \\
hl-msm & -0.3020 $\pm$ 0.2266 & -0.3929 $\pm$ 0.1659 & -0.2649 $\pm$ 0.2659 & -0.3121 $\pm$ 0.2704 \\
hl-mss & -0.2383 $\pm$ 0.1986 & -0.4634 $\pm$ 0.1906 & -0.1290 $\pm$ 0.1958 & -0.1884 $\pm$ 0.2220 \\
hl-smm & -0.2907 $\pm$ 0.1869 & -0.4324 $\pm$ 0.1718 & -0.2406 $\pm$ 0.2427 & -0.3055 $\pm$ 0.2547 \\
hl-sms & -0.2650 $\pm$ 0.1822 & -0.5088 $\pm$ 0.1961 & -0.0937 $\pm$ 0.1879 & -0.1160 $\pm$ 0.1809 \\
hl-ssm & -0.2664 $\pm$ 0.1983 & -0.5066 $\pm$ 0.1859 & -0.1610 $\pm$ 0.2175 & -0.2048 $\pm$ 0.2516 \\
hl-sss &\textbf{ -0.2364 $\pm$ 0.2080} & -0.5201 $\pm$ 0.1758 & \textbf{-0.0691 $\pm$ 0.1720} & \textbf{-0.0747 $\pm$ 0.1719}
\end{tabular}
\label{tabA:lasaShapAcc}
\end{table}
\clearpage
\section{\gcr Results}

\begin{table}[h!]
\centering
\footnotesize
\setlength{\tabcolsep}{4pt}
\caption{Average \gcr performance grouped by \laam configurations, relative to the SAX models. Including only the good models (accuracy $>75\%$).}
\begin{tabular}{l|cccc}
Good & Accuracy & Precision & Recall & F1 \\ \hline
hl-mm & -0.0337 $\pm$ 0.1300 & -0.0459 $\pm$ 0.1334 & -0.0386 $\pm$ 0.1302 & -0.0610 $\pm$ 0.1479 \\
hl-ms & -0.0305 $\pm$ 0.1320 & -0.0418 $\pm$ 0.1339 & -0.0352 $\pm$ 0.1292 & -0.0573 $\pm$ 0.1496 \\
hl-sm & -0.0209 $\pm$ 0.1339 & -0.0360 $\pm$ 0.1385 & -0.0289 $\pm$ 0.1312 & -0.0495 $\pm$ 0.1520 \\
hl-ss & \textbf{-0.0172 $\pm$ 0.1340} & \textbf{-0.0324 $\pm$ 0.1386} & \textbf{-0.0267 $\pm$ 0.1327} & \textbf{-0.0457 $\pm$ 0.1521}
\end{tabular}

\label{tabA:gcrPerfGood}
\end{table}

\begin{table}[h!]
\centering
\footnotesize
\setlength{\tabcolsep}{5pt}
\caption{Average \gcr performance grouped by \laam configurations, relative to the SAX models. Considering all models.}
\begin{tabular}{l|cccc}
All & Accuracy & Precision & Recall & F1 \\ \hline
hl-mm & 0.0551 $\pm$ 0.1757 & 0.0479 $\pm$ 0.1749 & 0.0716 $\pm$ 0.1815 & 0.0268 $\pm$ 0.1874 \\
hl-ms & 0.0554 $\pm$ 0.1757 & 0.0500 $\pm$ 0.1737 & 0.0762 $\pm$ 0.1788 & 0.0290 $\pm$ 0.1869 \\
hl-sm & 0.0613 $\pm$ 0.1739 & 0.0537 $\pm$ 0.1735 & 0.0787 $\pm$ 0.1779 & 0.0339 $\pm$ 0.1860 \\
hl-ss &\textbf{ 0.0630 $\pm$ 0.1734} & \textbf{0.0562 $\pm$ 0.1727} & \textbf{0.0826 $\pm$ 0.1763} & \textbf{0.0374 $\pm$ 0.1851}
\end{tabular}
\label{tabA:gcrPerf}
\end{table}

\begin{table}[h!]
\centering
\footnotesize
\caption{Number of datasets for which type of \gcr performed best.}
\begin{tabular}{l|cccc|cccc}
 & All &  &  &  & Good &  &  &  \\
 & hl-mm & hl-ms & hl-sm & hl-ss & hl-mm & hl-ms & hl-sm & hl-ss \\ \hline
\fcam & \cellcolor[HTML]{C0C0C0} & \cellcolor[HTML]{C0C0C0} & \cellcolor[HTML]{C0C0C0} & \cellcolor[HTML]{C0C0C0} &\cellcolor[HTML]{C0C0C0}   &\cellcolor[HTML]{C0C0C0}   &\cellcolor[HTML]{C0C0C0}   &\cellcolor[HTML]{C0C0C0}   \\ \hline
Sum & 70 & 67 & 64 & 55 & 19 & 18 & 17 & 13 \\
r. Avg. & 44 & 47 & 51 & 65 & 8 & 11 & 15 & 19 \\ \hline
\gtm & \cellcolor[HTML]{C0C0C0} & \cellcolor[HTML]{C0C0C0} & \cellcolor[HTML]{C0C0C0} & \cellcolor[HTML]{C0C0C0} & \cellcolor[HTML]{C0C0C0} & \cellcolor[HTML]{C0C0C0}   &  \cellcolor[HTML]{C0C0C0}  &  \cellcolor[HTML]{C0C0C0} \\ \hline
Max of sum & 16 & 19 & 19 & 20 & 2 & 2 & 2 & 4 \\
Max of r. avg. & 8 & 15 & 14 & 15 & 0 & 0 & 0 & 0 \\
Avg. of sum & 6 & 1 & 2 & 2 & 1 & 0 & 0 & 0 \\
Avg. of r. avg. & 9 & 10 & 12 & 7 & 5 & 6 & 3 & 1 \\
Med. of sum & 8 & 5 & 2 & 3 & 3 & 1 & 1 & 1 \\
Med. of r. avg. & 9 & 6 & 6 & 4 & 0 & 0 & 0 & 0
\end{tabular}

\label{tabA:grcCount}
\end{table}

\eat{
\begin{table}[h!]
\centering
\footnotesize
\caption{Effect of the \gcr certainty on the train accuracy of the good models.}
\begin{tabular}{l|cccc}
Good & hl-mm & hl-ms & hl-sm & hl-ss \\ \hline
Train Cert. 100 & ? & ? & ? & ? \\
Train Cert. 80 & 0.8761 $\pm$ 0.1290 & 0.8815 $\pm$ 0.1290 & 0.8953 $\pm$ 0.1244 & 0.9036 $\pm$ 0.1240 \\
Train Cert. 50 & 0.8831 $\pm$ 0.1317 & 0.8868 $\pm$ 0.1279 & 0.9029 $\pm$ 0.1186 & 0.9150 $\pm$ 0.1189 \\
Train Cert. 20 & 0.8994 $\pm$ 0.1344 & 0.9040 $\pm$ 0.1350 & 0.9196 $\pm$ 0.1111 & 0.9371 $\pm$ 0.1046 \\
Train Cert. 10 & 0.9086 $\pm$ 0.1456 & 0.9103 $\pm$ 0.1467 & 0.9270 $\pm$ 0.1099 & 0.9495 $\pm$ 0.0922
\end{tabular}

\label{tabA:certTrain}
\end{table}
}

\begin{table}[h!]
\centering
\footnotesize
\setlength{\tabcolsep}{5pt}
\caption{\gcr model fidelity for the good performing models split between the different model configurations.}
\begin{tabular}{l|cccc}
Good & \begin{tabular}[c]{@{}c@{}}Ori Train \\ Model Fidelity\end{tabular} & \begin{tabular}[c]{@{}c@{}}Ori Test \\ Model Fidelity\end{tabular} & \begin{tabular}[c]{@{}c@{}}SAX Train \\ Model Fidelity\end{tabular} & \begin{tabular}[c]{@{}c@{}}SAX Test \\ Model Fidelity\end{tabular} \\ \hline
All & 0.7497 $\pm$ 0.1958 & 0.7047 $\pm$ 0.1939 & 0.8259 $\pm$ 0.1536 & 0.7900 $\pm$ 0.1586 \\
3s-2l-8h & 0.7072 $\pm$ 0.1462 & 0.6429 $\pm$ 0.1646 & 0.8124 $\pm$ 0.1420 & 0.7639 $\pm$ 0.1847 \\
3s-2l-16h & 0.7276 $\pm$ 0.2266 & 0.6623 $\pm$ 0.2238 & \textbf{0.8689 $\pm$ 0.1197} & 0.8199 $\pm$ 0.1413 \\
3s-5l-6h & 0.7840 $\pm$ 0.1206 & 0.7129 $\pm$ 0.1469 & 0.8111 $\pm$ 0.1163 & 0.7417 $\pm$ 0.1412 \\
5s-2l-8h & 0.7912 $\pm$ 0.1412 & 0.7513 $\pm$ 0.1344 & 0.8243 $\pm$ 0.1055 & 0.7959 $\pm$ 0.1015 \\
5s-2l-16h & 0.6596 $\pm$ 0.2602 & 0.6149 $\pm$ 0.2547 & 0.7857 $\pm$ 0.2446 & 0.7561 $\pm$ 0.2381 \\
5s-5l-6h & \textbf{0.8247 $\pm$ 0.1570} & \textbf{0.8118 $\pm$ 0.1099} & 0.8596 $\pm$ 0.0940 & \textbf{0.8381 $\pm$ 0.0635}
\end{tabular}

\label{tabA:gcrFidelity}
\end{table}

\eat{
\begin{table}[h!]
\centering
\footnotesize
\caption{Train certainty using a selective approach, including \gcr, Counting \pgcr and Entropy \pgcr.}
\begin{tabular}{l|cccc}
Good & hl-mm & hl-ms & hl-sm & hl-ss \\ \hline
Train Cert. 100 & ? & ? & ? & ? \\
Train Cert. 80 & 0.9038 $\pm$ 0.0992 & 0.9097 $\pm$ 0.0995 & 0.9195 $\pm$ 0.1016 & 0.9232 $\pm$ 0.1031 \\
Train Cert. 50 & 0.9191 $\pm$ 0.0987 & 0.9306 $\pm$ 0.0971 & 0.9352 $\pm$ 0.0927 & 0.9376 $\pm$ 0.0944 \\
Train Cert. 20 & 0.9507 $\pm$ 0.0866 & 0.9568 $\pm$ 0.0826 & 0.9565 $\pm$ 0.0870 & 0.9589 $\pm$ 0.0847 \\
Train Cert. 10 & 0.9667 $\pm$ 0.0754 & 0.9683 $\pm$ 0.0734 & 0.9674 $\pm$ 0.0753 & 0.9687 $\pm$ 0.0735
\end{tabular}
\label{tabA:pgrcTrainCert}
\end{table}
}

\begin{table}[h!]
\centering
\footnotesize
\setlength{\tabcolsep}{5pt}
\caption{\gcr model performance using a selective approach, including \gcr, Counting \pgcr and Entropy \pgcr. For only the good performing models.}
\begin{tabular}{l|cccc}
Good & Accuracy & Precision & Recall & F1 \\ \hline
hl-mm & -0.0072 $\pm$ 0.1164 & -0.0343 $\pm$ 0.1295 & -0.0277 $\pm$ 0.1282 & -0.0447 $\pm$ 0.1371 \\
hl-ms & -0.0028 $\pm$ 0.1164 & -0.0296 $\pm$ 0.1295 & -0.0229 $\pm$ 0.1272 & -0.0402 $\pm$ 0.1380 \\
hl-sm & 0.0010 $\pm$ 0.1194 & -0.0264 $\pm$ 0.1339 & -0.0229 $\pm$ 0.1282 & -0.0366 $\pm$ 0.1421 \\
hl-ss & \textbf{0.0039 $\pm$ 0.1203} & \textbf{-0.0226 $\pm$ 0.1346} & \textbf{-0.0206 $\pm$ 0.1304} & \textbf{-0.0334 $\pm$ 0.1429}
\end{tabular}
\label{tab:pgrcPerformanceGood}
\end{table}

\begin{table}[h!]
\centering
\footnotesize
\caption{\gcr model performance using a selective approach, including \gcr, Counting \pgcr and Entropy \pgcr.}
\begin{tabular}{l|cccc}
All & Accuracy & Precision & Recall & F1 \\ \hline
hl-mm & 0.0757 $\pm$ 0.1635 & 0.0603 $\pm$ 0.1722 & 0.0854 $\pm$ 0.1785 & 0.0434 $\pm$ 0.1809 \\
hl-ms & 0.0778 $\pm$ 0.1635 & 0.0644 $\pm$ 0.1715 & 0.0919 $\pm$ 0.1752 & 0.0470 $\pm$ 0.1804 \\
hl-sm & 0.0829 $\pm$ 0.1630 & 0.0681 $\pm$ 0.1731 & 0.0932 $\pm$ 0.1760 & 0.0518 $\pm$ 0.1817 \\
hl-ss & \textbf{0.0850 $\pm$ 0.1614} & \textbf{0.0696 $\pm$ 0.1723} & \textbf{0.0975 $\pm$ 0.1747} & \textbf{0.0537 $\pm$ 0.1812}
\end{tabular}
\label{tab:pgrcPerformance}
\end{table}

\begin{table}[h!]
\centering
\footnotesize
\caption{Number of datasets, how often which kind of \gcr variant performed best.}
\begin{tabular}{l|cccc|cccc}
 & All &  &  &  & Good &  &  &  \\
 & hl-mm & hl-ms & hl-sm & hl-ss & hl-mm & hl-ms & hl-sm & hl-ss \\ \hline
\gcr & \cellcolor[HTML]{FFFFFF}62 & \cellcolor[HTML]{FFFFFF}60 & \cellcolor[HTML]{FFFFFF}64 & \cellcolor[HTML]{FFFFFF}67 & \cellcolor[HTML]{FFFFFF}10 & \cellcolor[HTML]{FFFFFF}10 & \cellcolor[HTML]{FFFFFF}11 & \cellcolor[HTML]{FFFFFF}15 \\
Entropy \pgcr & \cellcolor[HTML]{FFFFFF}46 & \cellcolor[HTML]{FFFFFF}45 & \cellcolor[HTML]{FFFFFF}39 & \cellcolor[HTML]{FFFFFF}40 & \cellcolor[HTML]{FFFFFF}15 & \cellcolor[HTML]{FFFFFF}15 & \cellcolor[HTML]{FFFFFF}14 & \cellcolor[HTML]{FFFFFF}11 \\
Counting \pgcr & \cellcolor[HTML]{FFFFFF}62 & \cellcolor[HTML]{FFFFFF}65 & \cellcolor[HTML]{FFFFFF}67 & \cellcolor[HTML]{FFFFFF}63 & \cellcolor[HTML]{FFFFFF}13 & \cellcolor[HTML]{FFFFFF}13 & \cellcolor[HTML]{FFFFFF}13 & \cellcolor[HTML]{FFFFFF}12
\end{tabular}
\label{tabA:pgcrCount}
\end{table}

\begin{table}[h!]
\centering
\footnotesize
\setlength{\tabcolsep}{5pt}
\caption{Performance of the \tgcr, using different \laam configurations. \textit{All \tgcr} is a selective approach over all \tgcrs options.}
\begin{tabular}{l|cccc}
 & \gcr Avg. t1.0 & \gcr Avg. t1.3 & \gcr Avg. t1.6 & All \tgcr \\ \hline
All & \cellcolor[HTML]{C0C0C0} & \cellcolor[HTML]{C0C0C0} & \cellcolor[HTML]{C0C0C0} & \cellcolor[HTML]{C0C0C0} \\ \hline
hl-mm & 0.0377 $\pm$ 0.1811 & 0.0525 $\pm$ 0.1771 & 0.0539 $\pm$ 0.1770 & 0.0674 $\pm$ 0.1693 \\
hl-ms & 0.0411 $\pm$ 0.1777 & 0.0565 $\pm$ 0.1745 & 0.0554 $\pm$ 0.1756 & 0.0698 $\pm$ 0.1663 \\
hl-sm & 0.0494 $\pm$ 0.1722 & 0.0612 $\pm$ 0.1726 & 0.0618 $\pm$ 0.1733 & 0.0742 $\pm$ 0.1645 \\
hl-ss & \textbf{0.0606 $\pm$ 0.1633} & \textbf{0.0637 $\pm$ 0.1718} & \textbf{0.0631 $\pm$ 0.1730} & \textbf{0.0759 $\pm$ 0.1637} \\ \hline
Good & \cellcolor[HTML]{C0C0C0} & \cellcolor[HTML]{C0C0C0} & \cellcolor[HTML]{C0C0C0} & \cellcolor[HTML]{C0C0C0} \\ \hline
hl-mm & -0.0696 $\pm$ 0.1395 & -0.0379 $\pm$ 0.1257 & -0.0327 $\pm$ 0.1266 & -0.0237 $\pm$ 0.1231 \\
hl-ms & -0.0646 $\pm$ 0.1372 & -0.0283 $\pm$ 0.1283 & -0.0293 $\pm$ 0.1290 & -0.0168 $\pm$ 0.1203 \\
hl-sm & -0.0478 $\pm$ 0.1291 & -0.0194 $\pm$ 0.1288 & -0.0200 $\pm$ 0.1320 & -0.0093 $\pm$ 0.1215 \\
hl-ss & \textbf{-0.0200 $\pm$ 0.1258} & \textbf{-0.0158 $\pm$ 0.1307} & \textbf{-0.0177 $\pm$ 0.1331} & \textbf{-0.0058 $\pm$ 0.1241}
\end{tabular}

\label{tabA:tgcrPerf}
\end{table}

\begin{table}[h!]
\centering
\footnotesize
\caption{Number of datasets where the different \tgcrs performed best}
\label{tabA:countingTGCR}
\begin{tabular}{l|ccc}
 & \gcr Avg. t1.0 & \gcr Avg. t1.3 & \gcr Avg. t1.6 \\ \hline
All & \cellcolor[HTML]{C0C0C0} & \cellcolor[HTML]{C0C0C0} & \cellcolor[HTML]{C0C0C0} \\ \hline
hl-mm & 54 & 65 & 51 \\
hl-ms & 64 & 71 & 35 \\
hl-sm & 70 & 76 & 24 \\
hl-ss & 90 & 68 & 22 \\ \hline
Good & \cellcolor[HTML]{C0C0C0} & \cellcolor[HTML]{C0C0C0} & \cellcolor[HTML]{C0C0C0} \\ \hline
hl-mm & 9 & 12 & 17 \\
hl-ms & 9 & 19 & 10 \\
hl-sm & 11 & 17 & 10 \\
hl-ss & 13 & 17 & 8
\end{tabular}
\end{table}
\clearpage
\section{\gcr for Interpretation}
\begin{table}[h!]
\centering
\footnotesize
\caption{SAX model fidelity of the \gcr variations (all) when interpreting the SAX model results.}
\begin{tabular}{l|cc}
All & \begin{tabular}[c]{@{}c@{}}SAX Train\\ Model Fidelity\end{tabular} & \begin{tabular}[c]{@{}c@{}}SAX Test\\ Model Fidelity\end{tabular} \\ \hline
hl-mm & 0.9410 $\pm$ 0.0807 & \textbf{0.8778 $\pm$ 0.0938} \\
hl-ms & 0.9446 $\pm$ 0.0787 & 0.8809 $\pm$ 0.0910 \\
hl-sm & 0.9469 $\pm$ 0.0811 & 0.8804 $\pm$ 0.0929 \\
hl-ss & \textbf{0.9497 $\pm$ 0.0807} & 0.8823 $\pm$ 0.0925
\end{tabular}
\label{tabA:saxOptGCRFi}
\end{table}

\begin{table}[h!]
\centering
\footnotesize
\caption{SAX model fidelity of the Shapelets model if it is trained to approximate the SAX-based Transformer model.}
\begin{tabular}{c|cc}
\begin{tabular}[c]{@{}c@{}}Shapelets\\ fits SAX Model\end{tabular} & \begin{tabular}[c]{@{}c@{}}SAX Train\\ Model Fidelity\end{tabular} & \begin{tabular}[c]{@{}c@{}}SAX Test\\ Model Fidelity\end{tabular} \\ \hline
\begin{tabular}[c]{@{}c@{}}All\\ slen 2\end{tabular} & \cellcolor[HTML]{C0C0C0} & \cellcolor[HTML]{C0C0C0} \\ \hline
Ori & 0.9639 $\pm$ 0.1645 & 0.7739 $\pm$ 0.1739 \\
SAX 5 & 0.9321 $\pm$ 0.1894 & 0.7576 $\pm$ 0.1969 \\
SAX 3 & 0.9289 $\pm$ 0.1770 & 0.7891 $\pm$ 0.1810 \\ \hline
\begin{tabular}[c]{@{}c@{}}All\\ slen 0.3\end{tabular} & \cellcolor[HTML]{C0C0C0} & \cellcolor[HTML]{C0C0C0} \\ \hline
Ori & \textbf{0.9675 $\pm$ 0.1645} & 0.7797 $\pm$ 0.1708 \\
SAX 5 & 0.9496 $\pm$ 0.1869 & 0.7897 $\pm$ 0.1853 \\
SAX 3 & 0.9638 $\pm$ 0.1668 & \textbf{0.8236 $\pm$ 0.1734} \\ \hline
\begin{tabular}[c]{@{}c@{}}Good \\ slen 2\end{tabular} & \cellcolor[HTML]{C0C0C0} & \cellcolor[HTML]{C0C0C0} \\ \hline
Ori & 0.9959 $\pm$ 0.0141 & 0.8343 $\pm$ 0.1430 \\
SAX 5 & 0.9657 $\pm$ 0.0699 & 0.8061 $\pm$ 0.1712 \\
SAX 3 & 0.9310 $\pm$ 0.1286 & 0.7856 $\pm$ 0.1756 \\ \hline
\begin{tabular}[c]{@{}c@{}}Good\\ slen 0.3\end{tabular} & \cellcolor[HTML]{C0C0C0} & \cellcolor[HTML]{C0C0C0} \\ \hline
Ori & \textbf{1.0000 $\pm$ 0.0000} & 0.8396 $\pm$ 0.1377 \\
SAX 5 & 0.9982 $\pm$ 0.0050 & \textbf{ 0.8566 $\pm$ 0.1222} \\
SAX 3 & 0.9981 $\pm$ 0.0046 & 0.8292 $\pm$ 0.1551
\end{tabular}
\label{tabA:saxShapeletsFitFi}
\end{table}

\begin{table}[h!]
\centering
\footnotesize
\caption{Number of the best performances of the \gcr variations for each \laam setting, when learning based on the SAX model predictions.}
\begin{tabular}{l|cccc|cccc}
 & All &  &  &  & Good &  &  &  \\
 & hl-mm & hl-ms & hl-sm & hl-ss & hl-mm & hl-ms & hl-sm & hl-ss \\ \hline
\gcr & 49 & 56 & 63 & 62 & 3 & 7 & 6 & 8 \\
Entropy \pgcr & 44 & 42 & 41 & 36 & 16 & 17 & 18 & 18 \\
Counting \pgcr & 41 & 37 & 45 & 38 & 9 & 9 & 10 & 9 \\
1.0 \tgcr & 16 & 24 & 14 & 25 & 4 & 4 & 2 & 2 \\
1.3 \tgcr & 17 & 10 & 5 & 7 & 6 & 0 & 1 & 0 \\
1.6 \tgcr & 3 & 1 & 2 & 2 & 0 & 1 & 1 & 1
\end{tabular}
\label{tabA:gcrSaxFitCount}
\end{table}



\end{appendices}